\documentclass[final]{cvpr}

\usepackage{times}
\usepackage{epsfig}
\usepackage{graphicx}
\usepackage{amsmath}
\usepackage{amssymb}
\usepackage{microtype}

\usepackage{multirow}
\usepackage[table,xcdraw]{xcolor}
\usepackage{gensymb}
\usepackage{tabularx, booktabs}
\usepackage{subcaption}

\usepackage{caption}
\newbox\jsavebox
\newcommand{\jsubfig}[2]{%
	\sbox\jsavebox{#1}%
	\parbox[t]{\wd\jsavebox}{\centering\usebox\jsavebox\\#2}%
	}
\newcommand{\whitetxt}[1]{{\color{white}#1}\normalfont}

\usepackage[pagebackref=true,breaklinks=true,colorlinks,bookmarks=false]{hyperref}

\def\cvprPaperID{1287} 
\def\confYear{CVPR 2021}

\begin{document}

\title{Extreme Rotation Estimation using Dense Correlation Volumes }

\author{
Ruojin Cai$^1$  \ \ \ 
Bharath Hariharan$^1$ \ \ \
Noah Snavely$^{1,2}$ \ \ \
Hadar Averbuch-Elor$^{1,2}$
\\[2mm]
\vspace{1em}
$^1$Cornell University \ \ \
$^2$Cornell Tech
}
\maketitle

\begin{abstract}
We present a technique for estimating the relative 3D rotation of an RGB image pair in an extreme setting, where the images have little or no overlap. 
We observe that, even when images do not overlap, there may be rich hidden cues as to their geometric relationship, such as light source directions, vanishing points, and symmetries present in the scene.
We propose a network design that can automatically learn such implicit cues by comparing all pairs of points between the two input images. 
Our method therefore constructs dense feature correlation volumes 
and processes these to predict relative 3D rotations. 
Our predictions are formed over a fine-grained discretization of rotations, bypassing difficulties associated with regressing 3D rotations.
We demonstrate our approach on a large variety of extreme RGB image pairs, including indoor and outdoor images captured under different lighting conditions and geographic locations.
Our evaluation shows that our model can successfully estimate relative rotations among non-overlapping images without compromising performance over overlapping image pairs.\footnote{\url{https://ruojincai.github.io/ExtremeRotation/}}
\end{abstract}


\section{Introduction}

Estimating the relative pose between a pair of RGB images is a fundamental task in computer vision with applications including 3D reconstruction~\cite{schonberger2016structure,ozyesil2017survey}, camera localization~\cite{brachmann2017dsac,schonberger2018semantic,taira2018inloc}, simultaneous localization and mapping~\cite{davison2007monoslam,mur2015orb} and novel view synthesis~\cite{mildenhall2020nerf, riegler2020fvs}. 
Standard methods for computing relative pose are highly dependent on accurate correspondence.  
But what if the poses are so different that there is no overlap and hence no correspondence?

\begin{figure}
\begin{center}
\includegraphics[trim=0 10 10 0, clip,width=0.95\columnwidth]{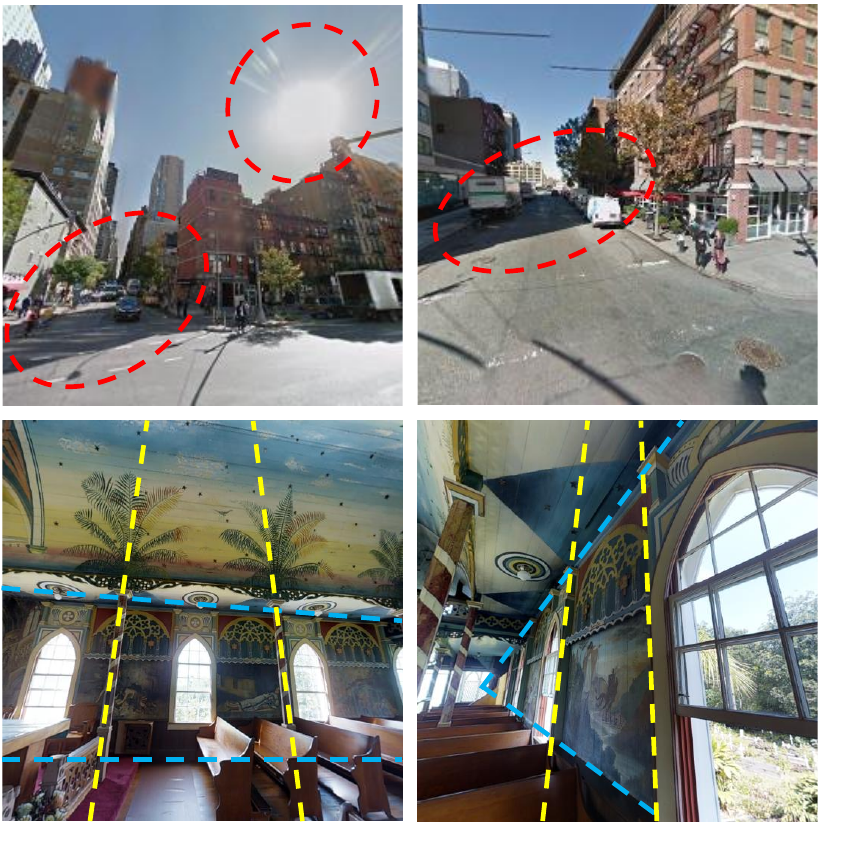}
\end{center}
\vspace{-8pt}
   \caption{\textbf{How can we estimate relative rotation between images in extreme non-overlapping cases?} Above we show two non-overlapping image pairs capturing an urban street scene (top) and a church (bottom). Possible cues to their relationship include sunlight and direction of shadows in outdoor scenes (highlighted in red) and lines parallel in 3D in indoor scenes (marked with yellow and blue line segments), from which vanishing points can be derived. 
}
\label{fig:example}
\end{figure}
\begin{figure*}
\centering
\includegraphics[trim=0 3 0 8, clip,width=\textwidth]{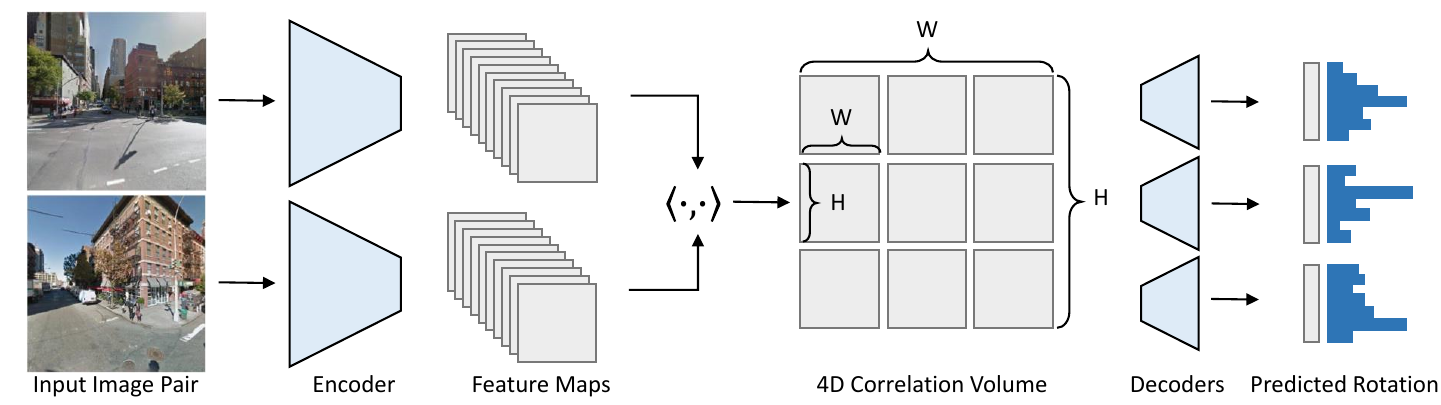}
   \caption{\textbf{Method overview.} 
    Given a pair of images, a shared-weight Siamese encoder extracts feature maps. We compute a 4D correlation volume using the inner product of features, from which our model predicts the relative rotation (here, as distributions over Euler angles). 
    }
\label{fig:overview}
\end{figure*}

Our work takes a step towards this seemingly impossible goal of estimating relative pose for pairs of RGB images that have little or no overlap.
In particular, we present a technique for estimating relative 3D rotation for a pair of images with (possibly) extreme relative motion.
There are many applications where dense imagery is difficult to obtain that can benefit from rotation estimation from non-overlapping views \cite{ApartmentsCVPR15, elor2015ringit}.
For example, when advertising homes on online real estate sites, 
users may only provide a small number of images---too sparse for current 3D reconstruction methods. Rotation estimation can simplify downstream tasks, such as 3D reconstruction from sparse views.

How can we reason about relative rotation in  extreme non-overlapping settings? 
As humans, there are a number of cues we might leverage. 
Consider the two image pairs in Fig.~\ref{fig:example}.
For the outdoor pair, we can infer relative orientation using illumination cues, e.g., by analyzing which buildings are lit or the directions of cast shadows. 
Geometric cues are also useful. 
For example, from the pair of indoor images we can infer a change in camera pitch from the set of parallel vertical lines in 3D (colored in yellow) that suggest a vanishing point, and  
we can infer a rightward camera rotation by analyzing symmetries and the layout of the benches.

Given the presence of such ``hidden'' cues, one approach to computing relative rotation would be to explicitly learn such cues via 
supervision, e.g., by labeling vanishing points and learning to predict them. %
However, in addition to the drawbacks of requiring additional supervision, we do not want to restrict our model to a set of handcrafted cues which may or may not be relevant for the image pair provided at test time.
Instead, we want to learn to predict relative rotation from pose supervision alone. As such, we ask: Can we guide the network to reason about such hidden cues implicitly? And what architecture would best achieve this goal? 

Our key insight is that reasoning about cues such as vanishing points and illumination---while not achievable from direct feature correspondence alone---nonetheless can be realized through comparison of local properties like line orientations (in the case of vanishing points) and shadows and light sources (in the case of illumination).
Crucially, any pair of points between the image pair can provide evidence for their geometric relationship.

We therefore turn to \emph{correlation volumes}, a tool used in correspondence tasks like optical flow or stereo. In a full correlation volume, every pair of points from feature maps derived from an image pair are compared.
While dense correlation volumes have 
demonstrated superior performance for tasks like optical flow \cite{xu2017accurate,lu2020devon,teed2020raft} and stereo matching  \cite{nie2019multi,liang2019stereo,yang2019hierarchical,gu2020cascade} that compare highly overlapping images, we find that they are also effective in finding implicit cues that are not in the form of direct correspondence. 
As such, 
we process image pairs---whether they overlap or not---by constructing a dense 4D correlation volume (see Fig.~\ref{fig:4d_volume}). 
This design allows us to both find explicit pixelwise correspondence, in the case of overlapping pairs, as well as leverage implicit cues for non-overlapping pairs.

To estimate the relative rotation, we process the correlation volume with another network that computes probabilities estimates over a fine-grained discretization over the space of 3D rotations. Our framework is end-to-end trainable and optimizes simple loss formulations, bypassing difficulties associated with regressing 3D rotations.

We evaluate our method on a large variety of extreme RGB image pairs, including indoor and outdoor images captured in different geographic locations under varying illumination. 
We also show that our models yield state-of-the-art performance for overlapping pairs. Our models generalize surprisingly well to new data---e.g., training a model on outdoor scenes in Manhattan yields median errors below 6$\degree$ for images captured in Pittsburgh and London.

\section{Related Work}

\noindent \textbf{Extreme relative pose estimation.}
Non-overlapping images have been addressed in the context of related problems such as 2D alignment and mosaicing \cite{poleg2012alignment,huang2013mind}, where the goal is to piece together images that have small gaps between them.
Most prior work on the 3D relative pose estimation problem rely on correspondences or overlap, and hence are not well suited for handling extreme cases. Several works address extreme relative pose estimation between two input RGB-D scans~\cite{yang2019extreme, yang2020extreme}. Provided with input depth values, these works perform scan completion and match the completed scans. Caspi and Irani~\cite{caspi2002aligning} consider image sequences where the two cameras are rigidly attached and move jointly, and search for consistent temporal behavior. Littwin \etal~\cite{littwin2015spherical} use inter-silhouette dissimilarities to estimate the relative rotations for a set of cameras. 
In this work, we propose a method for the more challenging problem of extreme relative rotation given only a pair of RGB images as input, without using depth, temporal coherence, or other forms of additional data.

\medskip
\noindent \textbf{Traditional pose estimation.}
The problem of relative pose estimation from overlapping views is traditionally divided into two sequential steps: correspondence estimation using local feature matching~\cite{lowe2004distinctive,bay2006surf}, followed by epipolar geometry--based pose estimation~\cite{hartley2003multiple}. As reliable correspondence estimation can be challenging even for overlapping images, several recent works propose end-to-end convolutional neural networks that regress from images to relative pose~\cite{melekhov2017relative,en2018rpnet,laskar2017camera}. 

In particular, several works focus on 3D rotations and propose parameterizations and architectures that allow for better estimation using neural networks \cite{zhou2019continuity,levinson2020analysis}. 
Zhou \etal~\cite{zhou2019continuity} evaluate commonly used representations, including quaternions and Euler angles, and discuss issues relating to discontinuities in the rotation representation. Peretroukhin \etal \cite{peretroukhin2019deep} regress
and combine multiple rotation estimates, producing  probabilistic estimates. Mohlin \etal~\cite{mohlin2020probabilistic} estimate 3D rotation uncertainty using the matrix Fisher distribution. In our work, we discretize the space of 3D rotations and estimate a distribution over this space to avoid issues associated with direct regression to rotations.

\medskip
\noindent \textbf{Dense volumes for computing image relations.} Dense volumes that encode pairwise pixel comparisons allow for an explicit representation of correspondence, and thus have been used in a number of tasks that require correspondence estimation. For instance, deep stereo matching approaches typically compute cost volumes that compare feature descriptors across disparity ranges \cite{kendall2017end,khamis2018stereonet, chang2018pyramid,yao2018mvsnet,nie2019multi,liang2019stereo,yang2019hierarchical,gu2020cascade}. 
Cost volumes are also used to compute optical flow, in which case 2D displacements are encoded within a volumetric representation \cite{xu2017accurate,sun2018pwc,lu2020devon}. 
To define a matching cost, several works compute dense correlations between image features, forming correlation volumes \cite{dosovitskiy2015flownet,sun2018pwc,teed2020raft}. In our work, we also form correlation volumes. However, unlike prior work, the input images are not necessarily highly correlated and therefore the correlation volumes are tasked with learning additional, more subtle, cues.

Several other recent techniques leverage dense volumes for reconstruction tasks. Zhou \etal~\cite{zhou2020learning} use cost volumes for the task of single-image reconstruction. Wei \etal~\cite{wei2020deepsfm} use them to refine structure-from-motion predictions. As part of their solution, they form a pose-based cost volume that is provided with initial camera pose parameters, then uniformly samples candidate poses around this initial estimate. In our case, we build dense volumes for all pairs of pixels and estimate rotations over the full space of possible rotations.

\section{Method}
Given a pair of RGB images $(I_1,I_2)$, our goal is to estimate the 3D rotation matrix $\mathbf{R}$ between the two images.
In order to allow for discovery and use of hidden cues in the difficult task of extreme rotation prediction, we use dense correlation volumes that allow for discovering implicit cues (Section \ref{sec:correlation}). These dense volumes are fed to a fine-grained relative rotation classification network (Section \ref{sec:classification}). 
An overview of our approach is provided in Fig.~\ref{fig:overview}.

\medskip
\noindent \textbf{Parameterization.} 
We must choose a parameterization for 3D rotations suited to our problem.
One standard representation for a 3D rotation matrix $\mathbf{R}$ is as three Euler angles $[\alpha,\beta,\gamma]$, denoting 
roll, pitch, and yaw angles, respectively:
\begin{equation}
    \mathbf{R}(\alpha, \beta, \gamma) = \mathbf{R}_x(\alpha)\mathbf{R}_y(\beta)\mathbf{R}_z(\gamma)
\end{equation}
This general parameterization can be integrated directly with our presented approach. However, we observe that for a wide variety of scenes the absolute pitch can be recovered from a single image using 
cues such as vanishing points.
Furthermore, cameras are typically upright, \emph{i.e.}, the input images have zero roll.

Given these observations, we can instead represent the relative orientation $\mathbf{R}$ with three angles $[\beta_1,\beta_2, \Delta \gamma]$, where $\Delta \gamma$ denotes the relative yaw angle, $\beta_1$ denotes the pitch of $I_1$ and $\beta_2$ denotes the pitch of $I_2$.
Using this parameterization, the rotation matrices of the two images are defined as $\mathbf{R}_1(0,\beta_1,0)$ and $\mathbf{R}_2(0,\beta_2,\Delta \gamma)$, and the relative rotation matrix is defined as 
$\mathbf{R}=\mathbf{R}_2\mathbf{R}_1^T$. In our experiments we show that encoding this prior knowledge into the parameterization leads to improved performance, although a generic parameterization also shows significant improvement over baseline methods.

\subsection{Dense Correlation Volumes}
\label{sec:correlation}
Our proposed solution is inspired by traditional methods that accumulate evidence for global quantities from local evidence via voting schemes or other mechanisms. 
For instance, as intuition, consider the problem of detecting vanishing points from a single image. One approach is to have local features like line segments vote on vanishing points using accumulation methods similar to Hough transforms~\cite{hough1962method}.
In our case, where we are given an image pair and want to estimate relative rotation, 
we observe that potentially any pair of image patches can provide evidence for the global geometric image relationship---for instance, two patches that support related vanishing point locations, or two patches that give evidence for light source directions.

\begin{figure}
\vspace{-2pt}
\begin{center}
\includegraphics[width=0.95\columnwidth, trim=0 20 0 0, clip]{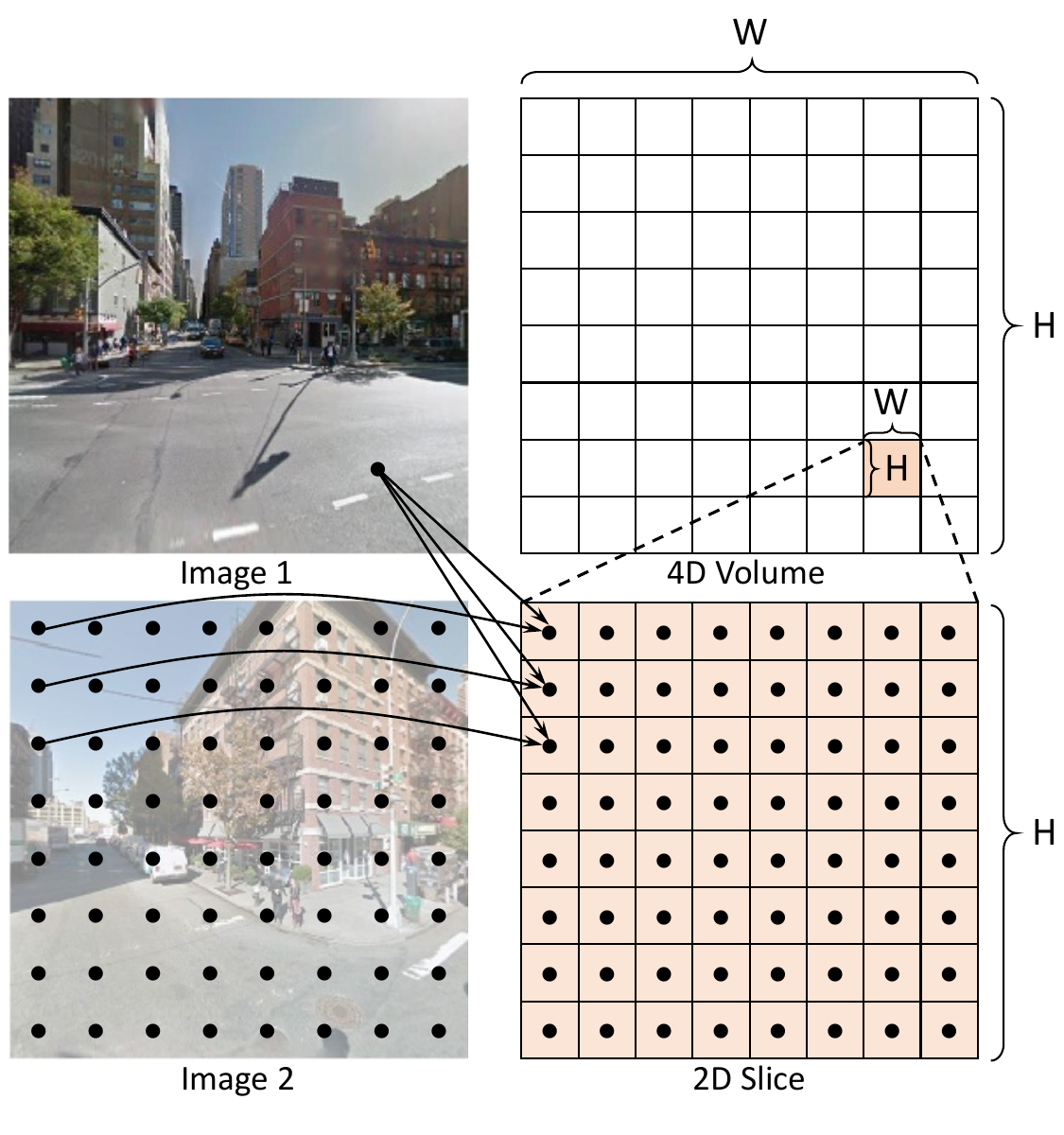}
\end{center}
\vspace{-8pt}
   \caption{\textbf{4D correlation volumes.} 
    A 4D correlation volume is calculated from a pair of image feature maps. Given a feature vector from Image 1, we compute the dot product with all feature vectors in Image 2, and build up a 2D slice of size $H\times W$. Combining all 2D slices over all feature vectors in Image 1, we obtain a 4D correlation volume of size $H\times W\times H\times W$.}
\label{fig:4d_volume}
\end{figure}

To operationalize this intuition, we devise a network structure that performs pairwise comparisons between all pairs of features across the two images.  
We first use convolutional networks
with shared weights to extract dense feature descriptors $\mathbf{f} (I_i)\in \mathbb{R}^{K \times H/4 \times W/ 4 }$, where $K$ is the number of channels, $W$ is the image width and $H$ is the image height. We then compute a 4D dense correlation volume $\mathcal{C}(\mathbf{f} (I_1), \mathbf{f} (I_2))$, such that for each pair of spatial positions $(p,q)$ in $I_1$ and $(r,s)$ in $I_2$, we define the correlation score at $(p, q, s, r)$ as the dot product of the corresponding vectors in the feature maps of the two images: 
\begin{equation}
    \mathcal{C}(\mathbf{f} (I_1), \mathbf{f} (I_2);p,q,r,s) = 
    \mathbf{f} (I_1; p, q) \cdot
    \mathbf{f} (I_2;r,s),
\end{equation}
where $\mathbf{f} (I; x, y)$ denotes the feature vector for image $I$ at spatial position $(x,y)$. 
Fig.~\ref{fig:4d_volume} illustrates the correlation volume for an image pair. 
Note that the correlation volume $\mathcal{C}$ can be computed efficiently using matrix multiplication.

Unlike prior work that uses correlation volumes to directly predict pixel-to-pixel correspondence as in optical flow \cite{xu2017accurate,lu2020devon,teed2020raft} or stereo \cite{nie2019multi,liang2019stereo,gu2020cascade}, our correlation volumes are implicitly assigned a dual role which emerges through training on both overlapping and non-overlapping pairs. When the input image pair contains significant overlap, pointwise correspondence can be computed and transferred onward to the rotation prediction module. When the input image pair contains little to no overlap, the correlation volume can assume the novel role of detecting implicit cues.

\begin{figure} 
    \centering
    \jsubfig{\includegraphics[height=2.06cm]{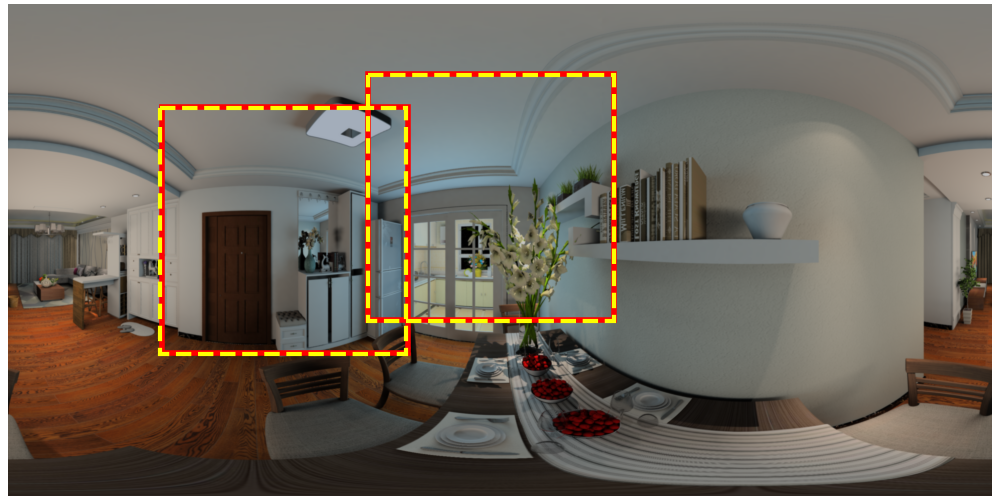}}{}
    \hfill 
    \jsubfig{\includegraphics[height=2.06cm]{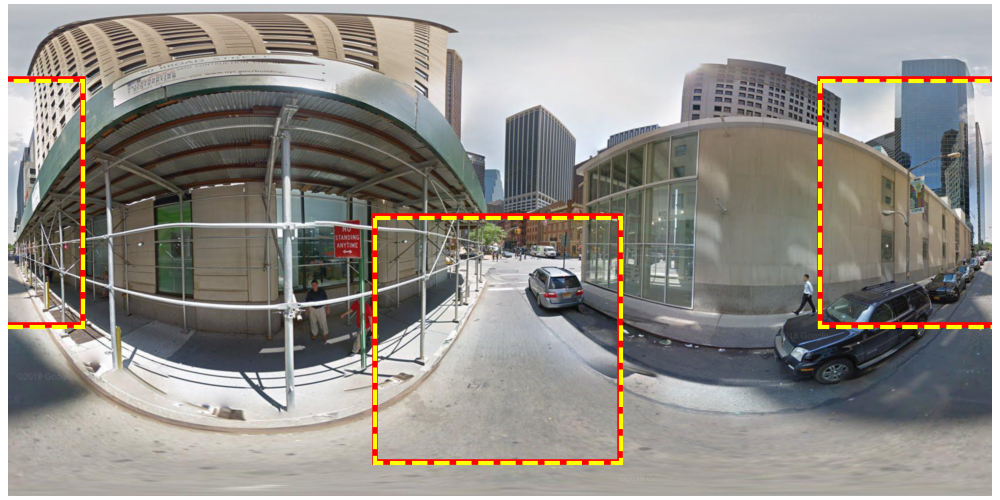}}{}
    \\ 
    \jsubfig{\includegraphics[height=1.67cm]{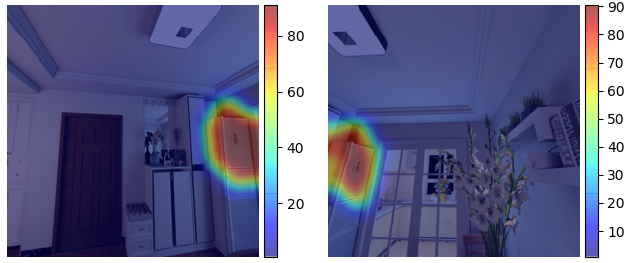}}{} 
    \hfill
    \jsubfig{\includegraphics[height=1.66cm]{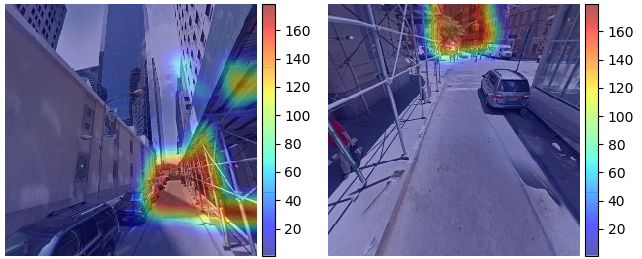}}{} 

    \caption{\textbf{Visualizing cues detected by our model for overlapping (left) and non-overlapping (right) image pairs.} We show regions which, when blocked, affect the rotation error, with warmer colors depicting larger errors (according to their associated color bars). The full panoramas are shown above, with the ground-truth and predicted perspective image regions marked in red and yellow, respectively. 
    These visualizations suggest that our method reasons about pointwise correspondences for overlapping pairs (\emph{e.g.} on top of the refrigerator) and implicit cues for non-overlapping pairs (\emph{e.g.} related to vanishing points).
   }
    \label{fig:clues_method}
\end{figure}

We visualize this dual role in Fig.~\ref{fig:clues_method}, where we create a heatmap for each image in a pair occluding each image region in turn with a sliding window before feeding it to our network, in order to assess each region's approximate importance towards computing the relative pose~\cite{zeiler2014visualizing}. 
As illustrated in the left pair, covering the region of overlap significantly affect the model's prediction for overlapping pairs. For non-overlapping pairs, covering regions corresponding to a strong vanishing point yields a steep drop in performance. Please refer to the supplementary material for additional visualizations.

Our dense correlation volume $\mathcal{C}$ is provided as input to a rotation classification network  $g_\sigma$ 
that is tasked with predicting the relative rotation. 

\subsection{Relative Rotation Classification}
\label{sec:classification}
The relative rotation classification network $g_\sigma$ is constructed from three identical networks (without weight sharing), where each network predicts one of the angles describing the relative rotation according to our three angle parameterization. 
Learning-based pose estimation methods typically regress to rotation and translation parameters \cite{melekhov2017relative,en2018rpnet}. However, commonly used 3D rotation representations, including Euler angles and quaternions, are discontinuous and hence challenging for direct regression in deep networks.
To overcome the problem of regressing discontinuous 3D rotations, prior works have suggested using higher dimensional continuous representations, e.g., in 5D and 6D \cite{zhou2019continuity}. 

As an alternative to directly regressing the relative angles, we discretize the space of rotations, such that for each angle we estimate a probability distribution over $N$ bins. We empirically set $N=360$, and let each bin capture an angle in the range $[-180^\circ,180^\circ]$. This discretization is related to concurrent work that discretizes 3D rotations in terms of rotation matrix columns~\cite{directionnet}.
Our fine-grained discrete angle parameterization enables using a simple cross-entropy loss to train our network. Overall, a sum of three cross-entropy loss functions is used, one per angle. During training, the ground truth is set using one-hot vectors (that empirically yielded similar results as smoothed vectors).

\section{Experiments}
\label{sec:experiment}
To validate our approach, we conduct extensive experiments on a variety of extreme RGB image pairs capturing both indoor and outdoor scenes.
We compare with several baseline techniques, evaluating performance on both overlapping and non-overlapping image pairs. We also demonstrate the generalization power of our model, testing it on image pairs from unseen cities. 
Finally, we present an ablation study to examine the impact of the different components in our proposed approach.
Additional results and implementation details are provided in the supplementary material.

\subsection{Datasets} 
\label{sec:dataset}

\noindent \textbf{StreetLearn}~\cite{mirowski2019streetlearn} is an outdoor dataset that contains approximately 143K panoramic views covering the cities of Manhattan and Pittsburgh. 
We focus on a set of roughly 56K images from Manhattan, randomly allocating 1,000 panoramic views for testing.

\smallskip \noindent \textbf{SUN360}~\cite{xiao2012recognizing} is an indoor dataset that contains 9,962 panoramic views of different scenes downloaded from the Internet, grouped into 50 different categories. We use approximately 7,500 panoramas for training and 830 among the remaining panoramas for testing.

\smallskip \noindent \textbf{InteriorNet}~\cite{InteriorNet18} is a synthetic indoor dataset. We use a subset of InteriorNet that contains 10,050 panoramic views from 112 different houses, where 82 houses are allocated for training and the remaining 30 houses for testing.

\smallskip

For each panoramic view, we randomly sample 100 perspective views to obtain images with a resolution of $256\times 256$ and a 90$\degree$  FoV. We sample images distributed uniformly over the range of $[-180,180]$ for yaw angles. 
This yields, for example, an average of 3147 samples (per $1\degree$ bin) in our StreetLearn dataset. 
We assume zero roll (and demonstrate in the supplementary material that our models are insensitive to small roll angles at test time).
To avoid generating textureless images that focus on the ceiling/sky or the floor, we limit the range over pitch angles to $[-30,30]$ for the indoor datasets and $[-45,45]$ for the outdoor dataset. We note that the StreetLearn dataset contains subtle watermarks at fixed locations. However, as we show in the supplementary material, removing these at test time (using Photoshop's content-aware fill) does not significantly affect the model's performance.
For the InteriorNet dataset, since the dataset is synthetic and panoramas are rendered at random camera positions---leading to cases where the image is too close to the geometry (\emph{e.g.} the scene is largely occluded by a close object or most of the panorama observes a wall), we use MiDaS~\cite{Ranftl2020}, a single view depth estimation method, to filter out images that are too close the scene.

From these perspective images, we construct datasets with and without camera translations, in order to understand the effect of camera translation on our rotation estimation problem. Datasets without translations are constructed by sampling image pairs originating from the same panorama, and datasets with translations, denoted as InteriorNet-T and StreetLearn-T, are constructed by pairing up images from different panoramas using translation distances smaller than 3m and 10m, respectively (SUN360 is constructed from Internet panoramas that are not physically related in space).
For all datasets, we have $\sim$1M training pairs sampled from the same panorama, $\sim$700K training pairs sampled from different panoramas, and 1K test pairs.
There is no overlap between train and test scenes.

\definecolor{graytext}{RGB}{130,130,130}

\begin{table*}[t]
\setlength{\tabcolsep}{1.0pt}
 \def\arraystretch{0.95}
\centering
\resizebox{\textwidth}{!}{%
\begin{tabular}{llccclccclccclccclccc}
\toprule
\multicolumn{2}{l}{}& \multicolumn{3}{c}{InteriorNet} &  & \multicolumn{3}{c}{InteriorNet-T} &  & \multicolumn{3}{c}{SUN360} &  & \multicolumn{3}{c}{StreetLearn} &  & \multicolumn{3}{c}{StreetLearn-T}\\ \cline{3-5} \cline{7-9} \cline{11-13} \cline{15-17} \cline{19-21}
Overlap 
& Method
& Avg(\degree$\downarrow$) & Med(\degree$\downarrow$) & \multicolumn{1}{c}{10\degree(\%$\uparrow$)} & & Avg(\degree$\downarrow$) & Med(\degree$\downarrow$) & \multicolumn{1}{c}{10\degree(\%$\uparrow$)} &  & Avg(\degree$\downarrow$) & Med(\degree$\downarrow$) & \multicolumn{1}{c}{10\degree(\%$\uparrow$)} &  & Avg(\degree$\downarrow$) & Med(\degree$\downarrow$) & \multicolumn{1}{c}{10\degree(\%$\uparrow$)} &  & Avg(\degree$\downarrow$) & Med(\degree$\downarrow$) & \multicolumn{1}{c}{10\degree(\%$\uparrow$)} \\ \midrule
\multirow{6}{*}{Large} & SIFT*~\cite{lowe2004distinctive}       & 6.09                 & 4.00                 & 84.86               &  & 7.78                 & 2.95                 & 55.52               &  & 5.46                       & 3.88                       & 93.10                      &  & 5.84                 & 3.16                 & 91.18               &  & \color{graytext}{18.86}                & \color{graytext}{3.13}                 & \color{graytext}{22.37}               \\
                         & SuperPoint*~\cite{detone18superpoint}  & 5.40                 & 3.53                 & 87.10               &  & 5.46                 & 2.79                 & 65.97               &  & 4.69                       & 3.18                       & 92.12                      &  & 6.23                 & 3.61                 & 91.18               &  & \color{graytext}{6.38}        & \color{graytext}{1.79}        & \color{graytext}{16.45}               \\
                         & Reg6D~\cite{zhou2019continuity}-o     & 5.43                 & 3.87                 & 87.10               &  & 10.45                & 6.91                 & 67.76               &  & 7.18                       & 5.79                       & 81.28                      &  & 3.36                 & 2.71                 & 97.65               &  & 12.31                & 6.02                 & 69.08               \\
                         & Reg6D~\cite{zhou2019continuity}        & 9.05                 & 5.90                 & 68.49               &  & 17.00                & 11.95                & 41.79               &  & 16.51                      & 12.43                      & 40.39                      &  & 11.70                & 8.87                 & 58.24               &  & 36.71                & 24.79                & 23.03               \\
                         & Ours-o                                & \textbf{1.53}        & 1.10                 & \textbf{99.26}     &  & \textbf{2.89}        & \textbf{1.10}        & \textbf{97.61}     &  & \textbf{1.00}              & \textbf{0.94}              & \textbf{100.00}           &  & \textbf{1.19}        & \textbf{1.02}        & \textbf{99.41}     &  & \textbf{9.12}                 & 2.91                 & \textbf{87.50}     \\
                         & Ours                                   & 1.82                 & \textbf{0.88}        & 98.76               &  & 8.86                 & 1.86                 & 93.13               &  & 1.37                       & 1.09                       & 99.51                      &  & 1.52                 & 1.09                 & \textbf{99.41}     &  & 24.98                & \textbf{2.48}                 & 78.95               \\ \midrule
\multirow{6}{*}{Small} & SIFT*~\cite{lowe2004distinctive}       & 24.18                & 8.57                 & 39.73               &  & \color{graytext}{18.16}                & \color{graytext}{10.01}                & \color{graytext}{18.52}               &  & 13.71                      & 6.33                       & 56.77                      &  & 16.22                & 7.35                 & 55.81               &  & \color{graytext}{38.78}                & \color{graytext}{13.81}                & \color{graytext}{5.68}                \\
                         & SuperPoint*~\cite{detone18superpoint}  & \color{graytext}{16.72}                & \color{graytext}{8.43}                 & \color{graytext}{21.58}               &  & \color{graytext}{11.61}                & \color{graytext}{5.82}                 & \color{graytext}{11.73}               &  & \color{graytext}{17.63}                      & \color{graytext}{7.70}                       & \color{graytext}{26.69}                      &  & \color{graytext}{19.29}                & \color{graytext}{7.60}                 & \color{graytext}{24.58}               &  & \color{graytext}{6.80}        & \color{graytext}{6.85}                 & \color{graytext}{0.95}                \\
                         & Reg6D~\cite{zhou2019continuity}-o     & 17.83                & 9.61                 & 51.37               &  & 21.87                & 11.43                & 44.14               &  & 18.61                      & 11.66                      & 39.85                      &  & 7.95                 & 4.34                 & 87.71               &  & 15.07                & 7.59                 & 63.41               \\
                         & Reg6D~\cite{zhou2019continuity}        & 25.71                & 15.56                & 33.56               &  & 42.93                & 28.92                & 23.15               &  & 42.55                      & 32.11                      & 9.40                       &  & 24.77                & 15.11                & 30.56               &  & 46.61                & 34.33                & 13.88               \\
                         & Ours-o                                & 6.45                 & 1.61                 & 95.89               &  & \textbf{10.24}       & \textbf{1.38}        & \textbf{89.81}     &  & \textbf{3.09}              & \textbf{1.41}              & \textbf{98.50}            &  & \textbf{2.32}        & \textbf{1.41}        & \textbf{98.67}     &  & \textbf{13.04}                & 3.49                 & \textbf{84.23}     \\
                         & Ours                                   & \textbf{4.31}        & \textbf{1.16}        & \textbf{96.58}     &  & 30.43                & 2.63                 & 74.07               &  & 6.13                       & 1.77                       & 95.86                      &  & 3.23                 & \textbf{1.41}        & 98.34               &  & 27.84                & \textbf{3.19}        & 74.76               \\ \midrule
\multirow{5}{*}{None}   & SIFT*~\cite{lowe2004distinctive}       & \color{graytext}{109.30}               & \color{graytext}{92.86}                & \color{graytext}{0.00}                &  & \color{graytext}{93.79}                & \color{graytext}{113.86}               & \color{graytext}{0.00}                &  & \color{graytext}{127.61}                     & \color{graytext}{129.07}                     & \color{graytext}{0.00}                       &  & \color{graytext}{83.49}                & \color{graytext}{90.00}                & \color{graytext}{0.38}                &  & \color{graytext}{85.90}                & \color{graytext}{106.84}               & \color{graytext}{0.38}                \\
                         & SuperPoint*~\cite{detone18superpoint}  & \color{graytext}{120.28}               & \color{graytext}{120.28}               & \color{graytext}{0.00}                &  & \color{graytext}{--}                   & \color{graytext}{--}                   & \color{graytext}{0.00}                &  & \color{graytext}{149.80}                     & \color{graytext}{165.24}                     & \color{graytext}{0.00}                       &  & \color{graytext}{--}                   & \color{graytext}{--}                   & \color{graytext}{0.00}                &  & \color{graytext}{--}                   & \color{graytext}{--}                   & \color{graytext}{0.00}                \\
                          & Reg6D~\cite{zhou2019continuity}-o  & 88.89                            & 79.24                            & 0.98                         &  & 110.50                           & 116.25                           & 1.17                         &  & 101.21                           & 99.64                            & 0.94                         &  & 133.08       & 167.19       & 2.46     &  & 132.67       & 158.45       & 1.88     \\
                         & Reg6D~\cite{zhou2019continuity}        & 48.36                & 32.93                & 10.82               &  & 60.91                & 51.26                & 11.14               &  & 64.74                      & 56.55                      & 3.77                       &  & 28.48                & 18.86                & 24.39               &  & 49.23                & 35.66                & 11.86               \\
                         & Ours                                   & \textbf{37.69}       & \textbf{3.15}        & \textbf{61.97}     &  & \textbf{49.44}       & \textbf{4.17}        & \textbf{58.36}     &  & \textbf{34.92}             & \textbf{4.43}              & \textbf{61.39}            &  & \textbf{5.77}        & \textbf{1.53}        & \textbf{96.41}     &  & \textbf{30.98}       & \textbf{3.50}        & \textbf{72.69}     \\ \midrule
\multirow{5}{*}{All}    & SIFT*~\cite{lowe2004distinctive}       & 13.68                & 5.04                 & 45.80               &  & \color{graytext}{12.24}                & \color{graytext}{5.69}                 & \color{graytext}{24.60}               &  & \color{graytext}{18.12}                      & \color{graytext}{5.02}                       & \color{graytext}{34.00}                      &  & \color{graytext}{17.29}                & \color{graytext}{5.53}                 & \color{graytext}{32.50}               &  & \color{graytext}{36.00}                & \color{graytext}{6.03}                 & \color{graytext}{5.40}                \\
                         & SuperPoint*~\cite{detone18superpoint}  & \color{graytext}{8.19}        & \color{graytext}{4.08}                 & \color{graytext}{41.40}               &  & \color{graytext}{6.62}        & \color{graytext}{3.38}                 & \color{graytext}{25.90}               &  & \color{graytext}{11.09}             & \color{graytext}{4.00}                       & \color{graytext}{25.80}                      &  & \color{graytext}{11.52}                & \color{graytext}{4.80}                 & \color{graytext}{22.90}               &  & \color{graytext}{6.42}        & \color{graytext}{2.62}        & \color{graytext}{2.80}                \\
                          & Reg6D~\cite{zhou2019continuity}-o     & 34.51                            & 9.71                             & 50.40                        &  & 48.27                            & 15.59                            & 37.40                        &  & 60.15                            & 34.51                            & 27.60                        &  & 73.36        & 20.03        & 44.30    &  & 77.09        & 55.29        & 31.60\\ 
                         & Reg6D~\cite{zhou2019continuity}        & 25.90                & 13.02                & 40.70               &  & 40.38                & 23.35                & 25.30               &  & 49.05                      & 34.37                      & 12.70                      &  & 24.51                & 15.31                & 32.00               &  & 46.50                & 33.14                & 29.90               \\
                         & Ours                                   & \textbf{13.49}                & \textbf{1.18}        & \textbf{86.90}     &  & \textbf{29.68}                & \textbf{2.58}        & \textbf{75.10}     &  & \textbf{20.45}                      & \textbf{2.23}              & \textbf{78.30}            &  & \textbf{4.40}        & \textbf{1.44}        & \textbf{97.50}     &  & \textbf{29.85}                & \textbf{3.20}                 & \textbf{74.30}     \\ 
                        \bottomrule 
\end{tabular}
}
\vspace{-8pt}
\caption{\textbf{Rotation estimation evaluation on the InteriorNet, the SUN360, and the StreetLearn datasets.} 
We report the mean and median geodesic error in degrees, and the percentage of pairs with a relative rotation error under 10$\degree$ for different levels of overlap, as detailed in Section \ref{sec:metrics}. 
Models trained only on overlapping pairs are denoted with ``-o''.
* indicates mean/median errors are computed only over successful image pairs, for which these algorithms output a pose estimate (failure over more than $50\%$ of the test pairs is shown in gray).}
\label{tab:main_result}
\end{table*}

\subsection{Baselines}
\label{sec:baseline}

\noindent \textbf{SIFT-based relative rotation estimation}. A geometry-based technique that computes SIFT~\cite{lowe2004distinctive} features and then estimates a rotation matrix with a 2-point algorithm~\cite{brown2007minimal} for image pairs from the same panorama, or an essential matrix for image pairs with translation, using RANSAC~\cite{fischler1981random}.

\smallskip
\noindent \textbf{Learning-based feature matching}, using pretrained networks for interest point detection and description (with model fitting as described above). We evaluate a pretrained SuperPointNet~\cite{detone18superpoint} (hereby called SuperPoint) and D2-Net~\cite{Dusmanu2019CVPR} (see supplemental material for results).

\smallskip
\noindent \textbf{End-to-end relative rotation regression}, where image features are concatenated and fed to a regression model. 
We evaluate models predicting a continuous representation in 6D (hereby denoted as Reg6D) as proposed by Zhou \etal~\cite{zhou2019continuity} and additional representations, including quaternions as proposed by En \etal~\cite{en2018rpnet}, in the supplementary material.

\smallskip
\noindent For all end-to-end techniques, we also train models with \emph{only} overlapping pairs to better understand the impact of training models with non-overlapping image pairs.

\subsection{Evaluation Metrics} 
\label{sec:metrics}
For an image pair, let $\mathbf{R}$ be the predicted rotation matrix and $\mathbf{R}^*$ be ground truth rotation matrix. We follow prior work and report the geodesic error $\arccos\left(\frac{tr(\mathbf{R}^T\mathbf{R}^*)-1}{2}\right)$.

To analyze the performance of methods across different overlap ratios, we divide the test image pairs into three categories: (i) \textbf{large}, indicating highly overlapping pairs (relative rotations up to 45$\degree$), (ii) \textbf{small}, indicating pairs that partially overlap (relative rotation angles $\in [45\degree,90\degree]$), and (i) \textbf{none}, indicating pairs with no overlap (relative rotations $>$90$\degree$).

\subsection{Quantitative Evaluation}
\label{sec:rotation_result}

Table \ref{tab:main_result} reports the mean and median geodesic error, as well as the percentage of image pairs with a relative rotation error under $10\degree$. 
Qualitative results are presented in Figure \ref{fig:predicted}. 
We analyze the results according to the amount of overlap: 

\smallskip
\noindent \textbf{Overlapping cases.}
For image pairs sampled from the same panorama, our model produces very accurate results for both indoor and outdoor scenes, with mean errors of 4.31$\degree$, 6.13$\degree$, and 3.23$\degree$ for small overlapping pairs.
For the regression baselines, adding non-overlapping pairs during training causes the regression baseline performance to suffer on overlapping pairs, while our method does not see such a drop.

The performance of models trained on datasets with camera translations is somewhat lower. For these, the overlapping regions can be smaller due to the camera motion. In particular, for StreetLearn-T, the translations are large (up to 10m), which can have a more dramatic effect on the overlap region. 
Nonetheless, our method still achieves low median errors for datasets with translation (around 3$\degree$).

\smallskip
SuperPoint~\cite{detone18superpoint} achieves the smallest average errors on StreetLearn-T large and small overlapping cases ($6.38\degree$ and $6.80\degree$). 
However, both SIFT and SuperPoint do not always output an answer (model fitting-based techniques require a sufficient number of detected inliers). Only successful image pairs are considered for evaluation, and hence these errors should be interpreted as errors over pairs for which they produced an answer (full numbers provided in the supplemental material).
Our method still significantly outperforms SuperPoint in terms of the fraction of pairs with less than $10^\circ$ error across all datasets, 
suggesting that 
SuperPoint fails to give an answer.

\smallskip
\noindent \textbf{Non-overlapping cases.}
Due to insufficient correspondences on image pairs with large viewpoint changes, feature matching-based methods (and regression models trained on only overlapping pairs) unsurprisingly fail on pairs with no overlap.
The median errors of the strongest regression baseline are all above $18\degree$ (and for several datasets are much larger).
In contrast, our method yields median errors consistently below 5$\degree$, while the mean error rises to 6$\degree$--49$\degree$. 
This indicates that our method is usually surprisingly accurate, but sometimes makes large errors. It turns out that such errors are primarily due to ambiguities, as detailed below.

To better understand cases when our method produces incorrect rotations, we can look at the full distributions predicted by our method. Because our network outputs a probability distribution over 3D rotations, we can compute a top-2 error (i.e., the smaller of the errors over both the most likely and second most likely prediction), to see whether the correct rotation angles have high peaks in the learned probability distributions. 
On the StreetLearn-T dataset, the mean error significantly drops, from $24.98\degree$ to $6.49\degree$ on large overlapping cases, $27.84\degree$ to $10.39\degree$ on small overlapping cases, and $30.98\degree$ to $15.72\degree$ on non-overlapping cases.
This gap illustrates that the model is able to predict with large probability the ground truth angle, but is confused with other likely rotation estimates. For instance, given two images facing two different roads at the crossing, there is an ambiguity between $90\degree$ and $180\degree$ rotations. We provide full distributions of prediction errors (that demonstrate, for example, that there are error modes at $90\degree$ and $180\degree$ rotations) and qualitative results of failure cases in the supplemental material.

\begin{figure} 
    \centering
\vspace{-7pt}
    \newcommand{\tal}{1.69cm}
    \newcommand{\tab}{0cm}
    \newcommand{\tar}{1.3cm}
    \newcommand{\tat}{1cm}
    \jsubfig{\includegraphics[height=3.8cm,trim={0cm} {\tab} {\tar} {\tat},clip]{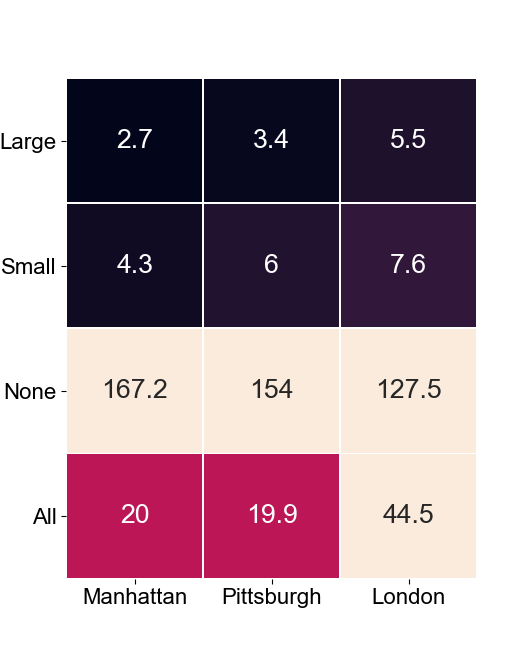}} {\vspace{-9pt}\quad Reg6D-o}
    \hfill
    \jsubfig{\includegraphics[height=3.8cm,trim={\tal} {\tab} {\tar} {\tat},clip]{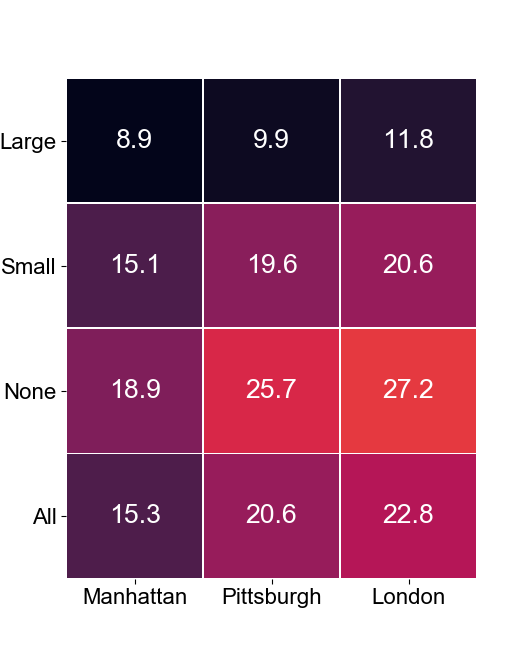}} {\vspace{-9pt}Reg6D}
    \hfill
    \jsubfig{\includegraphics[height=3.8cm,trim={\tal} {\tab} {\tar} {\tat},clip]{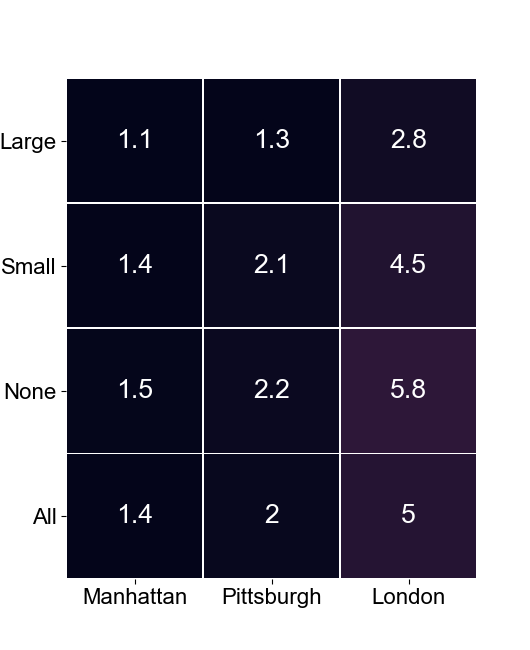}} {\vspace{-9pt}Ours}
\vspace{-12pt}
    \caption{\textbf{Generalization to new locations.} We show the median geodesic rotation error over test images from Manhattan, Pittsburgh, and London for the models trained on Manhattan. Results are reported for our model and the regression models proposed by Zhou \etal~\cite{zhou2019continuity} (Reg6D-o is trained only on overlapping pairs).  
   }
    \label{fig:generalization}
\end{figure}
\begin{table}[]
\centering
\scriptsize
\setlength{\tabcolsep}{1.0pt}
\resizebox{\columnwidth}{!}{
\begin{tabular}{llccclcclcc} 
\toprule
\multicolumn{2}{l}{}                         & \multicolumn{3}{c}{Rotation Error}&  & \multicolumn{2}{c}{Yaw Error}                                                                       &  & \multicolumn{2}{c}{Pitch Error}\\  \cline{3-5} \cline{7-8} \cline{10-11}
Overlap & Method
& Avg(\degree) & Med(\degree) & \multicolumn{1}{c}{10\degree(\%)} & & Avg(\degree) & Med(\degree) & & Avg(\degree) & Med(\degree)
\\ \midrule
\multirow{5}{*}{Large} & w/o RP                              & 1.02         & 1.05         & 100.00 &  & 0.48          & 0.47          &  & 0.52           & 0.48           \\
                       & w/o $\mathcal{L}_\mathsf{cls},\mathcal{C}$ & 40.19        & 33.41        & 8.87   &  & 33.22         & 23.19         &  & 16.50          & 14.01          \\
                       & w/o $\mathcal{L}_\mathsf{cls}$             & 7.93         & 6.96         & 78.33  &  & 5.07          & 3.90          &  & 3.22           & 2.91           \\
                       & w/o $\mathcal{C}$                   & 43.79        & 43.44        & 6.90   &  & 43.74         & 43.44         &  & 1.07           & 0.74           \\
                       & Ours (full)                         & 1.37         & 1.09         & 99.51  &  & 0.57          & 0.51          &  & 1.03           & 0.73           \\ \midrule
\multirow{5}{*}{Small} & w/o RP                              & 1.67         & 1.44         & 99.62  &  & 0.83          & 0.61          &  & 0.79           & 0.64           \\
                       & w/o $\mathcal{L}_\mathsf{cls},\mathcal{C}$ & 67.97        & 60.91        & 0.38   &  & 61.53         & 54.19         &  & 19.71          & 15.59          \\
                       & w/o $\mathcal{L}_\mathsf{cls}$             & 15.75        & 12.65        & 34.59  &  & 10.28         & 6.35          &  & 5.81           & 4.31           \\
                       & w/o $\mathcal{C}$                   & 65.31        & 69.32        & 15.79  &  & 65.13         & 69.31         &  & 2.04           & 1.10           \\
                       & Ours (full)                         & 6.13         & 1.77         & 95.86  &  & 4.70          & 0.72          &  & 1.76           & 0.97           \\ \midrule
\multirow{5}{*}{None}  & w/o RP                              & 39.01        & 25.81        & 23.92  &  & 36.32         & 21.17         &  & 5.96           & 3.39           \\
                       & w/o $\mathcal{L}_\mathsf{cls},\mathcal{C}$ & 83.29        & 75.08        & 0.56   &  & 78.42         & 68.89         &  & 18.16          & 14.72          \\
                       & w/o $\mathcal{L}_\mathsf{cls}$             & 46.26        & 39.53        & 2.64   &  & 40.98         & 34.08         &  & 13.93          & 11.08          \\
                       & w/o $\mathcal{C}$                   & 121.16       & 130.91       & 0.19   &  & 121.10        & 130.91        &  & 1.97           & 1.26           \\
                       & Ours (full)                         & 34.92        & 4.43         & 61.39  &  & 34.19         & 3.51          &  & 1.91           & 1.36           \\ \midrule
\multirow{5}{*}{All}   & w/o RP                              & 21.37        & 3.98         & 59.50  &  & 19.60         & 1.87          &  & 3.48           & 1.09           \\
                       & w/o $\mathcal{L}_\mathsf{cls},\mathcal{C}$ & 70.46        & 57.87        & 2.20   &  & 64.75         & 52.70         &  & 18.24          & 14.72          \\
                       & w/o $\mathcal{L}_\mathsf{cls}$             & 30.36        & 19.97        & 26.50  &  & 25.52         & 12.61         &  & 9.60           & 5.88           \\
                       & w/o $\mathcal{C}$                   & 90.60        & 89.37        & 5.70   &  & 90.51         & 89.35         &  & 1.81           & 1.10           \\
                       & Ours (full)                         & 20.45        & 2.23         & 78.30  &  & 19.52         & 1.07          &  & 1.69           & 1.07          
                       \\ \bottomrule              
\end{tabular}%
}
\vspace{-8pt}
\caption{\textbf{Ablation study},
evaluating the effect of our rotation parameterization (RP), classification objective ($\mathcal{L}_\mathsf{cls}$) and correlation volumes ($\mathcal{C}$) on SUN360.
}
\label{tab:ablation}
\end{table}

\subsection{Generalization to Other Datasets} 
\label{sec:generalization}
To demonstrate the generalization power of our model, we test the model trained on StreetLearn on outdoor images from Pittsburgh and London, using images from the Holicity dataset~\cite{zhou2020holicity}, and compare to the regression model of Zhou \etal~\cite{zhou2019continuity}. Test images are split according to the overlap levels detailed in Section \ref{sec:metrics}. Figure \ref{fig:new_cities} illustrates several qualitative examples, also demonstrating how different these samples are (\emph{e.g.} captured in rural areas or on water). 

Results are reported in Figure \ref{fig:generalization}. 
Our StreetLearn-trained model generalizes well to the other outdoor datasets, with all median errors below $6\degree$. In comparison, the median errors for Reg6D on small or non-overlapping pairs are larger on the other cities by 5-10$\degree$, suggesting that this baseline is more sensitive to image appearance. 
For overlapping pairs, we can see that Reg6D-o obtains only slightly larger median errors on the other cities (but still larger than ours), while it completely fails on non-overlapping pairs.

This experiment, and also the strong performance obtained on SUN360, a dataset constructed from Internet panoramas originating from various sources, suggest that we are not simply learning spurious correlations (\emph{e.g.} learning pose based on panorama stitching artifacts), as our models can successfully predict relative rotations for pairs that are significantly different from those they were trained on.

\subsection{Ablation Study}
\label{sec:ablation}
We perform an ablation study to analyze the effect of individual algorithmic components (see Table \ref{tab:ablation}). In particular, we replace (i) the rotation parameterization, (ii) dense correlation volumes with concatenated image features, and (iii) the relative rotation classification objective with a regression loss, regressing to a continuous representation in 6D.

For (i), instead of predicting the absolute pitch and the relative yaw, 
we predict the roll, pitch and yaw $[\alpha, \beta, \gamma]$ (directly decomposed from the relative rotation matrix).
The evaluation suggests that our parameterization generally does not have a large effect on the final result. However, decomposing the absolute pitch per image is helpful to accurately predict the pitch angle in non-overlapping cases.

Dense correlation volumes yield significant improvements, particularly over relative yaw predictions, and boost performance over regression models, indicating that networks with correlation volumes can better reason over geometric relations between images with little to no overlap.

The impact of the classification loss is visible for both overlapping and non-overlapping cases, especially for pitch predictions. Using a regression loss instead of a classification loss also degrades 
yaw accuracy for overlapping cases.
The significant improvements in the median error (\emph{e.g.} $39.53\degree$ to $4.43\degree$ for non-overlapping pairs) illustrates that predicting a distribution over discretized angles enables the model to learn richer information in comparison to a regression model.

\begin{figure*}
\begin{minipage}{\linewidth}
    \begin{center}
    \newcommand{\sizea}{0.235\textwidth}
    \newcommand{\tal}{0cm}
    \newcommand{\tab}{0cm}
    \newcommand{\tar}{0cm}
    \newcommand{\tat}{0cm}
    \newcommand{\smb}{0cm}
    \newcommand{\T}[1]{\raisebox{-0.5\height}{#1}}
    \setlength{\tabcolsep}{0pt}
    \renewcommand{\arraystretch}{0}
    \begin{tabular}{@{}ccccc@{}}
        \rotatebox[origin=c]{90}{Large} &
        \T{\includegraphics[width=\sizea, trim={\tal} {\tab} {\tar} {\tat},clip]{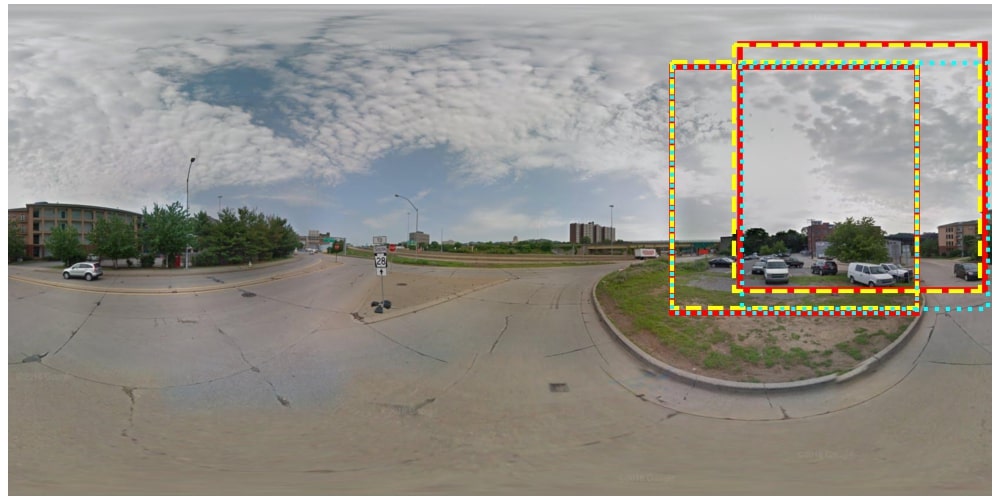}} &
        \T{\includegraphics[width=\sizea, trim={\tal} {\tab} {\tar} {\tat},clip]{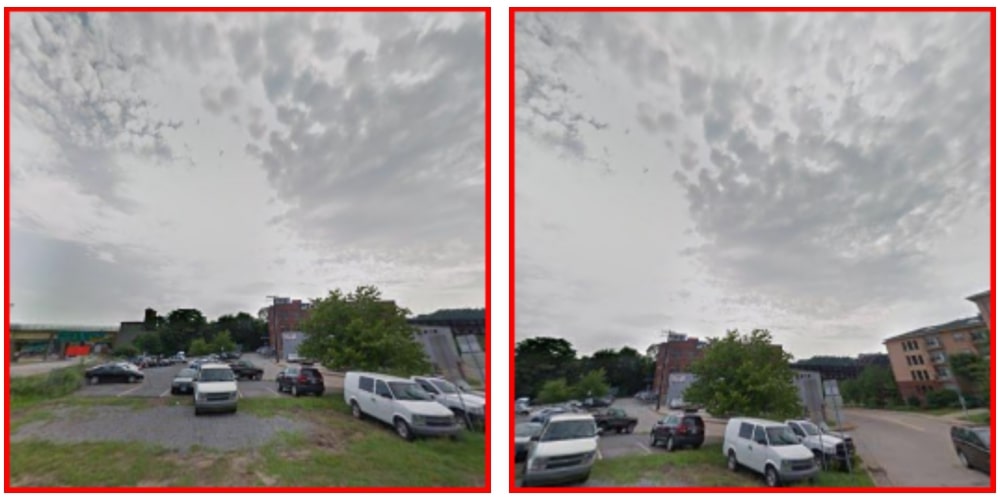}} &
        \T{\includegraphics[width=\sizea, trim={\tal} {\tab} {\tar} {\tat},clip]{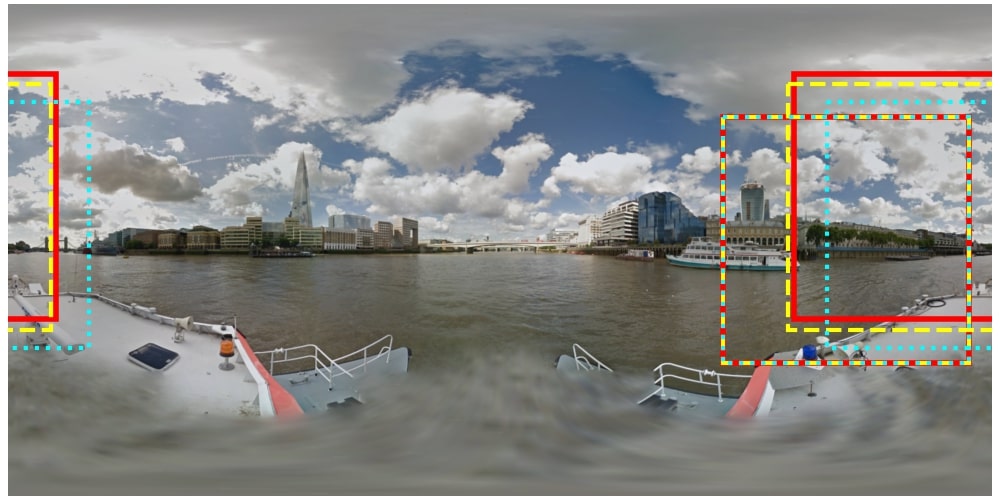}}&
        \T{\includegraphics[width=\sizea, trim={\tal} {\tab} {\tar} {\tat},clip]{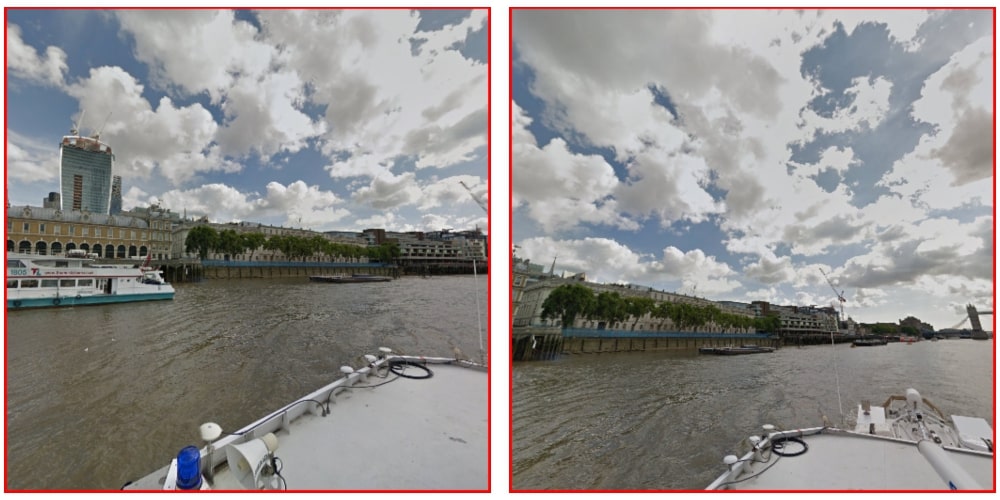}}
        \\ 
       
       \rotatebox[origin=c]{90}{Small} &
       \T{\includegraphics[width=\sizea, trim={\tal} {\tab} {\tar} {\tat},clip]{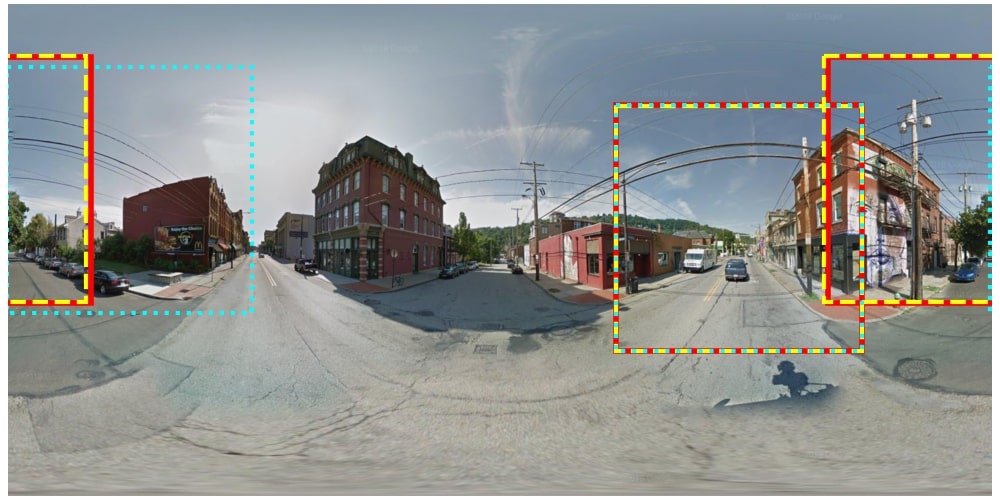}} &
        \T{\includegraphics[width=\sizea, trim={\tal} {\tab} {\tar} {\tat},clip]{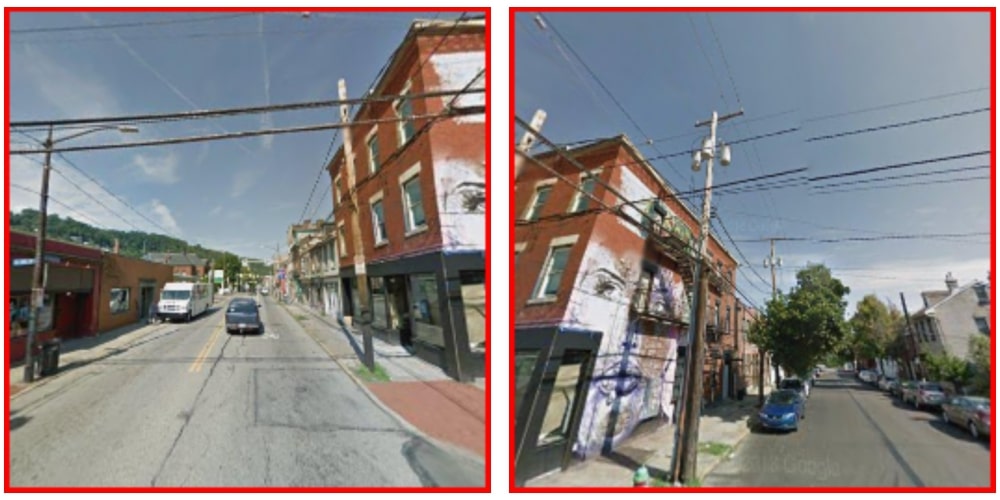}} &
        \T{\includegraphics[width=\sizea, trim={\tal} {\tab} {\tar} {\tat},clip]{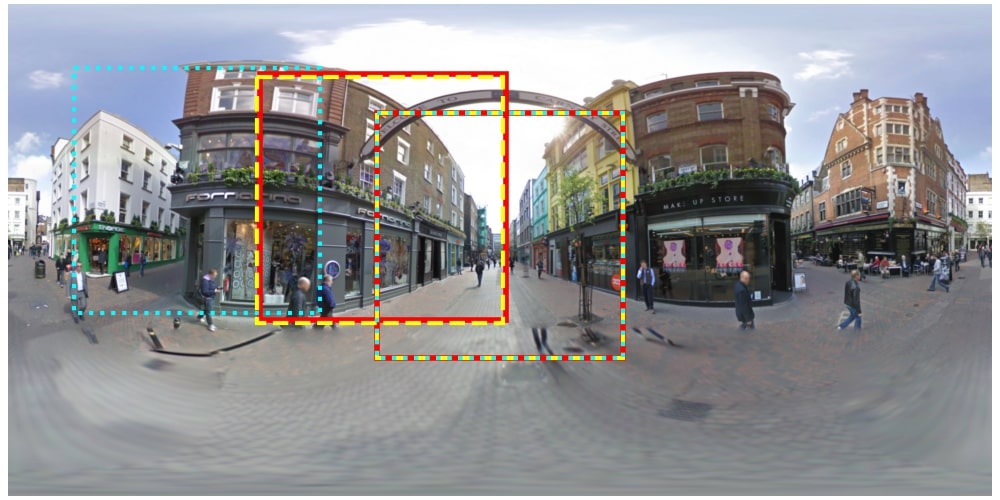}}&
        \T{\includegraphics[width=\sizea, trim={\tal} {\tab} {\tar} {\tat},clip]{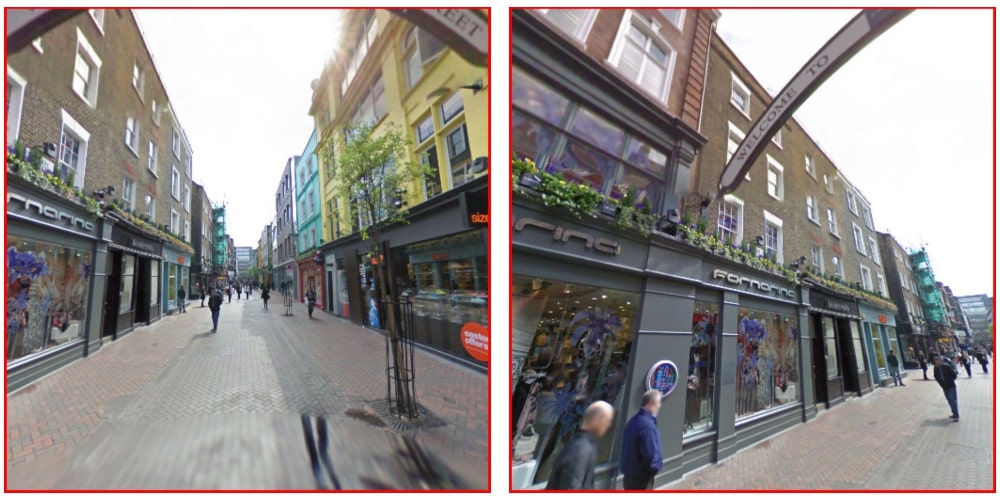}}
        \\ \vspace{+2pt}
       
        \rotatebox[origin=c]{90}{None} &
        \T{\includegraphics[width=\sizea, trim={\tal} {\tab} {\tar} {\tat},clip]{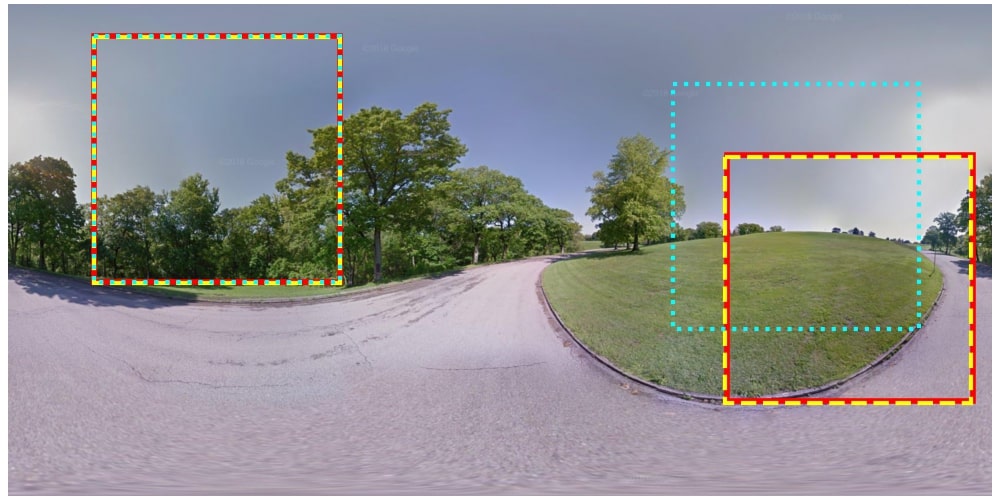}} &
        \T{\includegraphics[width=\sizea, trim={\tal} {\tab} {\tar} {\tat},clip]{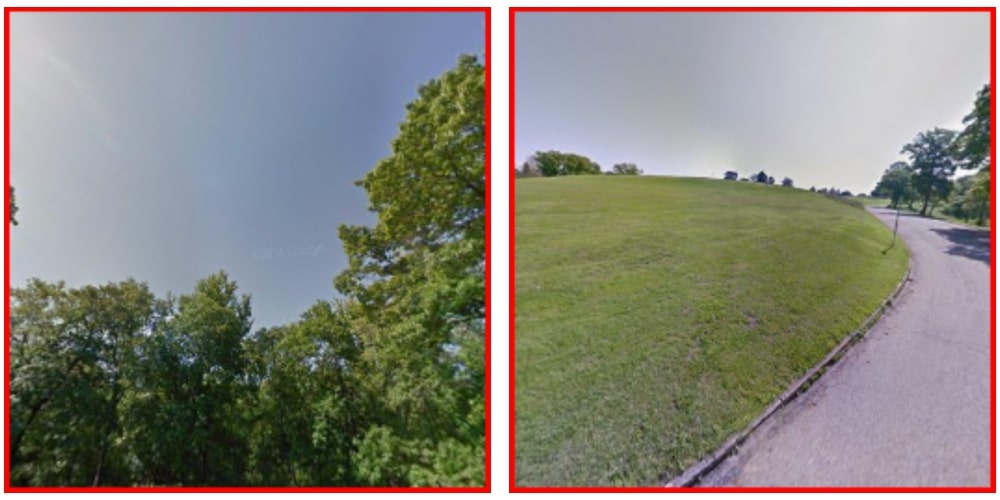}} &
        \T{\includegraphics[width=\sizea, trim={\tal} {\tab} {\tar} {\tat},clip]{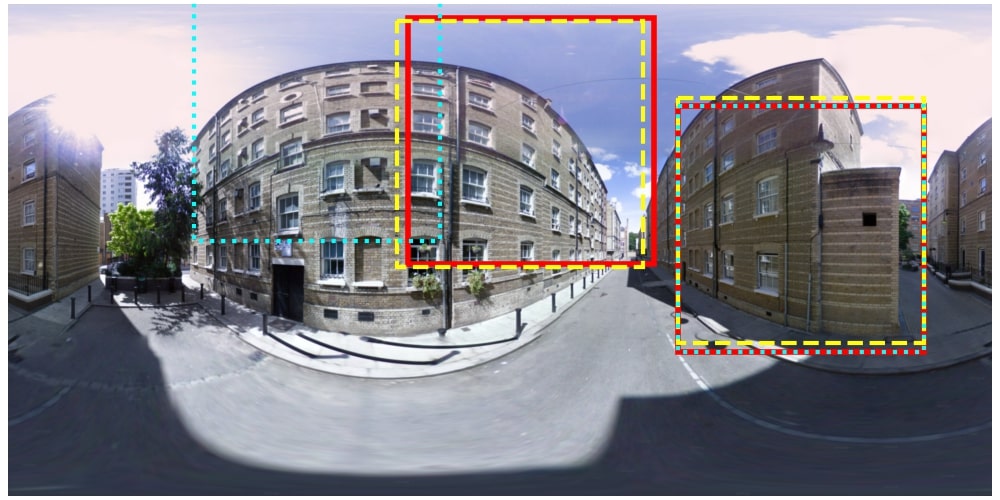}}&
        \T{\includegraphics[width=\sizea, trim={\tal} {\tab} {\tar} {\tat},clip]{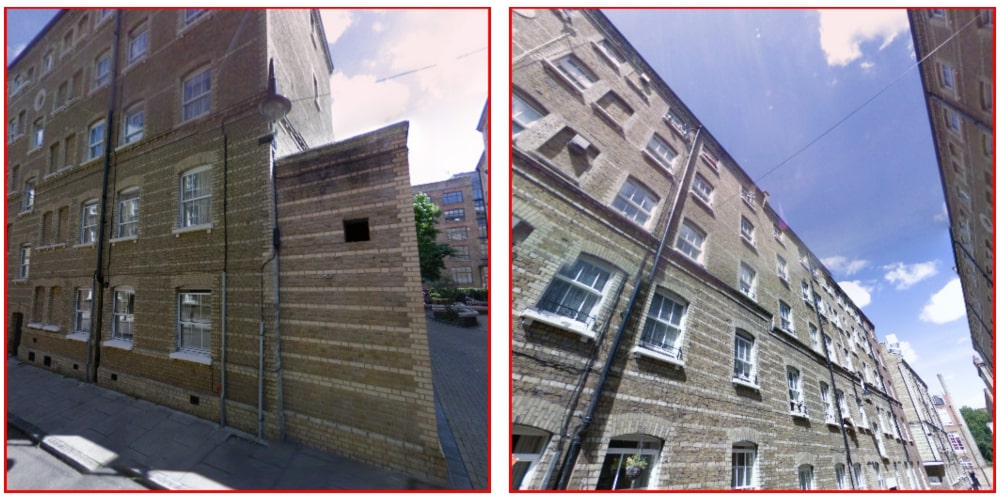}}
        \\ 
        & \multicolumn{2}{c}{Pittsburgh test images} 
        & \multicolumn{2}{c}{London test images} 
    \end{tabular}

    \end{center}
    \vspace{-15pt}
    \caption{\textbf{Generalization from Manhattan to Pittsburgh and London.} Above we show results obtained by our StreetLearn model on image pairs from new, never-before-seen cities. 
    The full panoramas are shown on the left with the ground-truth perspective images marked in red. 
    We show our predicted viewpoints (in yellow) and the result by the Reg6D regression model~\cite{zhou2019continuity} (in blue).
    }
    \label{fig:new_cities}
\end{minipage}

\begin{minipage}{\linewidth}
    \begin{center}
    \newcommand{\sizea}{0.235\textwidth}
    \newcommand{\tal}{0cm}
    \newcommand{\tab}{0cm}
    \newcommand{\tar}{0cm}
    \newcommand{\tat}{0cm}
    \newcommand{\smb}{0cm}
    \newcommand{\T}[1]{\raisebox{-0.5\height}{#1}}
    \setlength{\tabcolsep}{0pt}
    \renewcommand{\arraystretch}{0}
    \vspace{10pt}
    \begin{tabular}{@{}ccccc@{}}
        \multirow{2}{*}{\rotatebox[origin=c]{90}{Large \whitetxt{ssss}}} &
        \T{\includegraphics[width=\sizea, trim={\tal} {\tab} {\tar} {\tat},clip]{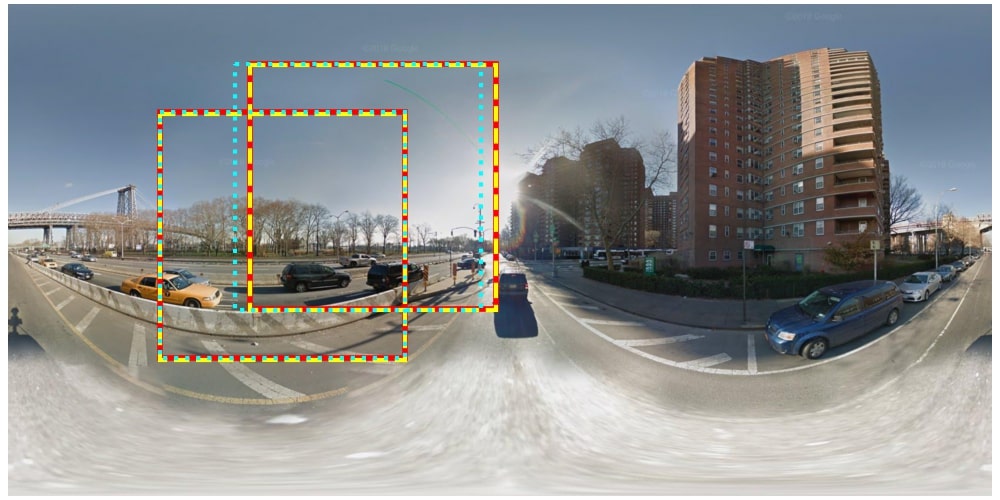}} &
        \T{\includegraphics[width=\sizea, trim={\tal} {\tab} {\tar} {\tat},clip]{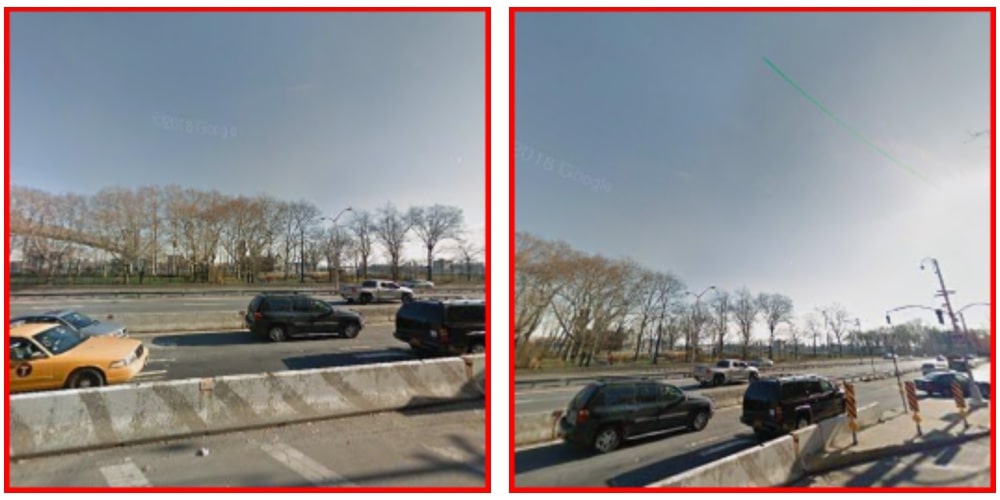}} &
        \T{\includegraphics[width=\sizea, trim={\tal} {\tab} {\tar} {\tat},clip]{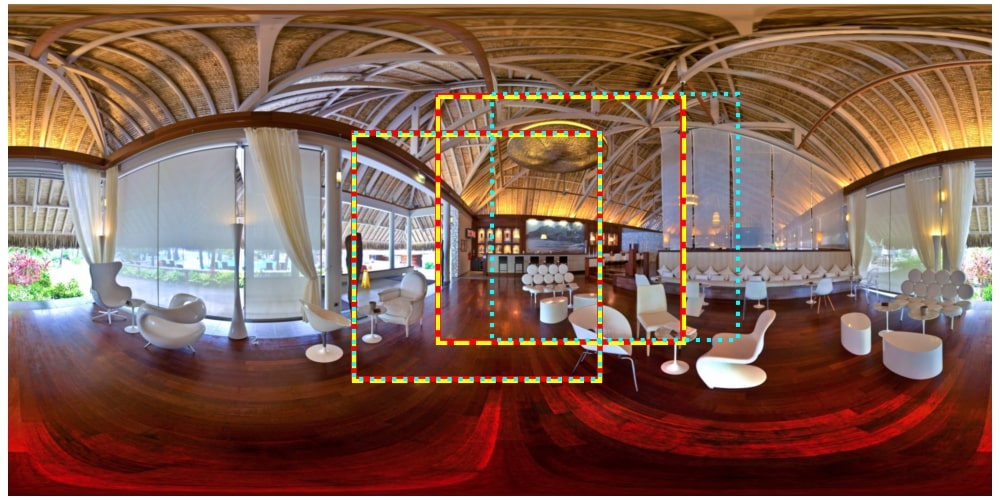}}&
        \T{\includegraphics[width=\sizea, trim={\tal} {\tab} {\tar} {\tat},clip]{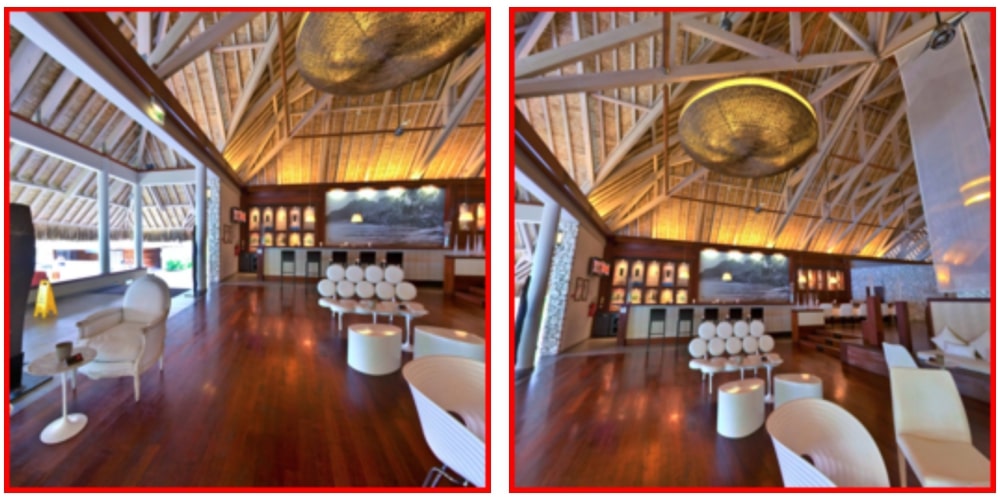}}
        \\  \vspace{+1pt}
        &
        \T{\includegraphics[width=\sizea, trim={\tal} {\tab} {\tar} {\tat},clip]{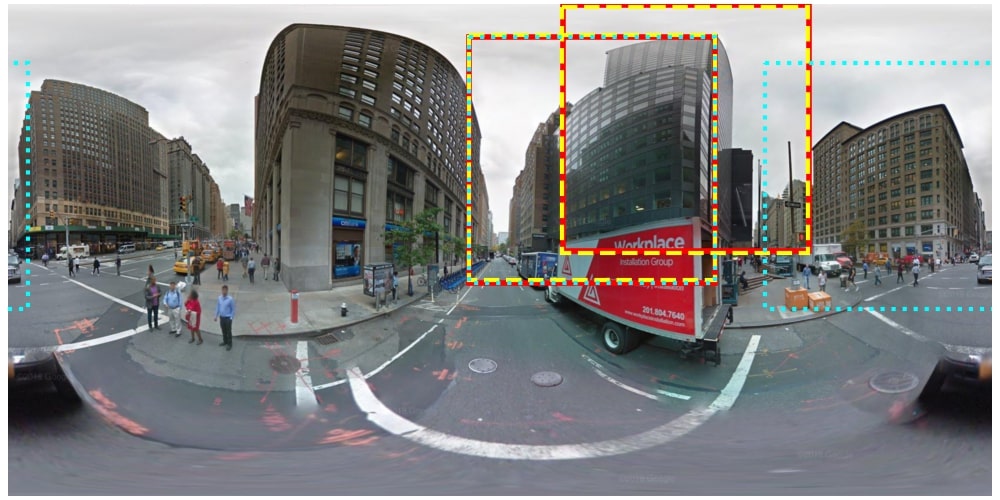}} &
        \T{\includegraphics[width=\sizea, trim={\tal} {\tab} {\tar} {\tat},clip]{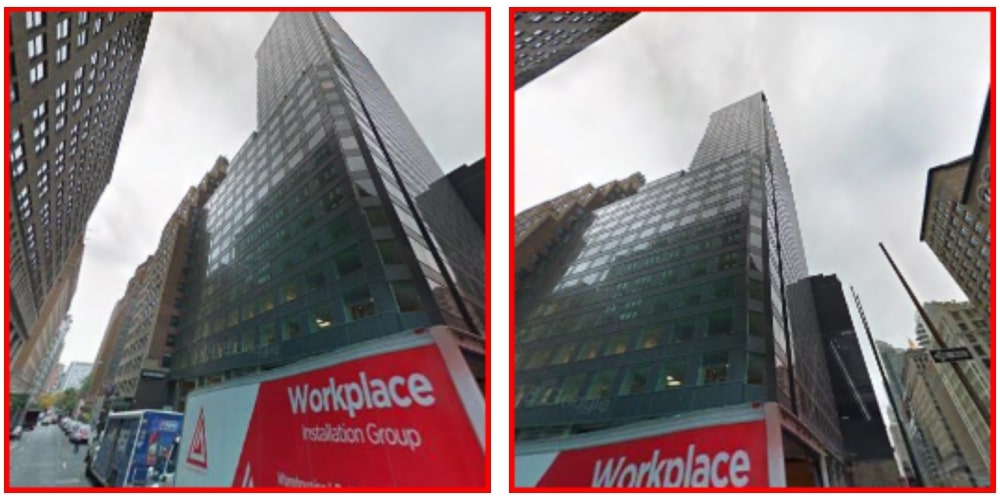}} &
        \T{\includegraphics[width=\sizea, trim={\tal} {\tab} {\tar} {\tat},clip]{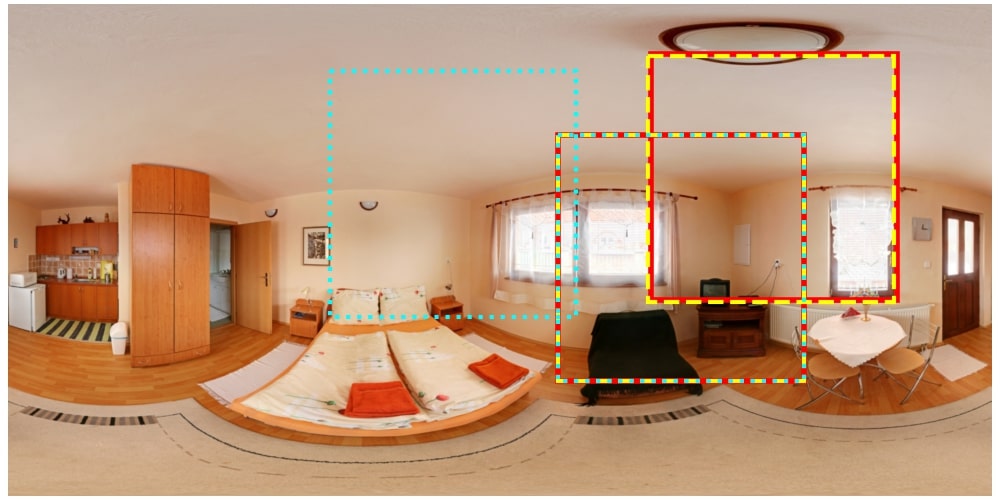}}&
        \T{\includegraphics[width=\sizea, trim={\tal} {\tab} {\tar} {\tat},clip]{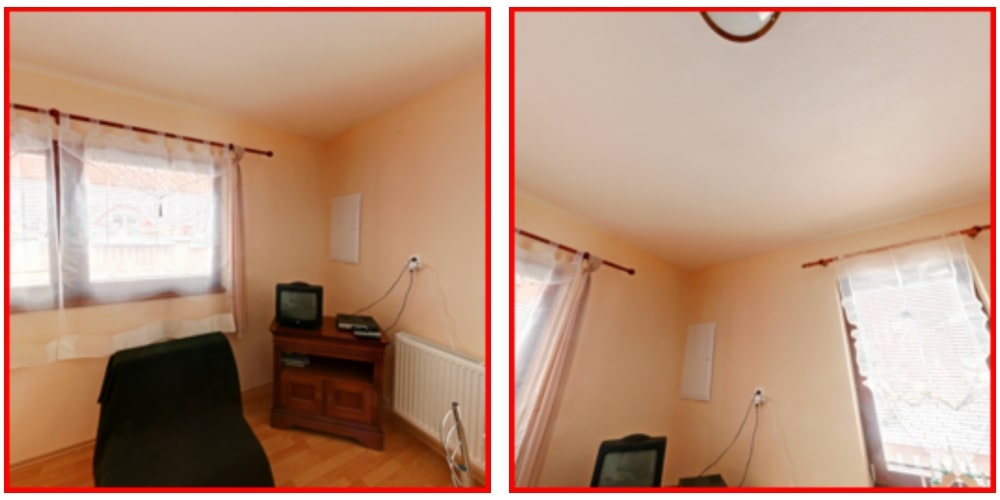}}
        \\ 
        
        \multirow{2}{*}{\rotatebox[origin=c]{90}{Small\whitetxt{ssss}}} &
        \T{\includegraphics[width=\sizea, trim={\tal} {\tab} {\tar} {\tat},clip]{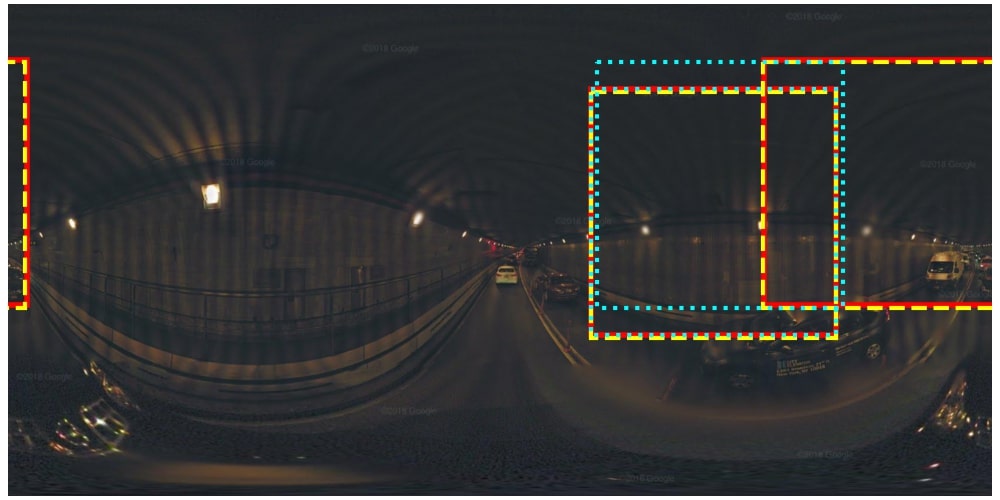}} &
        \T{\includegraphics[width=\sizea, trim={\tal} {\tab} {\tar} {\tat},clip]{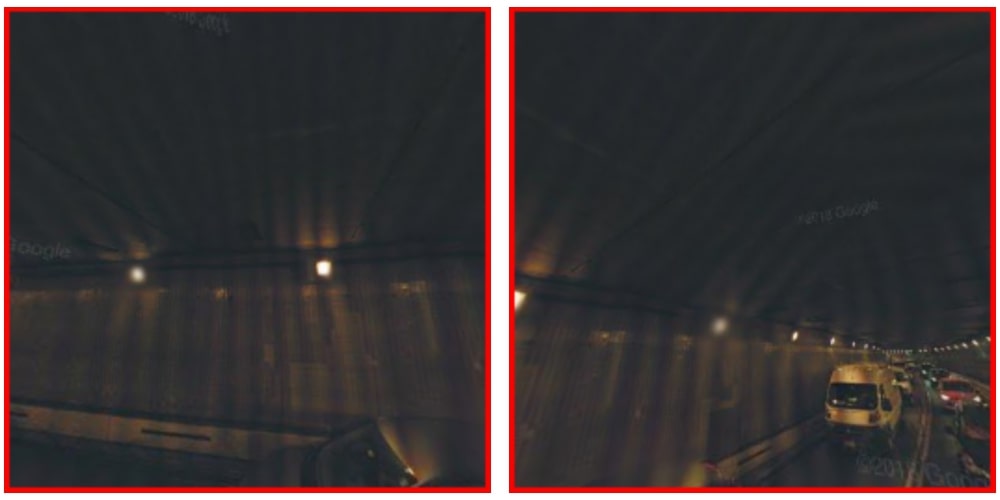}} &
        \T{\includegraphics[width=\sizea, trim={\tal} {\tab} {\tar} {\tat},clip]{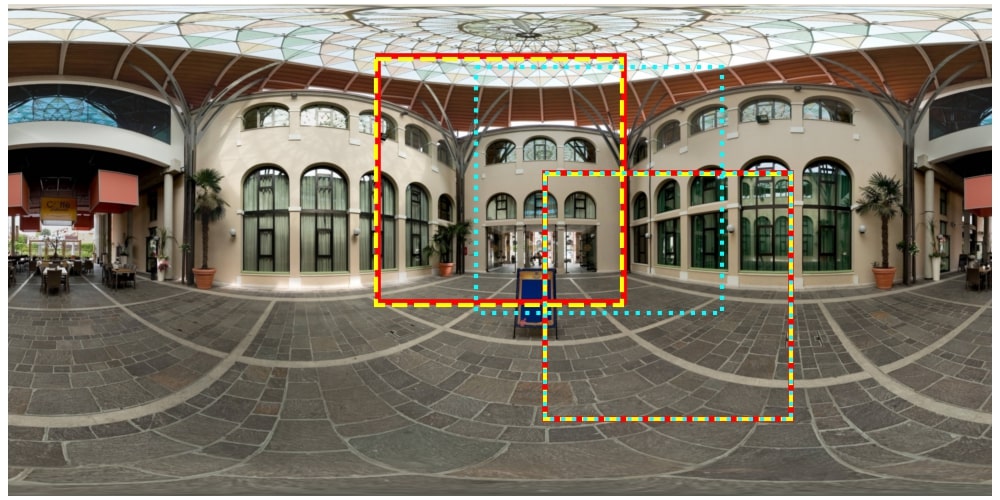}}&
        \T{\includegraphics[width=\sizea, trim={\tal} {\tab} {\tar} {\tat},clip]{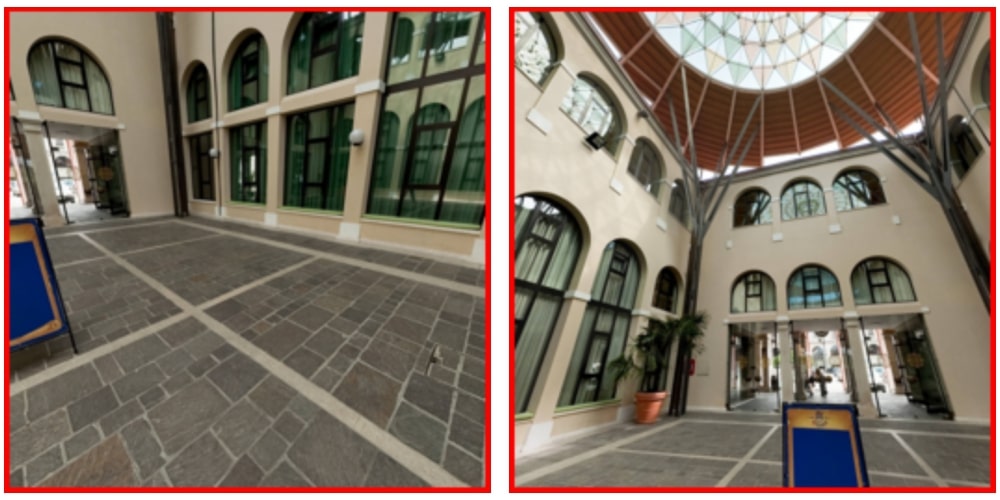}}
        \\ \vspace{+1pt}
        &
        \T{\includegraphics[width=\sizea, trim={\tal} {\tab} {\tar} {\tat},clip]{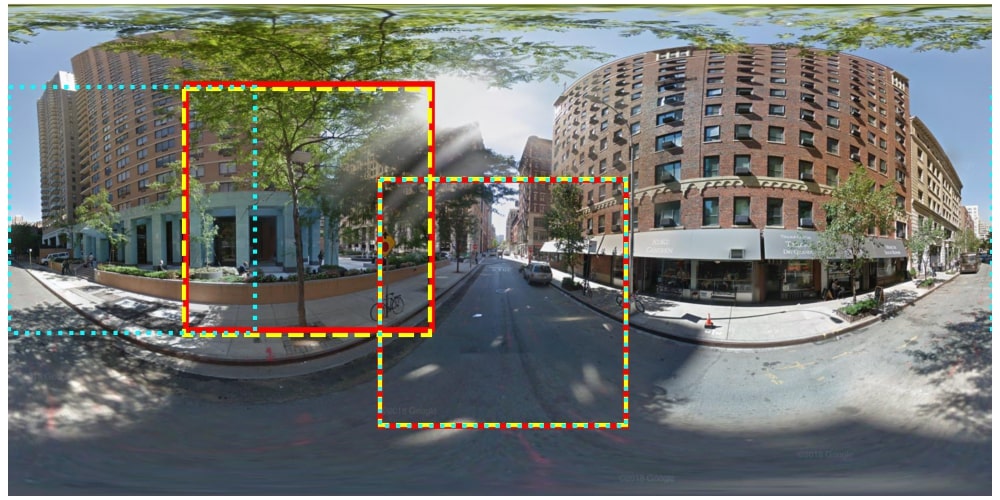}} &
        \T{\includegraphics[width=\sizea, trim={\tal} {\tab} {\tar} {\tat},clip]{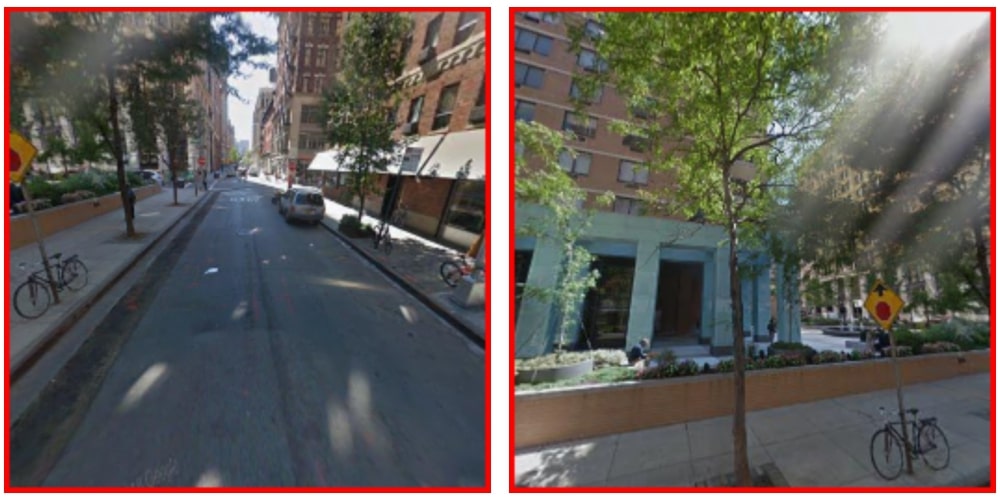}} &
        \T{\includegraphics[width=\sizea, trim={\tal} {\tab} {\tar} {\tat},clip]{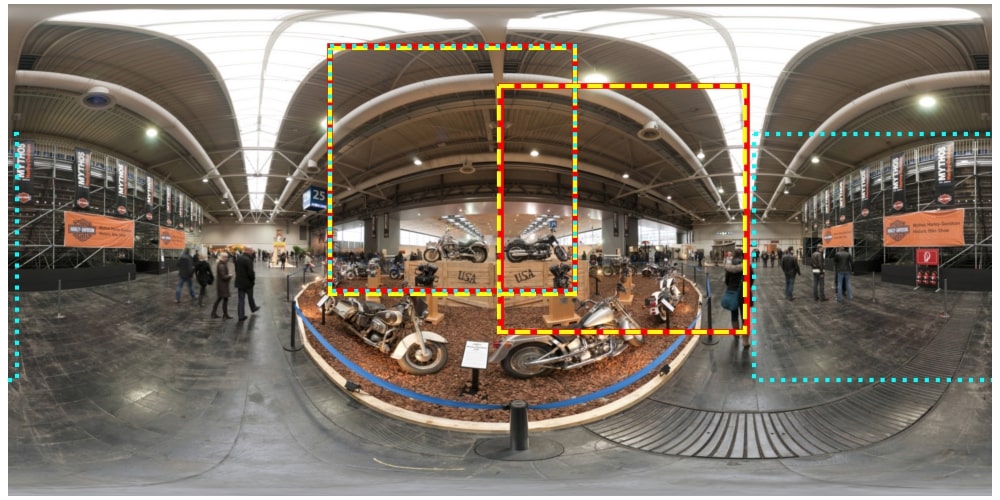}}&
        \T{\includegraphics[width=\sizea, trim={\tal} {\tab} {\tar} {\tat},clip]{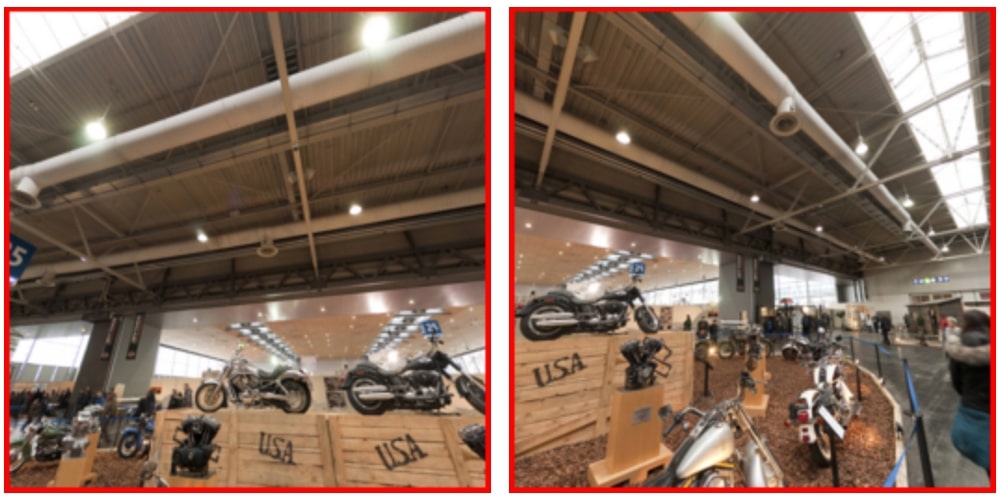}}
        \\
        \multirow{2}{*}{\rotatebox[origin=c]{90}{None\whitetxt{sssss}}} &
        \T{\includegraphics[width=\sizea, trim={\tal} {\tab} {\tar} {\tat},clip]{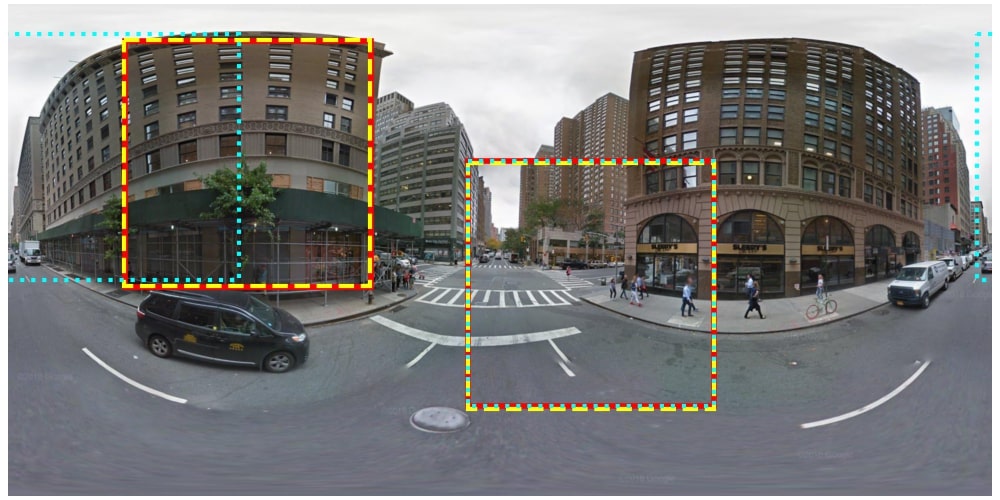}} &
        \T{\includegraphics[width=\sizea, trim={\tal} {\tab} {\tar} {\tat},clip]{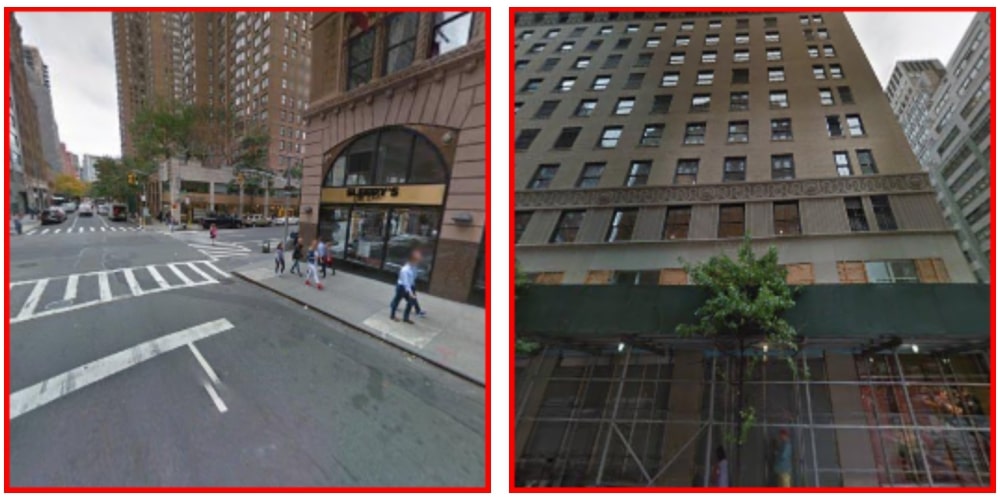}} &
        \T{\includegraphics[width=\sizea, trim={\tal} {\tab} {\tar} {\tat},clip]{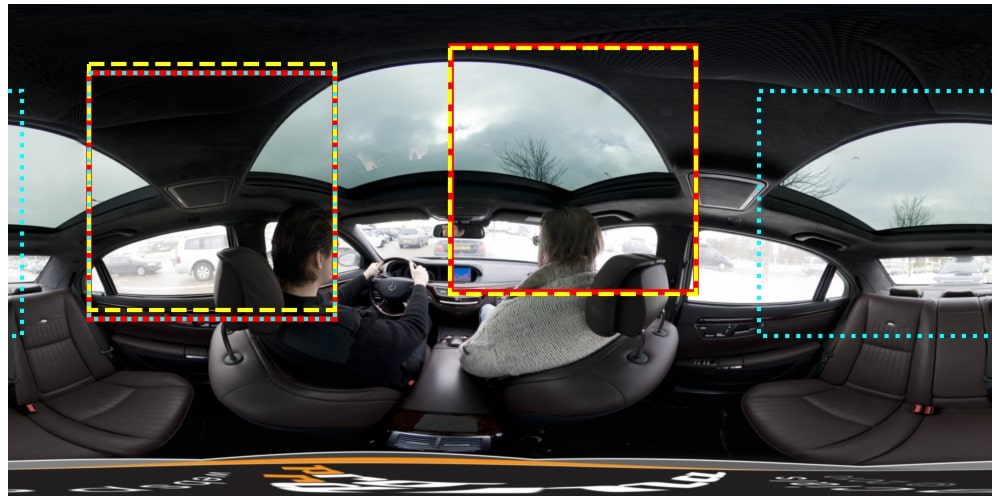}}&
        \T{\includegraphics[width=\sizea, trim={\tal} {\tab} {\tar} {\tat},clip]{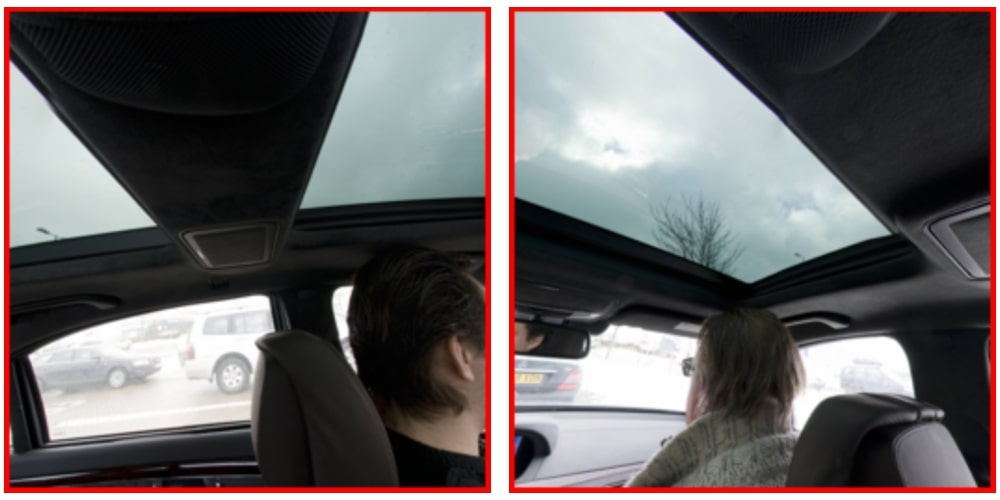}}
        \\  \vspace{+2pt}
        &
        \T{\includegraphics[width=\sizea, trim={\tal} {\tab} {\tar} {\tat},clip]{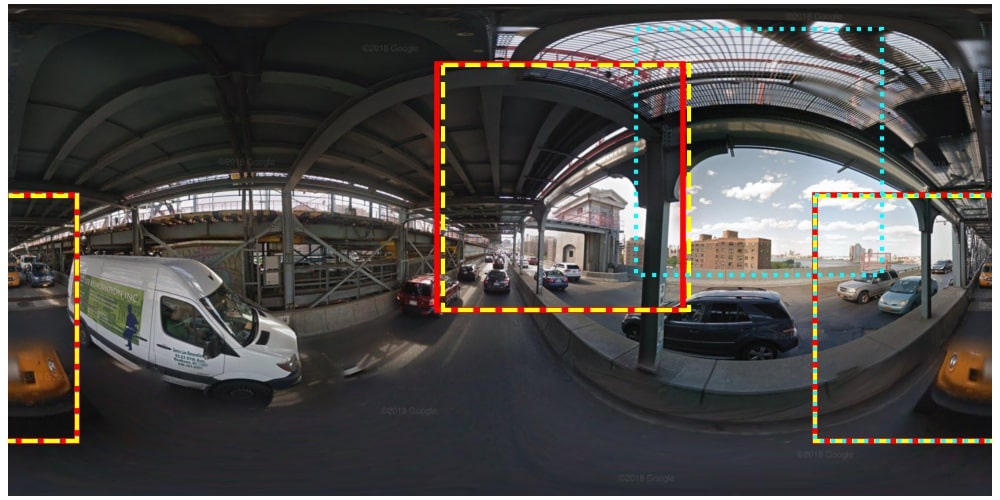}} &
        \T{\includegraphics[width=\sizea, trim={\tal} {\tab} {\tar} {\tat},clip]{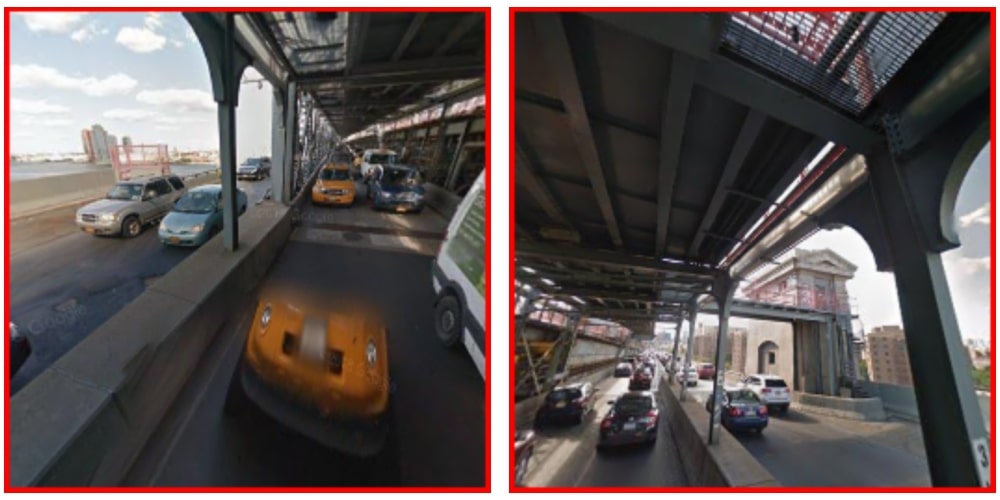}} &
        \T{\includegraphics[width=\sizea, trim={\tal} {\tab} {\tar} {\tat},clip]{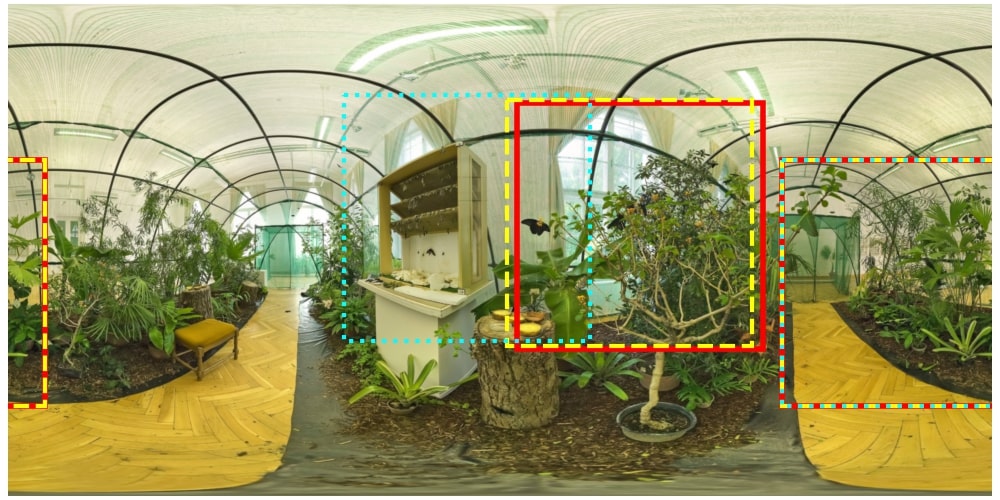}}&
        \T{\includegraphics[width=\sizea, trim={\tal} {\tab} {\tar} {\tat},clip]{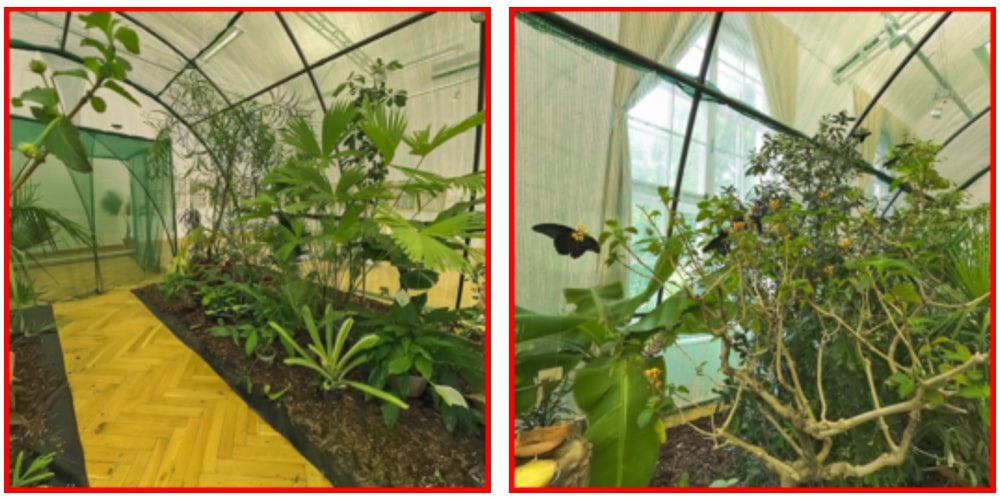}}
        \\ \vspace{-2pt}
        & \multicolumn{2}{c}{StreetLearn} 
        & \multicolumn{2}{c}{SUN360} 
    \end{tabular}
\vspace{-10pt}
    \end{center}
    \caption{\textbf{Predicted rotation results.} Full panoramas are shown on the left, with the ground-truth perspective images marked in red. We show our predicted viewpoints (in yellow) and results from the Reg6D regression model~\cite{zhou2019continuity} (in blue).
    }
    \label{fig:predicted}
\end{minipage}
\end{figure*}

\section{Conclusion}

We presented a method for estimating 3D rotations between a pair of RGB images in extreme settings where the images have little to no overlap. We demonstrate our method on a wide variety of example pairs and show that our model works surprisingly well on non-overlapping pairs, and can generalize to new scenes (including new cities) compared to the training data. Our key technical contribution is the use of dense correlation volumes that perform all-pairs feature correlations for this pose estimation problem, enabling our network to detect both explicit and implicit cues.
While we focus on relative rotations in our work, we believe that our method could be extended to compute whole camera poses, including camera intrinsics like focal length. However, we found that applying our approach directly to estimate relative translations did not work well for non-overlapping pairs. Addressing translations may require new architectures or by first predicting and removing rotation, as recently proposed by Chen \etal~\cite{directionnet}. 
In the future, we would also like to explore use of our internal features for other geometric tasks, like lighting or depth prediction---we believe that the use of non-overlapping relative pose as a proxy task may yield image features that are useful for a range of other tasks.

\smallskip
\noindent \textbf{Acknowledgements.} 
This work was supported in part by the National Science Foundation (IIS-2008313) and by the generosity of Eric and Wendy Schmidt by recommendation of the Schmidt Futures program and the Zuckerman STEM leadership program.

{\small
\bibliographystyle{ieee_fullname}
\bibliography{ref}
}

\newpage
\appendix

In this supplementary material, we provide additional implementation details (Section \ref{sec:implement}), experimental results (Section \ref{sec:quantitative}), and qualitative results (Section \ref{sec:qualitative}).

\section{Implementation Details}
\label{sec:implement}

\begin{figure*}
\centering
\includegraphics[width=\textwidth]{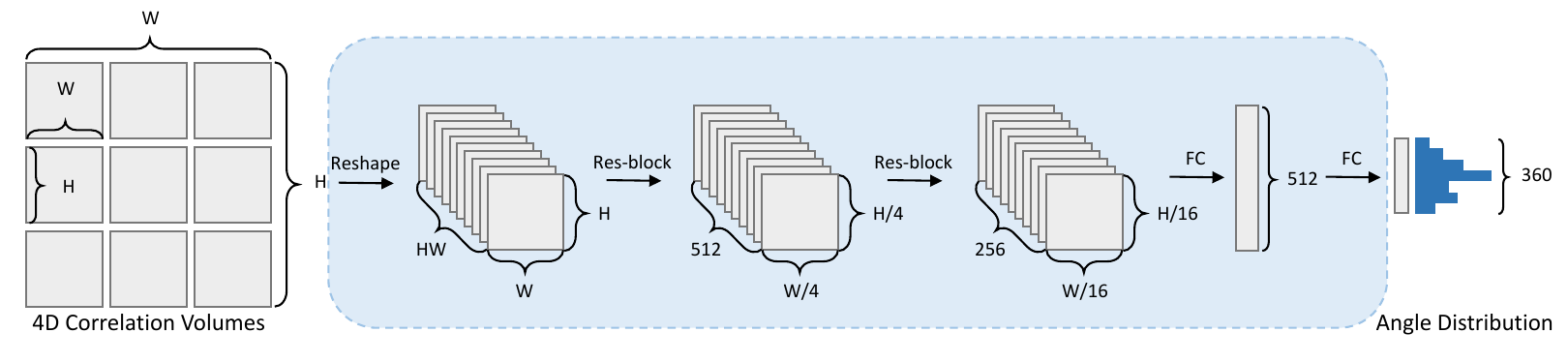}
   \caption{\textbf{Architecture of decoder.} 
    The 4D correlation volume is reshaped and processed with two pre-activation residual blocks, and eventually mapped by two fully-connected layers to a 360-dim distribution for each angle.
    }
\label{fig:decoder}
\end{figure*}
\subsection{Network architecture}

Our approach follows an encoder-decoder architecture as shown in Figure 2 in the main paper.
We adopt a Deep-Residual-Unet (ResUNet)~\cite{zhang2018road} architecture as the shared-weight Siamese \textbf{encoder}, which downsamples and upsamples the input image and outputs $32\times H/4 \times W/4$ embedded feature maps.
The image is first passed to a $7\times7$, stride-2 convolution layer, followed by three pre-activation residual blocks~\cite{he2016identity}, each of which downsamples by 2. 
Reaching the lowest resolution, the feature maps are upsampled by two ``up-convolution'' layers.
For each layer, the scale is set to 2 and a $3\times 3$ convolution is used.

We then efficiently compute pairwise extracted feature maps using matrix multiplication and output a \textbf{4D correlation volume} as a $H/4 \times W/4 \times H/4 \times W/4$ tensor.
The following \textbf{decoders}, as shown in Figure \ref{fig:decoder}, process the 4D correlation volume, which is reshaped to dimensionality $(H/4 \times W/4) \times H/4 \times W/4$, with two pre-activation residual blocks and uses two fully-connected layers to map to a 360-D distribution for each angle.
In total, our model contains $\sim$19M parameters, and the regression baseline contains $\sim$23M parameters.

\subsection{Experimental setting}

For all the experiments, we use the Adam optimizer ($\beta_1=0.5, \beta_2=0.9$). 
The initial learning rate for the model when using a classification loss is $5\times10^{-4}$ (and  $1\times10^{-4}$ for the models trained with a regression loss). 
All models are trained over 500k iterations with a batch size of 20, using a linear decay strategy, where the learning rate drops starting from 250k iteration and ends up at $5\times10^{-6}$ for model with classification loss (and $1\times10^{-6}$ for models trained using a regression loss).
All the models are trained and evaluated using the same training strategy.
The training time is about two days with one nVidia GeForce RTX 2080 Ti GPU.

\subsection{Baselines}
In this section, we provide more details on the baselines, including ones not described in detail in the main paper due to space constraints.

\medskip
\noindent \textbf{SIFT-based relative rotation estimation}. 
First, we detect local features in images of size 256$\times$256 with SIFT~\cite{lowe2004distinctive}, then features are matched across images by fitting a model using RANSAC~\cite{fischler1981random}.
The minimal number of inlier matches is set to 10. 
We compute rotation matrices with a 2-point algorithm~\cite{brown2007minimal} for image pairs from the same panorama, and for image pairs captured under camera translation we compute and decompose an essential matrix. We use the publicly available OpenCV implementation for each of these steps (except for the 2-point algorithm for which we use our own implementation).

\smallskip
\noindent \textbf{Learning-based feature matching}. We use the following pretrained networks:
\begin{itemize}
    \item \textbf{D2-Net}~\cite{Dusmanu2019CVPR}. 
    Detect-and-describe (D2) networks use an ImageNet pretrained VGG16~\cite{simonyan2014very} network as the feature extraction network. The extracted feature maps are used to detect keypoints with hard feature detections and serve as local descriptors at the same time.
    \item \textbf{SuperPoint}~\cite{detone18superpoint}.
    SuperPoint uses a VGG-style~\cite{simonyan2014very} encoder, and passes the feature maps to two separate decoders for interest point detection and description. 
    The head of the interest point decoder adopts sub-pixel convolution~\cite{shi2016real}, and the head of the descriptor decoder uses a model similar to Universal Correspondence Networks~\cite{choy2016universal}. Both decoders use non-learned upsampling modules to compute the output.
    
\end{itemize}
These networks detect interest points in images and generate corresponding dense feature descriptors. We then estimate the rotation matrix using a model fitting technique following the procedure described above.

\medskip
\noindent \textbf{End-to-end relative rotation regression}.
We use the following regression baselines:
\begin{itemize}
    \item Zhou \etal~\cite{zhou2019continuity}, hereby denoted by \emph{Reg6D}. 
    Image features are concatenated and fed to a regression model predicting a continuous representation in 6D. We report two different models: (1) \emph{Reg6D-128}, which adopts the same network architecture, pretrained weights, and the same input resolution ($128\times 128$) as our method, and (2) \emph{Reg6D}, where we replaced the ResUNet architecture of the encoder with ResNet, which outputs $1024\times H/64 \times W/64$ feature maps, as we found this modification yields improved performance. In addition, the resolution of the input images for this model is $256\times 256$.
    \item En \etal~\cite{en2018rpnet}, hereby denoted by \emph{Reg4D}. The learning-based method proposed in En \etal concatenates the pairwise images features and then regresses them to predict a quaternion representation. We adopt the same network architecture as for the Reg6D model, and the resolution of the input images for the baselines is also set to $256\times 256$.
    \item \emph{RegEuler}. We implemented an additional regression baseline built on our network architecture, which 
    regresses the concatenated feature to an Euler angle representation. For this additional baseline, we report results for input images of size $128\times 128$, which is the input resolution used by our model.
\end{itemize}

\section{Additional Experimental Results}
\label{sec:quantitative}

\subsection{Comparison to all baselines}

\definecolor{graytext}{RGB}{130,130,130}

\begin{table*}[t]
\setlength{\tabcolsep}{1.0pt}
\def\arraystretch{1.05}
 \def\arraystretch{1.00}
 \footnotesize
\scriptsize
\begin{center}
\begin{tabularx}{\textwidth}{llccclccclccclccclccc}
\toprule
\multicolumn{2}{l}{}& \multicolumn{3}{c}{InteriorNet} &  & \multicolumn{3}{c}{InteriorNet-T} &  & \multicolumn{3}{c}{SUN360} &  & \multicolumn{3}{c}{StreetLearn} &  & \multicolumn{3}{c}{StreetLearn-T}\\ \cline{3-5} \cline{7-9} \cline{11-13} \cline{15-17} \cline{19-21}
Overlap 
& Method
& Avg(\degree$\downarrow$) & Med(\degree$\downarrow$) & \multicolumn{1}{c}{10\degree(\%$\uparrow$)} & & Avg(\degree$\downarrow$) & Med(\degree$\downarrow$) & \multicolumn{1}{c}{10\degree(\%$\uparrow$)} &  & Avg(\degree$\downarrow$) & Med(\degree$\downarrow$) & \multicolumn{1}{c}{10\degree(\%$\uparrow$)} &  & Avg(\degree$\downarrow$) & Med(\degree$\downarrow$) & \multicolumn{1}{c}{10\degree(\%$\uparrow$)} &  & Avg(\degree$\downarrow$) & Med(\degree$\downarrow$) & \multicolumn{1}{c}{10\degree(\%$\uparrow$)} \\ \midrule
\multirow{13}{*}{Large} 
                         & SIFT*~\cite{lowe2004distinctive}       & 6.09                 & 4.00                 & 84.86               &  & 7.78                 & 2.95                 & 55.52               &  & 5.46                       & 3.88                       & 93.10                      &  & 5.84                 & 3.16                 & 91.18               &  & \color{graytext}{18.86}                & \color{graytext}{3.13}                 & \color{graytext}{22.37}               \\
                         & D2-Net*~\cite{Dusmanu2019CVPR}         & 8.27                 & 3.78                 & 69.48               &  & \color{graytext}{14.19}                & \color{graytext}{9.38}                 & \color{graytext}{15.82}               &  & 10.48                      & 4.22                       & 69.46                      &  & 12.85                & 4.42                 & 54.12               &  & \color{graytext}{8.76}                 & \color{graytext}{6.73}                 & \color{graytext}{1.32}                \\
                         & SuperPoint*~\cite{detone18superpoint}  & 5.40                 & 3.53                 & 87.10               &  & 5.46                 & 2.79                 & 65.97               &  & 4.69                       & 3.18                       & 92.12                      &  & 6.23                 & 3.61                 & 91.18               &  & \color{graytext}{6.38}        & \color{graytext}{1.79}        & \color{graytext}{16.45}               \\
                         & RegEuler-o                            & 4.89                 & 3.52                 & 91.56               &  & 8.36                 & 4.64                 & 80.00               &  & 6.57                       & 5.16                       & 85.22                      &  & 3.99                 & 3.20                 & 95.29               &  & 15.15                & 7.13                 & 62.50               \\
                         & RegEuler                               & 11.11                & 7.92                 & 60.05               &  & 20.98                & 14.59                & 33.43               &  & 14.19          & 10.77         & 46.31                       &  & 27.47                & 17.59                & 23.53               &  & 46.89                & 32.36                & 11.84               \\
                         & Reg4D~\cite{en2018rpnet}-o            & 4.95                 & 3.47                 & 91.56               &  & 9.58                 & 7.00                 & 69.25               &  & 7.86                       & 5.89                       & 76.85                      &  & 3.14                 & 2.67                 & 98.82               &  & 11.73                & 6.35                 & 74.34               \\
                         & Reg4D~\cite{en2018rpnet}            & 13.83                & 10.26                & 48.88               &  & 26.05                & 16.84                & 22.99               &  & 36.02                      & 24.23                      & 16.26                      &  & 20.57                & 13.05                & 40.00               &  & 41.44                & 28.27                & 20.39               \\
                         & Reg6D~\cite{zhou2019continuity}-o-128 & 7.48                 & 5.24                 & 75.19               &  & 14.58                & 10.97                & 46.27               &  & 11.97                      & 8.11                       & 57.64                      &  & 6.00                 & 5.20                 & 85.29               &  & 13.66                & 8.24                 & 58.55               \\
                         & Reg6D~\cite{zhou2019continuity}-128    & 14.04                & 9.06                 & 53.10               &  & 31.96                & 21.19                & 22.99               &  & 40.19                      & 33.41                      & 8.87                       &  & 14.09                & 8.63                 & 56.47               &  & 31.01                & 19.14                & 26.97               \\
                         & Reg6D~\cite{zhou2019continuity}-o     & 5.43                 & 3.87                 & 87.10               &  & 10.45                & 6.91                 & 67.76               &  & 7.18                       & 5.79                       & 81.28                      &  & 3.36                 & 2.71                 & 97.65               &  & 12.31                & 6.02                 & 69.08               \\
                         & Reg6D~\cite{zhou2019continuity}        & 9.05                 & 5.90                 & 68.49               &  & 17.00                & 11.95                & 41.79               &  & 16.51                      & 12.43                      & 40.39                      &  & 11.70                & 8.87                 & 58.24               &  & 36.71                & 24.79                & 23.03               \\
                         & Ours-o                                & \textbf{1.53}        & 1.10                 & \textbf{99.26}     &  & \textbf{2.89}        & \textbf{1.10}        & \textbf{97.61}     &  & \textbf{1.00}              & \textbf{0.94}              & \textbf{100.00}           &  & \textbf{1.19}        & \textbf{1.02}        & \textbf{99.41}     &  & 9.12                 & 2.91                 & \textbf{87.50}     \\
                         & Ours                                   & 1.82                 & \textbf{0.88}        & 98.76               &  & 8.86                 & 1.86                 & 93.13               &  & 1.37                       & 1.09                       & 99.51                      &  & 1.52                 & 1.09                 & \textbf{99.41}     &  & 24.98                & 2.48                 & 78.95               \\
\midrule 
\multirow{13}{*}{Small} 
                         & SIFT*~\cite{lowe2004distinctive}       & 24.18                & 8.57                 & 39.73               &  & \color{graytext}{18.16}                & \color{graytext}{10.01}                & \color{graytext}{18.52}               &  & 13.71                      & 6.33                       & 56.77                      &  & 16.22                & 7.35                 & 55.81               &  & \color{graytext}{38.78}                & \color{graytext}{13.81}                & \color{graytext}{5.68}                \\
                         & D2-Net*~\cite{Dusmanu2019CVPR}         & \color{graytext}{14.21}                & \color{graytext}{8.50}                 & \color{graytext}{3.42}                &  & \color{graytext}{--}                   & \color{graytext}{--}                   & \color{graytext}{0.00}                &  & \color{graytext}{25.49}                      & \color{graytext}{9.17}                       & \color{graytext}{4.51}                       &  & \color{graytext}{41.35}                & \color{graytext}{18.43}                & \color{graytext}{1.66}                &  & \color{graytext}{67.27}                & \color{graytext}{67.27}                & \color{graytext}{0.00}                \\
                         & SuperPoint*~\cite{detone18superpoint}  & \color{graytext}{16.72}                & \color{graytext}{8.43}                 & \color{graytext}{21.58}               &  & \color{graytext}{11.61}                & \color{graytext}{5.82}                 & \color{graytext}{11.73}               &  & \color{graytext}{17.63}                      & \color{graytext}{7.70}                       & \color{graytext}{26.69}                      &  & \color{graytext}{19.29}                & \color{graytext}{7.60}                 & \color{graytext}{24.58}               &  & \color{graytext}{6.80}        & \color{graytext}{6.85}                 & \color{graytext}{0.95}                \\
                         & RegEuler-o                            & 15.37                & 7.88                 & 59.59               &  & 19.27                & 7.41                 & 64.51               &  & 14.88                      & 8.92                       & 54.14                      &  & 8.24                 & 4.29                 & 86.71               &  & 20.41                & 9.96                 & 50.47               \\
                         & RegEuler                               & 28.58                & 17.77                & 31.85               &  & 32.87                & 21.96                & 18.52               &  & 32.28          & 25.41         & 12.41                       &  & 52.08                & 37.35                & 4.32                &  & 58.43                & 47.27                & 7.57                \\
                         & Reg4D~\cite{en2018rpnet}-o            & 16.62                & 8.46                 & 57.88               &  & 24.77                & 12.42                & 41.67               &  & 15.02                      & 9.81                       & 51.88                      &  & 6.55                 & 4.15                 & 91.03               &  & 14.56                & 7.46                 & 65.93               \\
                         & Reg4D~\cite{en2018rpnet}            & 32.86                & 22.41                & 15.75               &  & 41.52                & 28.46                & 10.80               &  & 66.59                      & 57.84                      & 0.38                       &  & 38.39                & 24.77                & 10.63               &  & 50.42                & 34.07                & 15.46               \\
                         & Reg6D~\cite{zhou2019continuity}-o-128 & 22.45                & 12.57                & 38.70               &  & 31.15                & 16.58                & 27.47               &  & 23.17                      & 15.08                      & 24.06                      &  & 9.12                 & 5.71                 & 82.06               &  & 20.70                & 10.69                & 45.43               \\
                         & Reg6D~\cite{zhou2019continuity}-128    & 36.37                & 23.81                & 18.84               &  & 54.24                & 39.94                & 14.81               &  & 67.97                      & 60.91                      & 0.38                       &  & 24.03                & 15.13                & 30.56               &  & 41.07                & 28.33                & 17.03               \\
                         & Reg6D~\cite{zhou2019continuity}-o     & 17.83                & 9.61                 & 51.37               &  & 21.87                & 11.43                & 44.14               &  & 18.61                      & 11.66                      & 39.85                      &  & 7.95                 & 4.34                 & 87.71               &  & 15.07                & 7.59                 & 63.41               \\
                         & Reg6D~\cite{zhou2019continuity}        & 25.71                & 15.56                & 33.56               &  & 42.93                & 28.92                & 23.15               &  & 42.55                      & 32.11                      & 9.40                       &  & 24.77                & 15.11                & 30.56               &  & 46.61                & 34.33                & 13.88               \\
                         & Ours-o                                & 6.45                 & 1.61                 & 95.89               &  & \textbf{10.24}       & \textbf{1.38}        & \textbf{89.81}     &  & \textbf{3.09}              & \textbf{1.41}              & \textbf{98.50}            &  & \textbf{2.32}        & \textbf{1.41}        & \textbf{98.67}     &  & \textbf{13.04}                & 3.49                 & \textbf{84.23}     \\
                         & Ours                                   & \textbf{4.31}        & \textbf{1.16}        & \textbf{96.58}     &  & 30.43                & 2.63                 & 74.07               &  & 6.13                       & 1.77                       & 95.86                      &  & 3.23                 & \textbf{1.41}        & 98.34               &  & 27.84                & \textbf{3.19}        & 74.76               \\
 \midrule 
\multirow{8}{*}{None}   
                         & SIFT*~\cite{lowe2004distinctive}       & \color{graytext}{109.30}               & \color{graytext}{92.86}                & \color{graytext}{0.00}                &  & \color{graytext}{93.79}                & \color{graytext}{113.86}               & \color{graytext}{0.00}                &  & \color{graytext}{127.61}                     & \color{graytext}{129.07}                     & \color{graytext}{0.00}                       &  & \color{graytext}{83.49}                & \color{graytext}{90.00}                & \color{graytext}{0.38}                &  & \color{graytext}{85.90}                & \color{graytext}{106.84}               & \color{graytext}{0.38}                \\
                         & D2-Net*~\cite{Dusmanu2019CVPR}         & \color{graytext}{--}                   & \color{graytext}{--}                   & \color{graytext}{0.00}                &  & \color{graytext}{--}                   & \color{graytext}{--}                   & \color{graytext}{0.00}                &  & \color{graytext}{171.21}                     & \color{graytext}{171.21}                     & \color{graytext}{0.00}                       &  & \color{graytext}{--}                   & \color{graytext}{--}                   & \color{graytext}{0.00}                &  & \color{graytext}{--}                   & \color{graytext}{--}                   & \color{graytext}{0.00}                \\
                         & SuperPoint*~\cite{detone18superpoint}  & \color{graytext}{120.28}               & \color{graytext}{120.28}               & \color{graytext}{0.00}                &  & \color{graytext}{--}                   & \color{graytext}{--}                   & \color{graytext}{0.00}                &  & \color{graytext}{149.80}                     & \color{graytext}{165.24}                     & \color{graytext}{0.00}                       &  & \color{graytext}{--}                   & \color{graytext}{--}                   & \color{graytext}{0.00}                &  & \color{graytext}{--}                   & \color{graytext}{--}                   & \color{graytext}{0.00}                \\
                         & RegEuler                               & 52.95                & 36.03                & 7.87                &  & 55.73                & 42.04                & 9.97                & & 70.93          & 59.43         & 5.46                       &  & 55.92                & 41.23                & 7.56                &  & 61.04                & 48.79                & 9.04                \\
                         & Reg4D~\cite{en2018rpnet}            & 62.04                & 48.92                & 4.59                &  & 59.85                & 48.81                & 4.99                &  & 80.08                      & 72.78                      & 1.32                       &  & 46.19                & 32.74                & 9.26                &  & 55.70                & 39.70                & 9.79                \\
                         & Reg6D~\cite{zhou2019continuity}-128    & 64.59                & 49.80                & 5.90                &  & 79.86                & 71.60                & 5.87                &  & 83.29                      & 75.08                      & 0.56                       &  & 34.78                & 23.16                & 17.77               &  & 50.96                & 36.50                & 9.23                \\
                         & Reg6D~\cite{zhou2019continuity}        & 48.36                & 32.93                & 10.82               &  & 60.91                & 51.26                & 11.14               &  & 64.74                      & 56.55                      & 3.77                       &  & 28.48                & 18.86                & 24.39               &  & 49.23                & 35.66                & 11.86               \\
                         & Ours                                   & \textbf{37.69}       & \textbf{3.15}        & \textbf{61.97}     &  & \textbf{49.44}       & \textbf{4.17}        & \textbf{58.36}     &  & \textbf{34.92}             & \textbf{4.43}              & \textbf{61.39}            &  & \textbf{5.77}        & \textbf{1.53}        & \textbf{96.41}     &  & \textbf{30.98}       & \textbf{3.50}        & \textbf{72.69}     \\
\midrule
\multirow{8}{*}{All}    
                         & SIFT*~\cite{lowe2004distinctive}       & 13.68                & 5.04                 & 45.80               &  & \color{graytext}{12.24}                & \color{graytext}{5.69}                 & \color{graytext}{24.60}               &  & \color{graytext}{18.12}                      & \color{graytext}{5.02}                       & \color{graytext}{34.00}                      &  & \color{graytext}{17.29}                & \color{graytext}{5.53}                 & \color{graytext}{32.50}               &  & \color{graytext}{36.00}                & \color{graytext}{6.03}                 & \color{graytext}{5.40}                \\
                         & D2-Net*~\cite{Dusmanu2019CVPR}         & \color{graytext}{8.56}                 & \color{graytext}{3.95}                 & \color{graytext}{29.00}               &  & \color{graytext}{14.19}                & \color{graytext}{9.38}                 & \color{graytext}{5.30}                &  & \color{graytext}{13.80}                      & \color{graytext}{4.62}                       & \color{graytext}{15.30}                      &  & \color{graytext}{16.41}                & \color{graytext}{5.38}                 & \color{graytext}{9.70}                &  & \color{graytext}{23.39}                & \color{graytext}{11.87}                & \color{graytext}{0.20}                \\
                         & SuperPoint*~\cite{detone18superpoint}  & \color{graytext}{8.19}        & \color{graytext}{4.08}                 & \color{graytext}{41.40}               &  & \color{graytext}{6.62}        & \color{graytext}{3.38}                 & \color{graytext}{25.90}               &  & \color{graytext}{11.09}             & \color{graytext}{4.00}                       & \color{graytext}{25.80}                      &  & \color{graytext}{11.52}                & \color{graytext}{4.80}                 & \color{graytext}{22.90}               &  & \color{graytext}{6.42}        & \color{graytext}{2.62}        & \color{graytext}{2.80}                \\
                         & RegEuler                               & 28.97                & 15.53                & 35.90               &  & 36.68                & 22.70                & 20.60               &  & 49.13          & 31.44         & 15.60                       &  & 49.93                & 35.08                & 9.30                &  & 58.06                & 45.54                & 9.00                \\
                         & Reg4D~\cite{en2018rpnet}          & 34.09                & 19.59                & 25.70               &  & 42.59                & 28.35                & 12.90               &  & 67.55                      & 55.40                      & 4.10                       &  & 39.48                & 25.76                & 14.90               &  & 51.86                & 36.89                & 13.20               \\
                         & Reg6D~\cite{zhou2019continuity}-128    & 35.98                & 18.65                & 28.70               &  & 55.51                & 36.90                & 14.50               &  & 70.46                      & 57.87                      & 2.20                       &  & 28.03                & 17.11                & 28.20               &  & 44.79                & 31.00                & 14.40               \\
                         & Reg6D~\cite{zhou2019continuity}        & 25.90                & 13.02                & 40.70               &  & 40.38                & 23.35                & 25.30               &  & 49.05                      & 34.37                      & 12.70                      &  & 24.51                & 15.31                & 32.00               &  & 46.50                & 33.14                & 29.90               \\
                         & Ours                                   & \textbf{13.49}                & \textbf{1.18}        & \textbf{86.90}     &  & \textbf{29.68}                & \textbf{2.58}        & \textbf{75.10}     &  & \textbf{20.45}                      & \textbf{2.23}              & \textbf{78.30}            &  & \textbf{4.40}        & \textbf{1.44}        & \textbf{97.50}     &  & \textbf{29.85}                & \textbf{3.20}                 & \textbf{74.30}     \\ 
                        \bottomrule 
\end{tabularx}
\end{center}
\vspace{-5pt}
\caption{\textbf{Rotation estimation evaluation on the InteriorNet, SUN360, and StreetLearn datasets.} 
We report the mean and median geodesic error in degrees, and the percentage of pairs with a relative rotation error under 10$\degree$, for different overlapping levels (Large, Small, and None), as detailed in Section 4.3 of the main paper. For the percentage of pairs (10$\degree$\%), higher is better. Models trained only on overlapping pairs are denoted with ``-o''.
*For the indicated methods, mean and median errors are computed only over successful image pairs, for which these algorithms output an estimated rotation matrix (cases where there is failure over more than $50\%$ of the test pairs is shown in gray). 
}
\label{tab:main_result_supp}
\end{table*}

We compare our approach to all the baselines described in the previous section 
and report the results in Table \ref{tab:main_result_supp}. 
These results further validate that our approach consistently outperforms alternative techniques. Note that the strongest baselines are reported also in the main paper.

We show the number of successful image pairs for SIFT, D2-Net and SuperPoint in Table \ref{tab:sift_result}. We observe a significant drop in SuperPoint's ability to find enough correct matches to estimate the rotation matrix 
when the image pairs have larger rotations and translations. 

Also, we can see that using smaller input images ($128\times 128$ vs.\ $256\times 256$) negatively affects the performance of Reg6D in non-overlapping cases, increasing the gap with our models (which are trained using images of size $128\times 128$) from a mean error of 29.82$\degree$ to 48.37$\degree$ on SUN360 and from 22.71$\degree$ to 29.01$\degree$ on StreetLearn. 

\begin{table*}[]
\setlength{\tabcolsep}{2.3pt}
\small
\begin{center}
\begin{tabularx}{\textwidth}{llcccccccccccccccccccc}
\toprule

 & &\multicolumn{3}{c}{InteriorNet}   &  & \multicolumn{3}{c}{InteriorNet-T} &  & \multicolumn{3}{c}{SUN360} &  & \multicolumn{3}{c}{StreetLearn}  &  & \multicolumn{3}{c}{StreetLearn-T} \\  \cline{3-5} \cline{7-9} \cline{11-13} \cline{15-17} \cline{19-21}
\%Overlap & Method     & Success & Fail & Total & & Success & Fail & Total & & Success & Fail & Total &  & Success & Fail & Total &  & Success & Fail & Total \\ \midrule
\multirow{3}{*}{Large} & SIFT~\cite{lowe2004distinctive}      & 371 & 32  & 403  &  & 252 & 83  & 335  &  & 201 & 2   & 203  &  & 166 & 4   & 170  &  & 50  & 102 & 152  \\
                       & D2-Net~\cite{Dusmanu2019CVPR}        & 330 & 73  & 403  &  & 102 & 233 & 335  &  & 176 & 27  & 203  &  & 119 & 51  & 170  &  & 3   & 149 & 152  \\
                       & SuperPoint~\cite{detone18superpoint} & 382 & 21  & 403  &  & 259 & 76  & 335  &  & 201 & 2   & 203  &  & 169 & 1   & 170  &  & 30  & 122 & 152  \\ \midrule
\multirow{3}{*}{Small} & SIFT~\cite{lowe2004distinctive}      & 195 & 97  & 292  &  & 121 & 203 & 324  &  & 217 & 49  & 266  &  & 267 & 34  & 301  &  & 39  & 278 & 317  \\
                       & D2-Net~\cite{Dusmanu2019CVPR}        & 17  & 275 & 292  &  & 0   & 324 & 324  &  & 23  & 243 & 266  &  & 17  & 284 & 301  &  & 1   & 316 & 317  \\
                       & SuperPoint~\cite{detone18superpoint} & 112 & 180 & 292  &  & 60  & 264 & 324  &  & 112 & 154 & 266  &  & 115 & 186 & 301  &  & 3   & 314 & 317  \\ \midrule
\multirow{3}{*}{None}  & SIFT~\cite{lowe2004distinctive}      & 8   & 297 & 305  &  & 5   & 336 & 341  &  & 32  & 499 & 531  &  & 27  & 502 & 529  &  & 15  & 516 & 531  \\
                       & D2-Net~\cite{Dusmanu2019CVPR}        & 0   & 305 & 305  &  & 0   & 341 & 341  &  & 2   & 529 & 531  &  & 0   & 529 & 529  &  & 0   & 531 & 531  \\
                       & SuperPoint~\cite{detone18superpoint} & 1   & 304 & 305  &  & 0   & 341 & 341  &  & 4   & 527 & 531  &  & 0   & 529 & 529  &  & 0   & 531 & 531  \\ \midrule
\multirow{3}{*}{All}   & SIFT~\cite{lowe2004distinctive}      & 574 & 426 & 1000 &  & 378 & 622 & 1000 &  & 450 & 550 & 1000 &  & 460 & 540 & 1000 &  & 104 & 896 & 1000 \\
                       & D2-Net~\cite{Dusmanu2019CVPR}        & 347 & 653 & 1000 &  & 102 & 898 & 1000 &  & 201 & 799 & 1000 &  & 136 & 864 & 1000 &  & 4   & 996 & 1000 \\
                       & SuperPoint~\cite{detone18superpoint} & 495 & 505 & 1000 &  & 319 & 681 & 1000 &  & 317 & 683 & 1000 &  & 284 & 716 & 1000 &  & 33  & 967 & 1000 \\
                        \bottomrule 
\end{tabularx}%
\end{center}
\vspace{-5pt}
\caption{\textbf{Number of pairs for which SIFT, D2-Net, and SuperPoint return an answer. }  For each dataset and overlap ratio, we report the number of pairs for which RANSAC successfully outputs a model (regardless of whether that model is accurate or not).
}
\label{tab:sift_result}
\end{table*}

\begin{figure*} 
\begin{minipage}{\linewidth}
    \centering
    \jsubfig{\includegraphics[height=0.40cm]{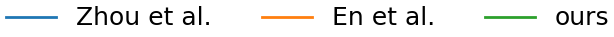}}{} \\    
    \jsubfig{\includegraphics[height=4.10cm]{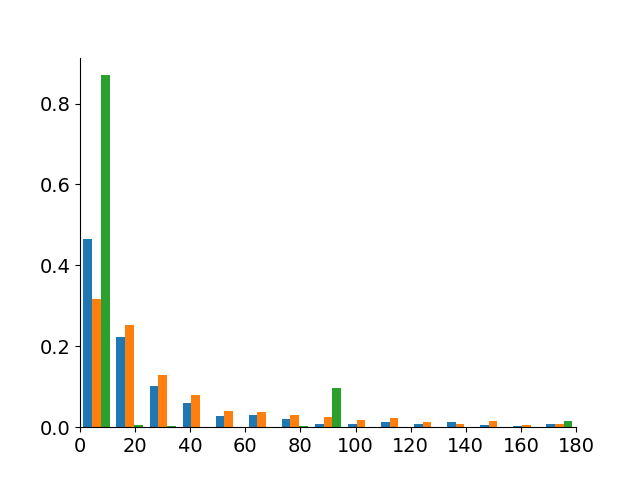}}{InteriorNet}
    \hfill
    \jsubfig{\includegraphics[height=4.10cm]{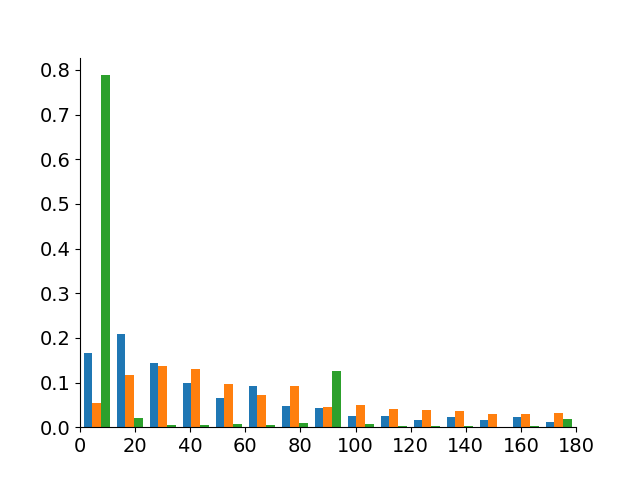}}{SUN360}
    \hfill
    \jsubfig{\includegraphics[height=4.10cm]{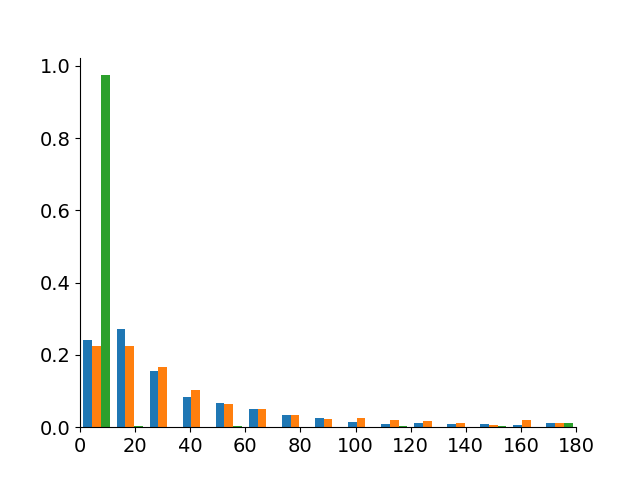}}{StreetLearn} \\
    \jsubfig{\includegraphics[height=4.10cm]{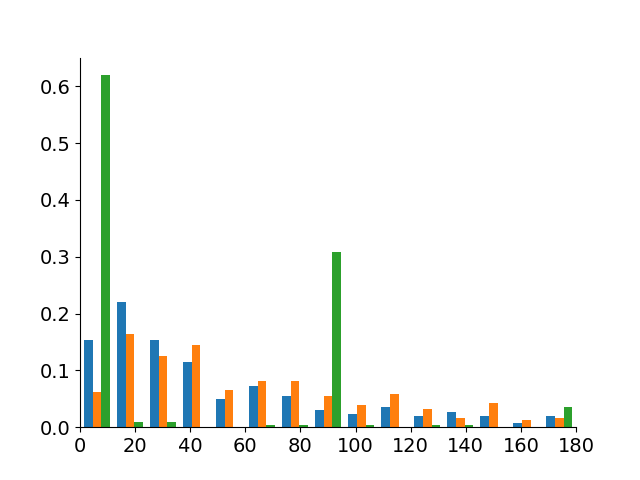}}{InteriorNet}
    \hfill
    \jsubfig{\includegraphics[height=4.10cm]{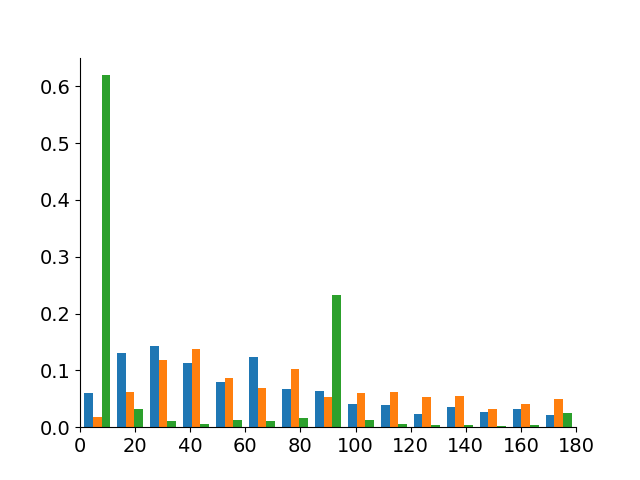}}{SUN360}
    \hfill
    \jsubfig{\includegraphics[height=4.10cm]{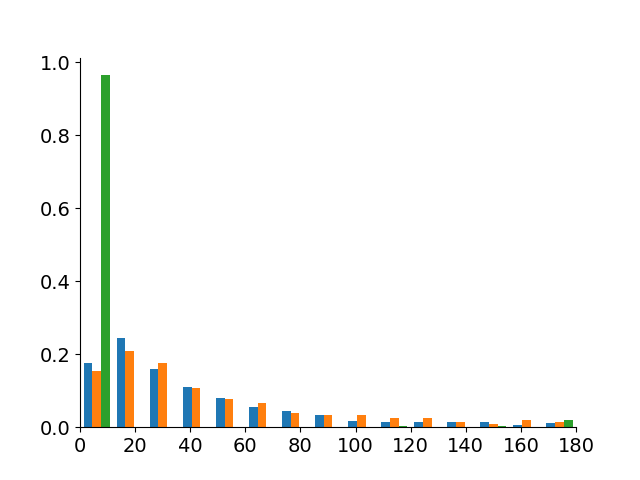}}{StreetLearn}
\vspace{5pt}
    \caption{\textbf{Geodesic error histograms.} The $x$-axis is geodesic error (each bin covers 10\degree, \emph{e.g.} the first bin is over errors in the range $[0\degree, 10\degree]$, and so on). The $y$-axis is the frequency of each bin. The top row shows histograms over all pairs and the bottom row is over non-overlapping pairs.
   }
    \label{fig:hist}
\end{minipage}
\begin{minipage}{\linewidth}
    \centering
    \jsubfig{\includegraphics[height=0.40cm]{figure/analysis_figures/fig2/legend.png}}{} \\    
    \jsubfig{\includegraphics[height=4.10cm]{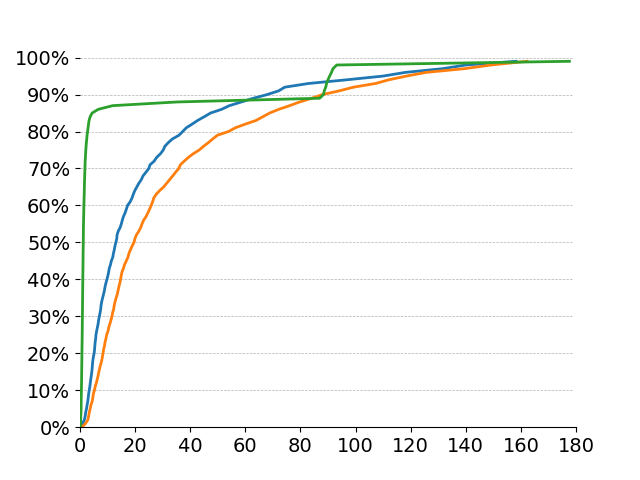}}{InteriorNet}
    \hfill
    \jsubfig{\includegraphics[height=4.10cm]{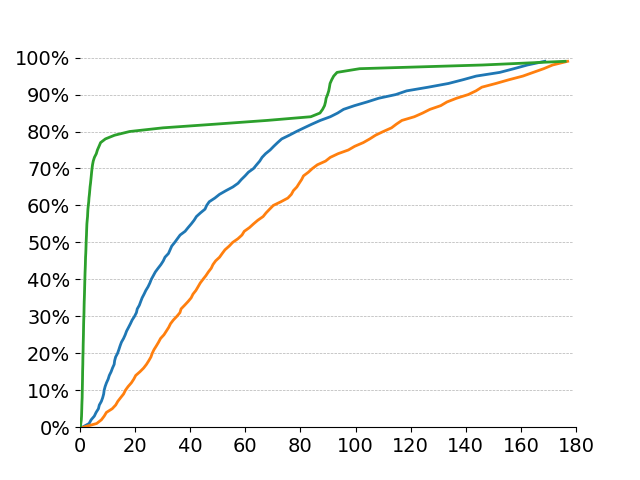}}{SUN360}
    \hfill
    \jsubfig{\includegraphics[height=4.10cm]{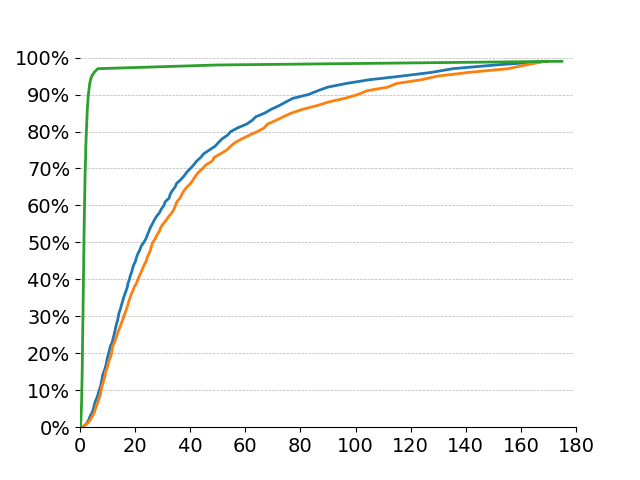}}{StreetLearn}\\
    \jsubfig{\includegraphics[height=4.10cm]{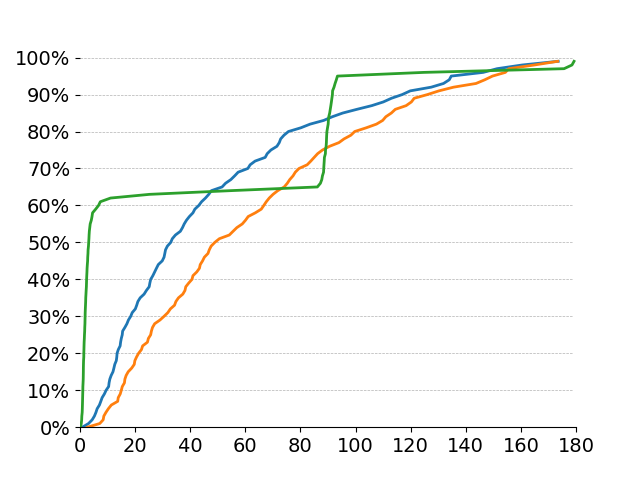}}{InteriorNet}
    \hfill
    \jsubfig{\includegraphics[height=4.10cm]{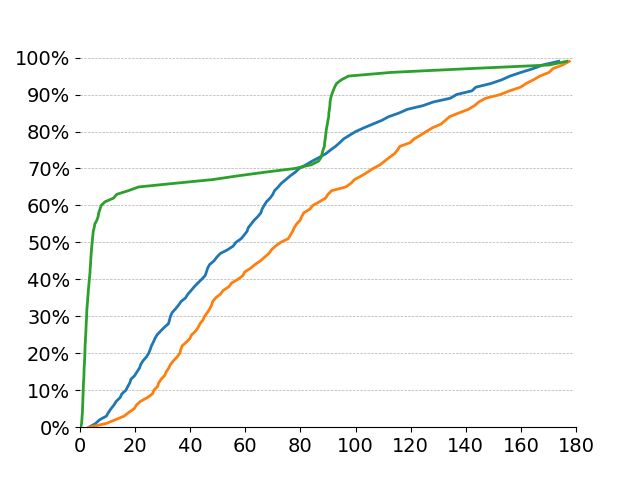}}{SUN360}
    \hfill
    \jsubfig{\includegraphics[height=4.10cm]{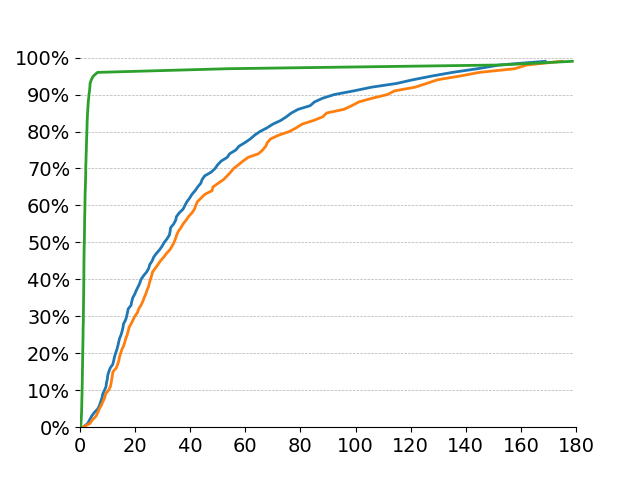}}{StreetLearn}
\vspace{5pt}
    \caption{\textbf{Cumulative distribution of geodesic error.} The $x$-axis is the geodesic error in degrees, and the $y$-axis is the percentage of pairs that achieve at most that given error. The top row shows the distribution over all pairs and the bottom row shows the distribution over non-overlapping pairs.}
    \label{fig:percentile}
\end{minipage}
\end{figure*}
\begin{figure*} 
    \centering
    \jsubfig{\includegraphics[height=4.10cm]{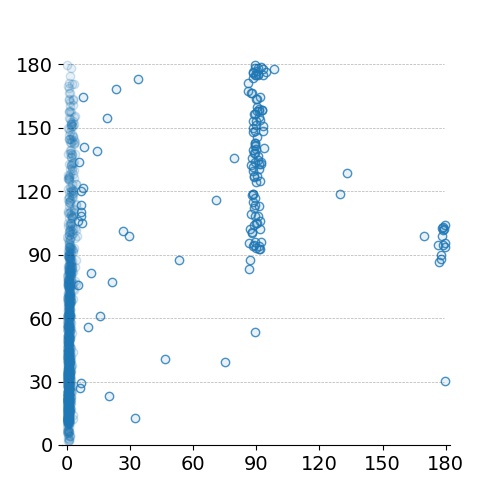}}{InteriorNet}
    \hfill
    \jsubfig{\includegraphics[height=4.10cm]{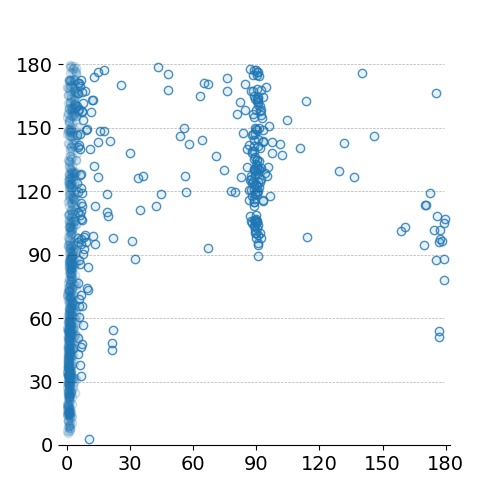}}{SUN360}
    \hfill
    \jsubfig{\includegraphics[height=4.10cm]{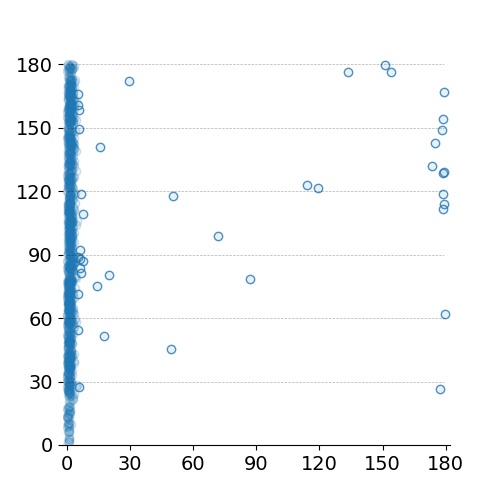}}{StreetLearn}
    \hfill
    \jsubfig{\includegraphics[height=4.10cm]{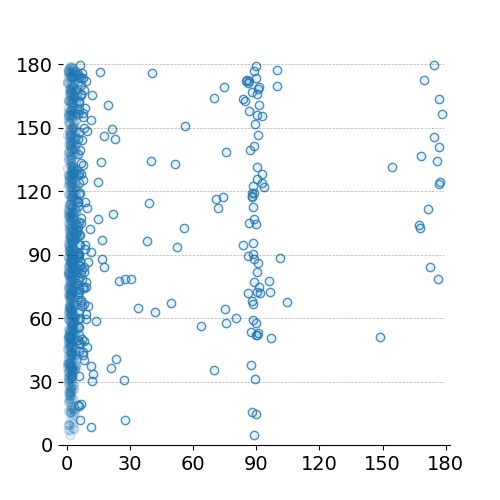}}{StreetLearn-T}
\vspace{5pt}
    \caption{\textbf{Geodesic error distribution.} 
    The $x$-axis is geodesic error in degrees, and the $y$-axis is the geodesic angle of the ground truth rotation of that pair. 
    Note the strong mode around 0$\degree$, as well as a weaker mode around 90$\degree$ for SUN360 and StreetLearn-T, and a small mode around 180$\degree$ for each dataset. These modes only manifest for non-overlapping pairs, and indicate ambiguities between rotations that differ by a multiple of 90$\degree$.
    }
    \label{fig:angle_error}
\end{figure*}

\begin{table*}[]
\centering
\setlength{\tabcolsep}{3.5pt}
\begin{tabularx}{0.92\textwidth}{llccclccclccc} 
\toprule
\multicolumn{2}{l}{}                         & \multicolumn{3}{c}{Rotation Error}                                                                     &  & \multicolumn{3}{c}{Yaw Error}                                                                       &  & \multicolumn{3}{c}{Pitch Error}                                                                                  \\ \cline{3-5} \cline{7-9} \cline{11-13}
\%Overlap & Roll
& Avg(\degree$\downarrow$) & Med(\degree$\downarrow$) & \multicolumn{1}{c}{10\degree(\%$\uparrow$)} &  & Avg(\degree$\downarrow$) & Med(\degree$\downarrow$) & \multicolumn{1}{c}{10\degree(\%$\uparrow$)} &  & Avg(\degree$\downarrow$) & Med(\degree$\downarrow$) & \multicolumn{1}{c}{10\degree(\%$\uparrow$)}  
\\ \midrule
\multirow{2}{*}{Large}                        & $=0\degree$  & 2.27 & 1.07 & 99.41 &  & 1.86 & 0.67 & 99.41 &  & 0.70 & 0.53 & 100.00 \\
 & $<5\degree$ & 7.71 & 3.18 & 97.08 &  & 5.20 & 0.84 & 97.65 &  & 1.55 & 1.55 & 99.41  \\ \midrule
\multirow{2}{*}{Small}                        & $=0\degree$  & 2.66 & 1.41 & 98.34 &  & 2.12 & 0.87 & 98.67 &  & 0.87 & 0.61 & 99.67  \\
 & $<5\degree$ & 6.43 & 3.39 & 96.33 &  & 3.43 & 0.95 & 98.01 &  & 1.56 & 0.87 & 99.00  \\\midrule
\multirow{2}{*}{None}                        & $=0\degree$  & 6.48 & 1.58 & 96.60 &  & 5.70 & 0.92 & 96.79 &  & 1.18 & 0.59 & 99.24  \\
  & $<5\degree$ & 8.54 & 3.44 & 96.22 &  & 6.88 & 1.16 & 96.22 &  & 1.37 & 0.87 & 99.24  \\\midrule
\multirow{2}{*}{All}                        & $=0\degree$  & 4.61 & 1.42 & 97.60 &  & 3.97 & 0.85 & 97.80 &  & 1.00 & 0.59 & 99.50  \\
   & $<5\degree$ & 7.77 & 3.39 & 96.40 &  & 5.56 & 1.04 & 97.00 &  & 1.46 & 0.88 & 99.20 \\
                        \bottomrule              
\end{tabularx}%

\caption{\textbf{Roll angle experiments on StreetLearn}, evaluating the effect of adding small roll angles to the ground truth rotation on relative rotation estimates of image pairs at test time. As illustrated in the table, adding roll at test time does not break the model, but rather leads to a modest increase in up to $2\degree$ error (across all overlap levels).
}
\label{tab:roll_supp}
\end{table*}
\begin{figure*}
    \begin{center}
    \newcommand{\sizea}{0.2425\textwidth}
    \newcommand{\tal}{0cm}
    \newcommand{\tab}{0cm}
    \newcommand{\tar}{0cm}
    \newcommand{\tat}{0cm}
    \newcommand{\smb}{0cm}
    \newcommand{\T}[1]{\raisebox{-0.5\height}{#1}}
    \setlength{\tabcolsep}{0pt}
    \renewcommand{\arraystretch}{0}
    \begin{tabular}{@{}cccc@{}}
        \includegraphics[width=\sizea, trim={\tal} {\tab} {\tar} {\tat},clip]{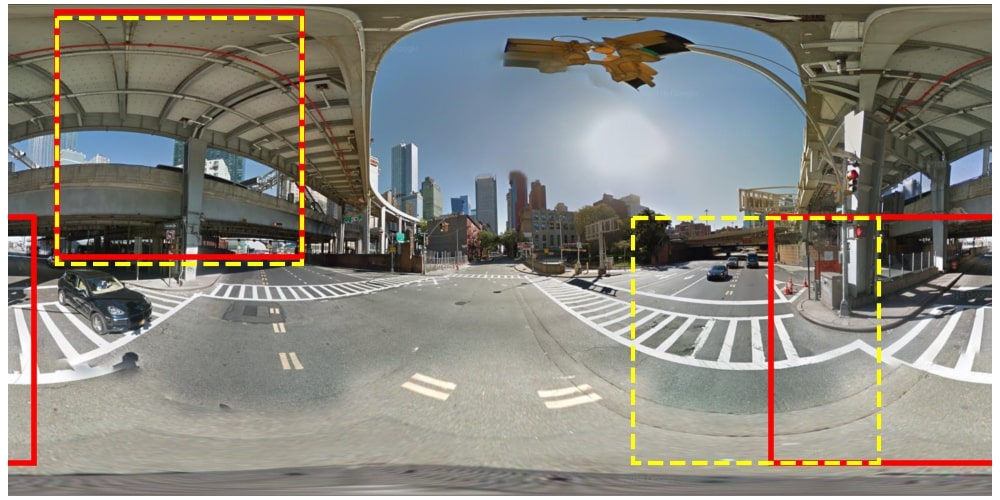} & 
        \includegraphics[width=\sizea, trim={\tal} {\tab} {\tar} {\tat},clip]{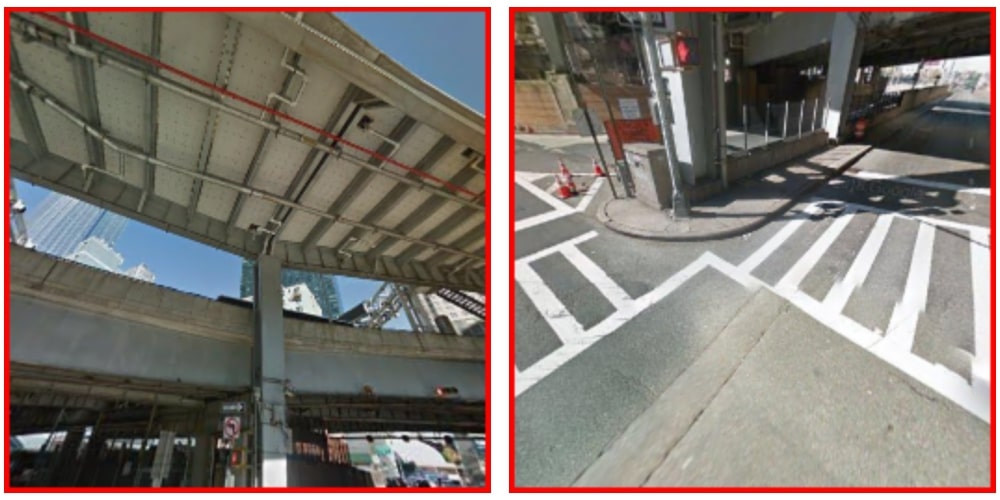} & 
        \includegraphics[width=\sizea, trim={\tal} {\tab} {\tar} {\tat},clip]{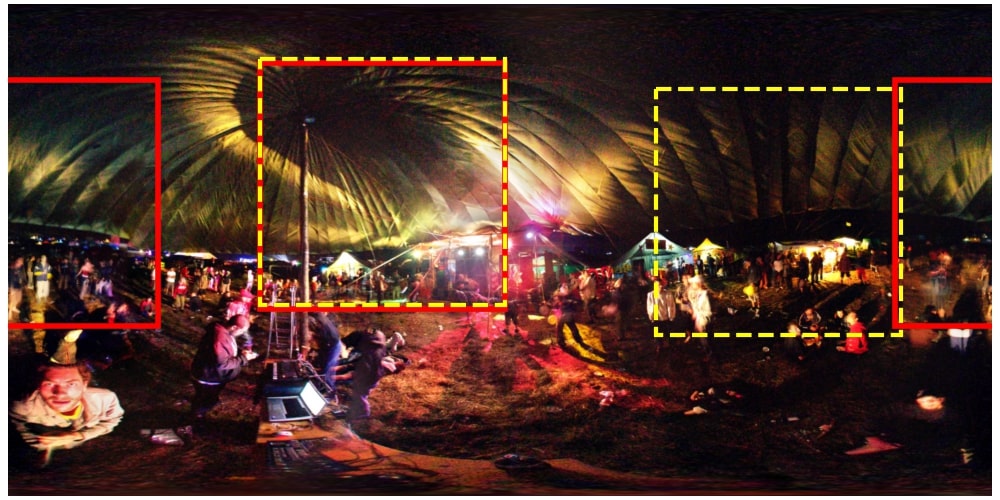}&
        \includegraphics[width=\sizea, trim={\tal} {\tab} {\tar} {\tat},clip]{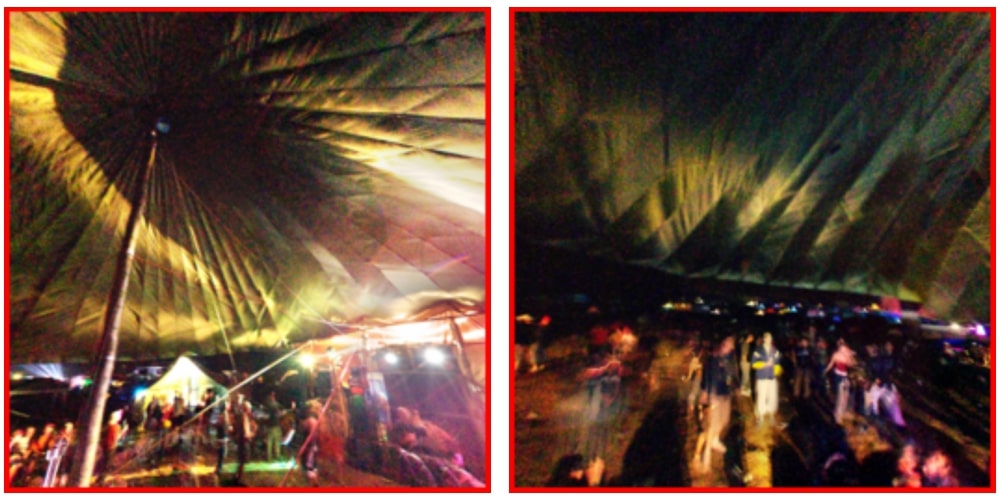}
        \\ 
        \includegraphics[width=\sizea, trim={\tal} {\tab} {\tar} {\tat},clip]{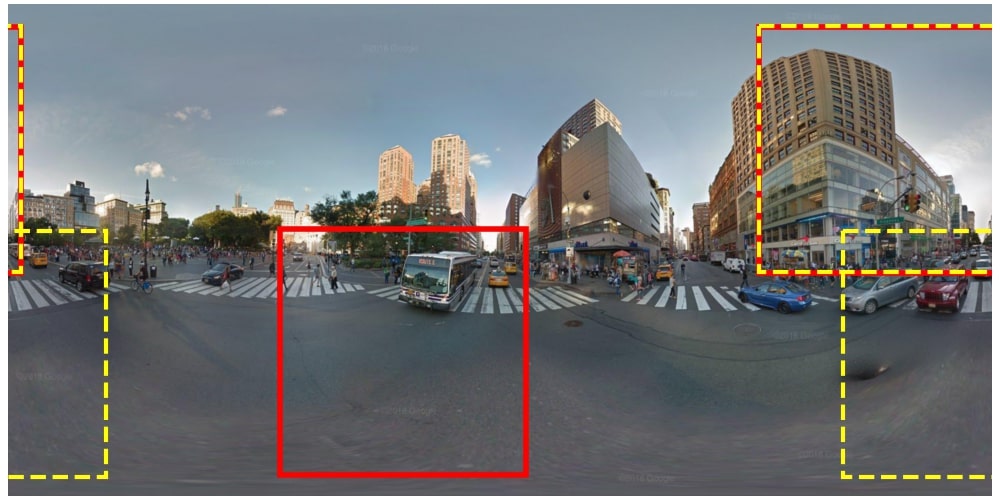} & 
        \includegraphics[width=\sizea, trim={\tal} {\tab} {\tar} {\tat},clip]{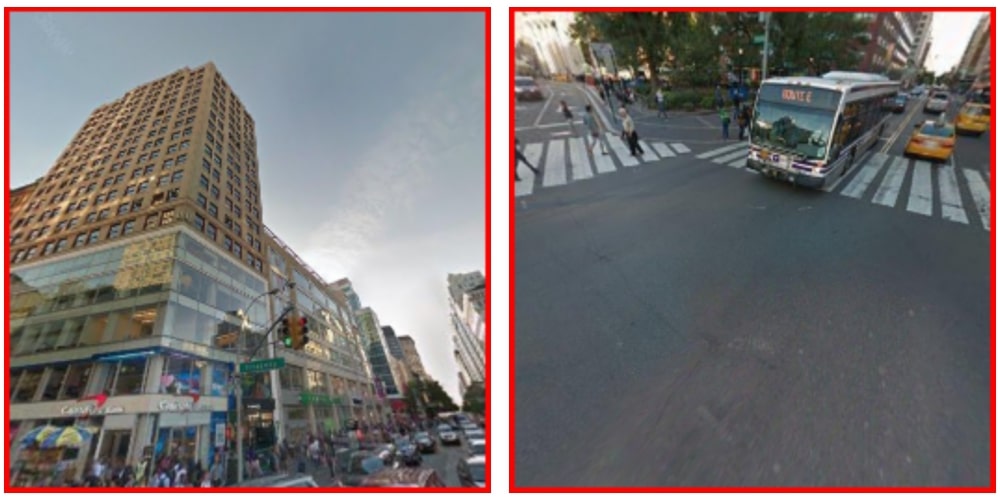} & 
        \includegraphics[width=\sizea, trim={\tal} {\tab} {\tar} {\tat},clip]{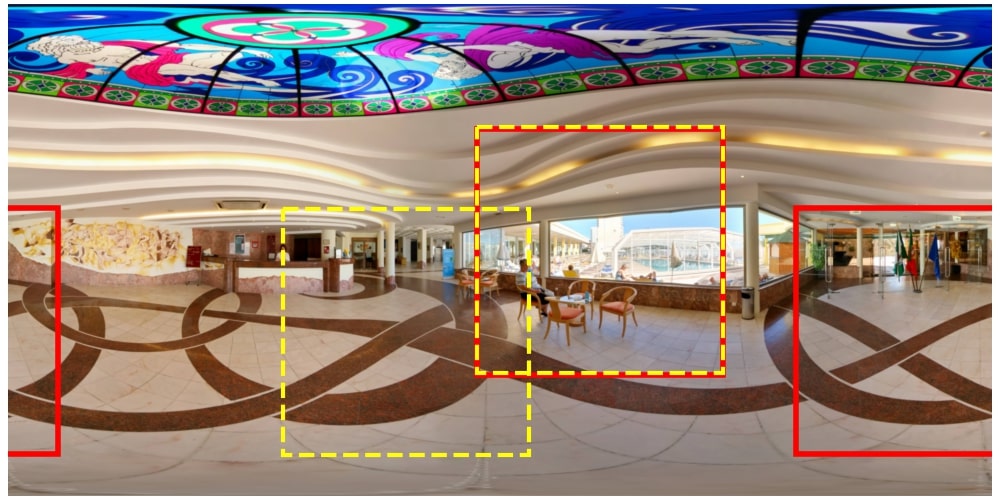}&
        \includegraphics[width=\sizea, trim={\tal} {\tab} {\tar} {\tat},clip]{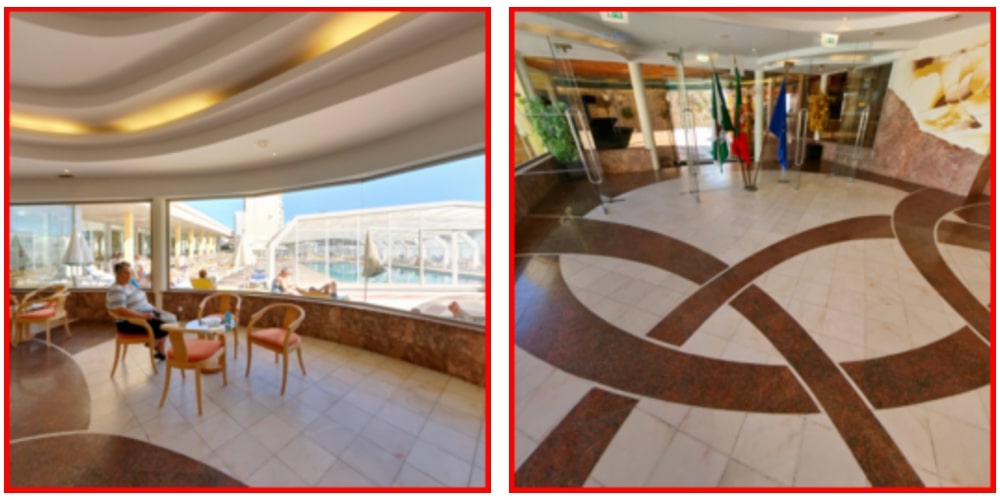}
        \\ 
        \includegraphics[width=\sizea, trim={\tal} {\tab} {\tar} {\tat},clip]{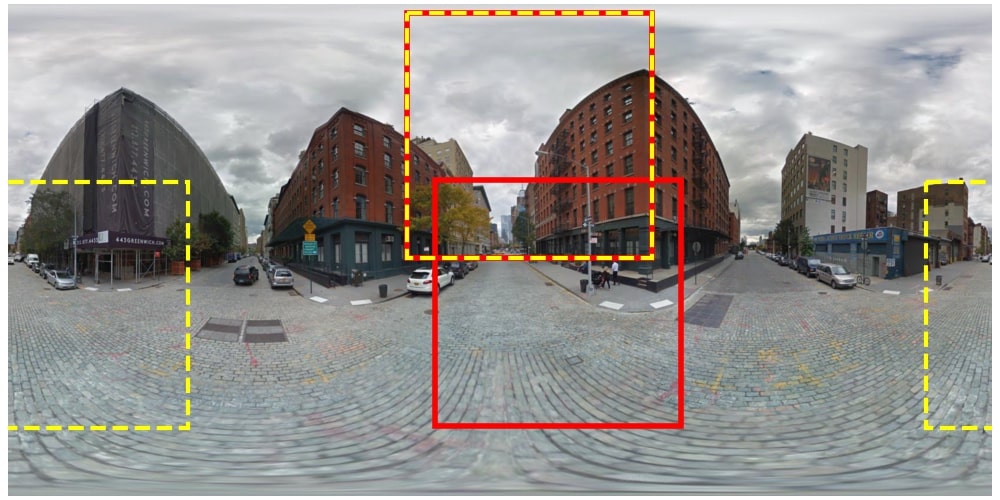} & 
        \includegraphics[width=\sizea, trim={\tal} {\tab} {\tar} {\tat},clip]{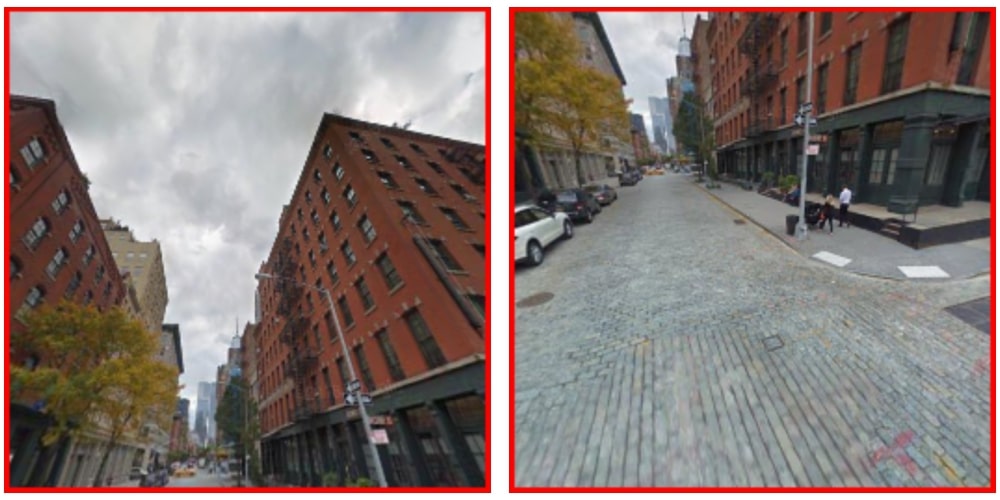} & 
        \includegraphics[width=\sizea, trim={\tal} {\tab} {\tar} {\tat},clip]{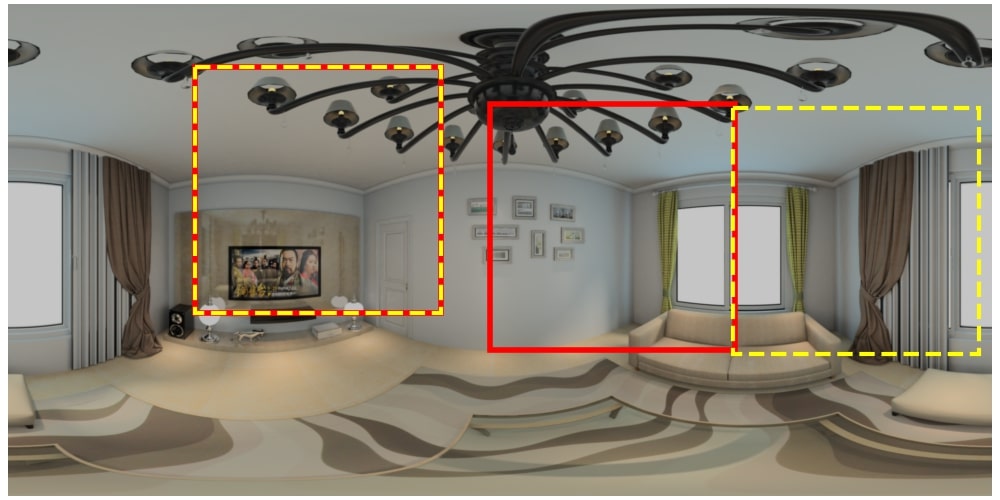}&
        \includegraphics[width=\sizea, trim={\tal} {\tab} {\tar} {\tat},clip]{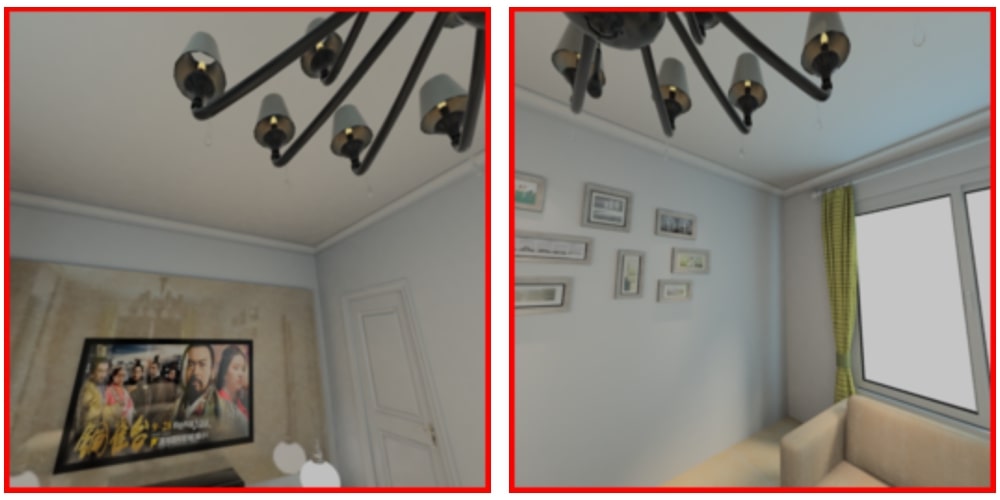}
        \\ 
        \includegraphics[width=\sizea, trim={\tal} {\tab} {\tar} {\tat},clip]{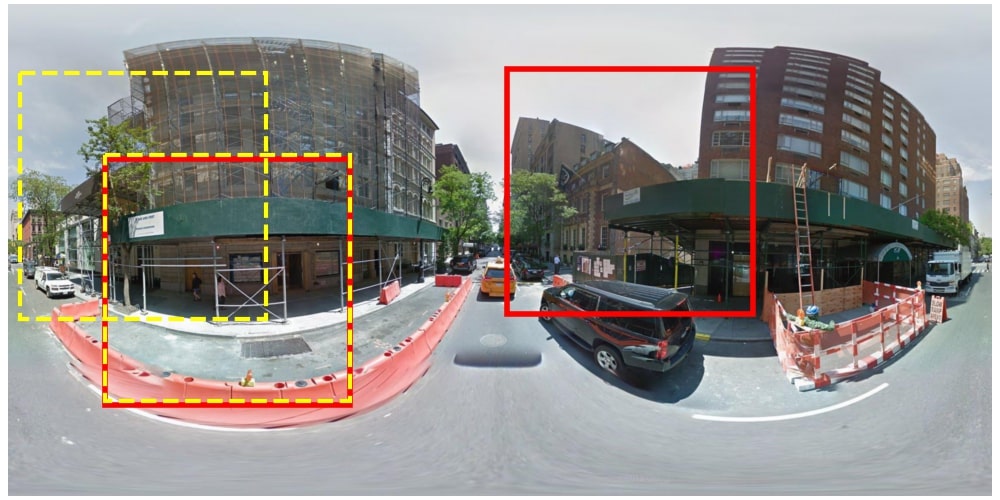} & 
        \includegraphics[width=\sizea, trim={\tal} {\tab} {\tar} {\tat},clip]{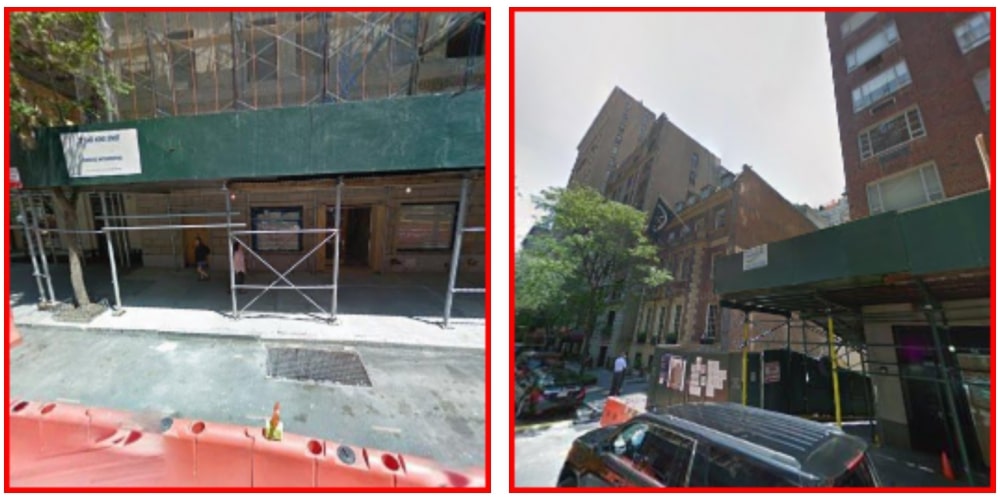}& 
        \includegraphics[width=\sizea, trim={\tal} {\tab} {\tar} {\tat},clip]{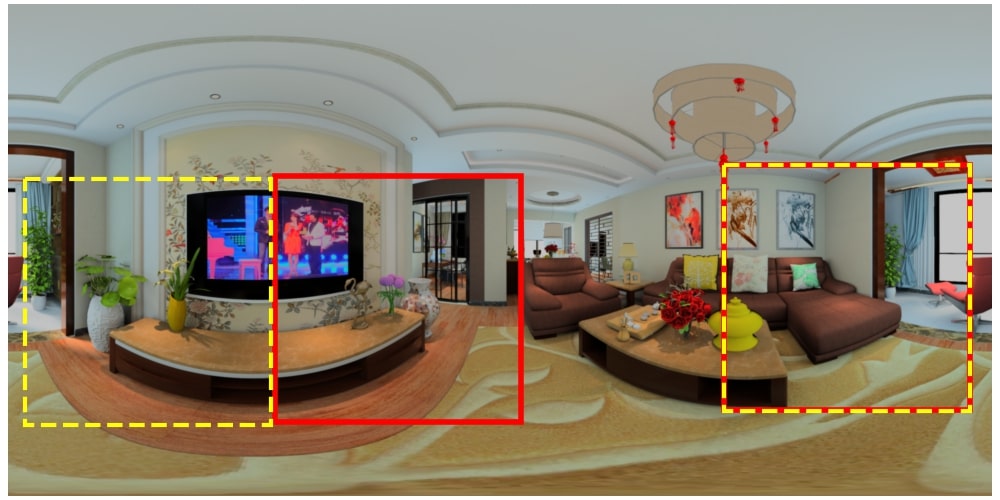}&
        \includegraphics[width=\sizea, trim={\tal} {\tab} {\tar} {\tat},clip]{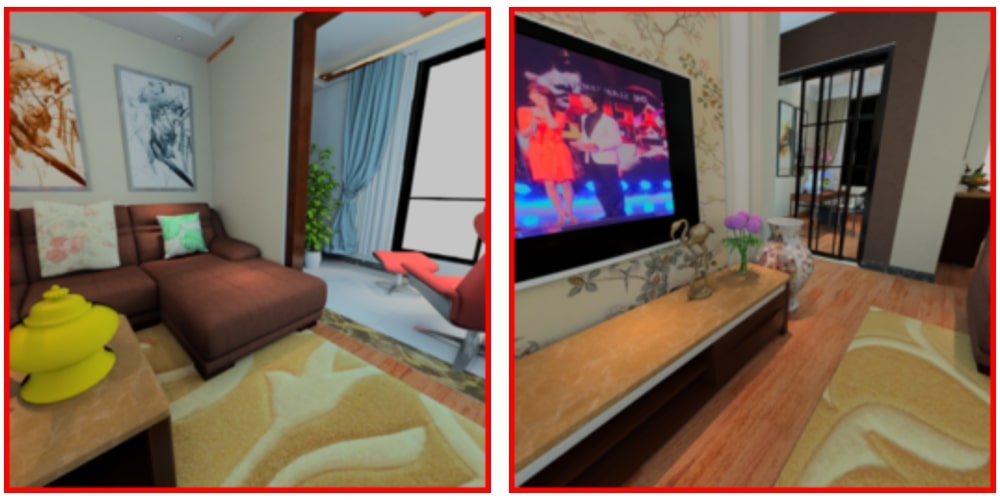}
        \\ 
        \multicolumn{2}{c}{StreetLearn} & \multicolumn{2}{c}{Indoor} 
    \end{tabular}
    \end{center}
    \caption{\textbf{Failure cases.} The full panoramas are shown on the left, with the ground-truth perspective images marked in red. We show our predicted viewpoints in yellow.
    }
    \label{fig:failure}
\end{figure*}
\begin{table}[]
\centering
\scriptsize
\setlength{\tabcolsep}{2pt}
\begin{tabularx}{0.92\columnwidth}{llccclccc} 
\toprule
 & \multicolumn{3}{c}{w/ Watermarks}                                                                     &  & \multicolumn{3}{c}{w/o Watermarks}                                                                                                                                                       \\ \cline{2-4} \cline{6-8}
\%Overlap
& Avg(\degree$\downarrow$) & Med(\degree$\downarrow$) & \multicolumn{1}{c}{10\degree(\%$\uparrow$)} &  & Avg(\degree$\downarrow$) & Med(\degree$\downarrow$) & \multicolumn{1}{c}{10\degree(\%$\uparrow$)} 
\\ \midrule
Large & 1.52         & 1.09         & 99.41    &  & 3.52         & 1.13         & 98.24    \\
Small & 3.23         & 1.41         & 98.34    &  & 4.65         & 1.54         & 97.67    \\
None  & 5.77         & 1.53         & 96.41    &  & 9.18         & 1.81         & 94.33   \\
All   & 4.40         & 1.44         & 97.50    &  & 6.86         & 1.59         & 96.00   \\
                        \bottomrule              
\end{tabularx}%
\vspace{-5pt}
\caption{\textbf{Watermark removal experiment on StreetLearn}, showing results on the original test set on the left (w/ Watermarks) and on the modified test set on the right (w/o Watermarks). 
This experiment further demonstrates that our model is not merely learning from spurious features like watermarks, but is leveraging a broader range of cues.
}
\label{tab:watermark_supp}
\end{table}


\subsection{Statistical analysis}
We conduct a statistical analysis of the geodesic error for the relative rotations estimated by different methods across a range of datasets.
As illustrated in the histograms shown in Figure \ref{fig:hist}, most errors made by our method fall in the first bin ($[0\degree,10\degree]$), and the frequency of errors below $10\degree$ achieved by our method is significantly larger than the equivalent number for the baselines. Figure \ref{fig:percentile} shows a cumulative distribution of angular errors for each method. That figure shows that approximately $80\%$ of the errors made by our method are below 5\degree.
From inspecting the histograms and cumulative distributions for the interior datasets (InteriorNet and SUN360), we observe a distinct peak of error at 90\degree, indicating that our method can sometimes make errors that are likely due to ambiguities in Manhattan-world scenes like rooms.
From Figure \ref{fig:angle_error}, we can see the errors at $90\degree$ and $180\degree$ mainly stem from non-overlapping image pairs (Figure \ref{fig:failure} that shows failure cases also demonstrates this).

\subsection{Roll angle experiments}
We conduct a study of the influence of small roll angles by feeding our models (trained without roll) with image pairs containing small roll angles. Results are reported in Table \ref{tab:roll_supp}. As illustrated in the table, the mean geodesic errors over pairs with up to $5\degree$ roll increase by up to $2\degree$ (across all overlap levels), demonstrating that adding roll at test time does not break the model, but rather leads to a modest increase in error.

\subsection{Pitch angle prediction from a single image}
We evaluate to what extent pitch angles can be estimated from a single image by inputting the same image twice into our framework, using the network trained on the StreetLearn dataset (without retraining). The mean error (over pitch angle) is $0.75\degree$, with almost all images having errors smaller than $10\degree$ ($99.7\%$).
This result suggests that pitch can be predicted from a single image. We should note that pitch angles are generally easier to reason about (as can be seen in the ablation study in the main paper). This experiment also shows that the model correctly predicts an identity rotation matrix (when the same image is fed to the network twice), with an average geodesic error of $0.62\degree$ over the StreetLearn dataset. 

\subsection{Removing watermarks from StreetLearn}
Panoramas belonging to the StreetLearn dataset contain small (and barely visible) watermarks with a copyright notice at 12 fixed locations. In the paper, we show that the model trained on StreetLearn can generalize to a different outdoor dataset---Holicity~\cite{zhou2020holicity}, capturing panoramas of London. To also show that it does not significantly affect the performance on StreetLearn, in Table \ref{tab:watermark_supp}, we report results on the StreetLearn test set for modified panoramas, for which we removed the watermarks using Photoshop's content-aware fill technique (as the locations are fixed, we could batch-process this procedure and perform it over the entire test set).

As the table illustrates, the median performance over watermark-free test images are affected in a minor way. The average errors show a modest increase, and there is a 2$\%$ drop on the $10\degree$ success rate for  non-overlapping cases. This suggests that watermarks serve as a (seemingly minor) additional cue that the network utilizes to estimate relative rotations, but that there are clearly other cues as performance remains strong without watermarks (even though watermarks were visible during training and only removed on test images, without retraining).

\section{Additional Qualitative Results}
\label{sec:qualitative}

We visualize some failure cases in Figure \ref{fig:failure}.
On the StreetLearn dataset, the large errors mainly stem from the ambiguities involving opposing (antipodal) street directions. 
For example, the top two rows show a street crossing. In both cases, the model correctly predicts the pitch angle, but is confused in predicting the correct direction at the intersection.
The remaining two rows show that when the image pair---while non-overlapping---has similar objects visible in the view (e.g., green scaffolding sheds in the last row), the model can erroneously infer that the two images are pointed in the same direction, and when the image pair has just a small overlap, as in the second to last row, the model may predict that they are non-overlapping pairs, and assign a large rotation angle (both of these cases have $\sim$180$\degree$ error).

On SUN360 and InteriorNet, the larger variety of indoor scene configurations makes the relative rotation task comparatively more difficult.
For example, the first two rows contain repetitive textures on the roof and floor, which confuse the model.
The last two rows illustrate ambiguities arising from non-overlapping pairs, where the possible horizontal rotation range is $[90^\circ, 270^\circ]$.
Even when we narrow down the choices, for example if both views observe a corner of the room, there still may exist three plausible rotations. 
Our results suggest that estimating rotations in indoor scenes is generally more difficult in comparison to outdoor scenes.

\begin{figure*} 
\begin{minipage}{0.49\linewidth}
    \centering
        \jsubfig{\includegraphics[height=2.06cm]{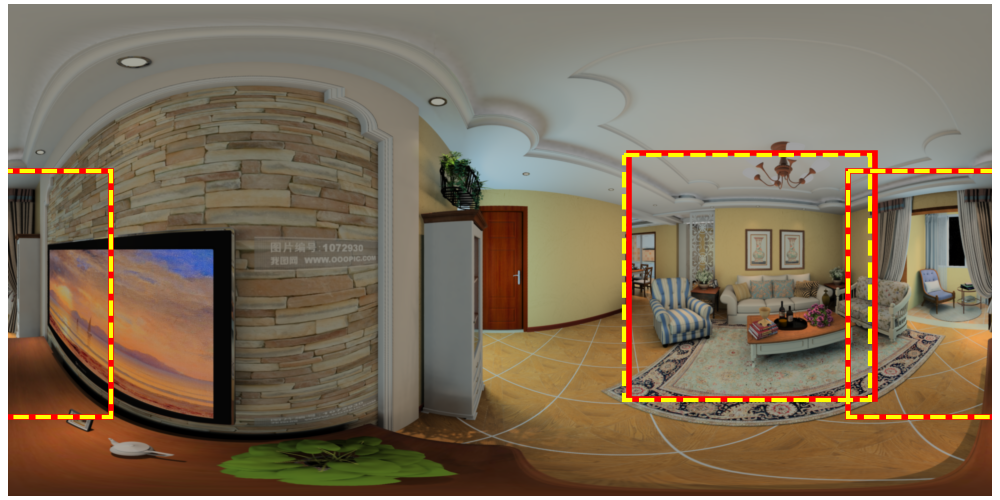}}{}
        \hfill
\jsubfig{\includegraphics[height=2.06cm]{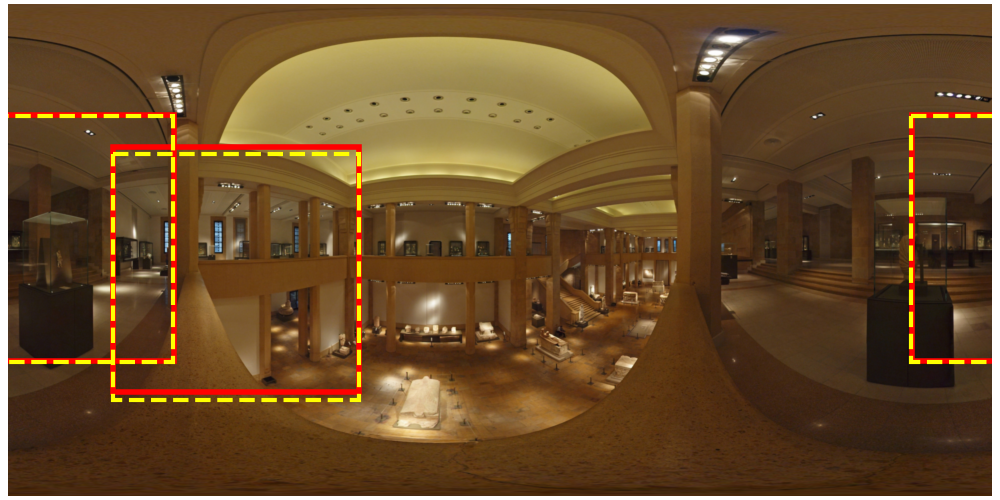}}{}
        \\ 
        \hfill
\jsubfig{\includegraphics[height=1.67cm]{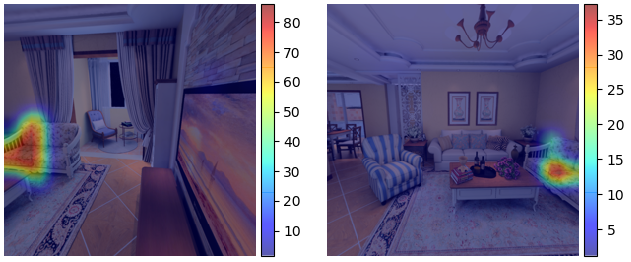}}{} 
        \hfill
\jsubfig{\includegraphics[height=1.63cm]{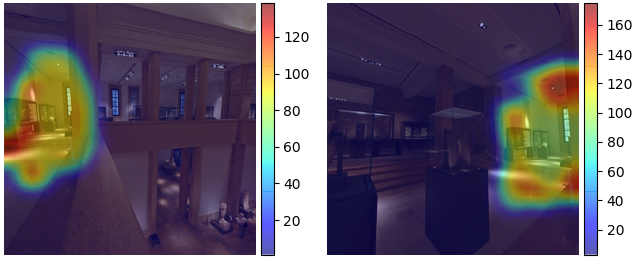}}{} 
\hfill
    \\ \vspace{7pt}
  \jsubfig{\includegraphics[height=2.06cm]{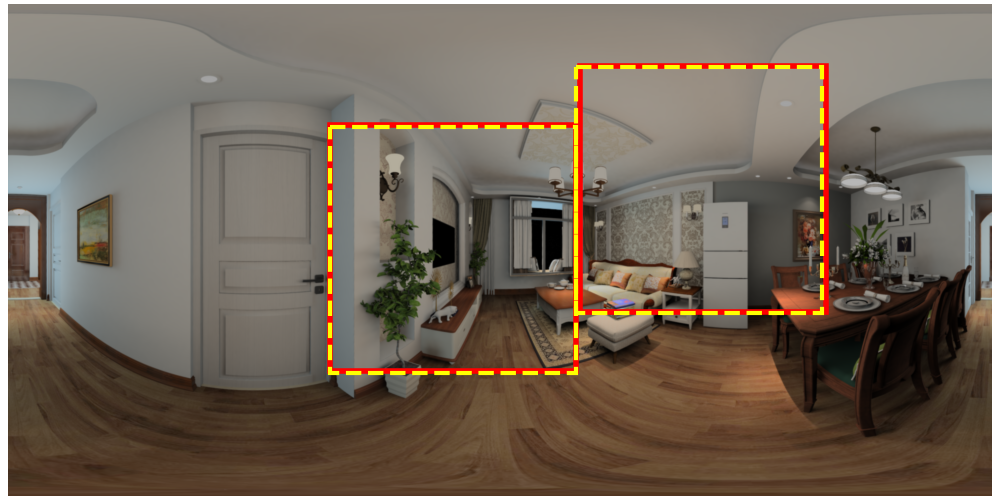}}{}
    \hfill
    \jsubfig{\includegraphics[height=2.06cm]{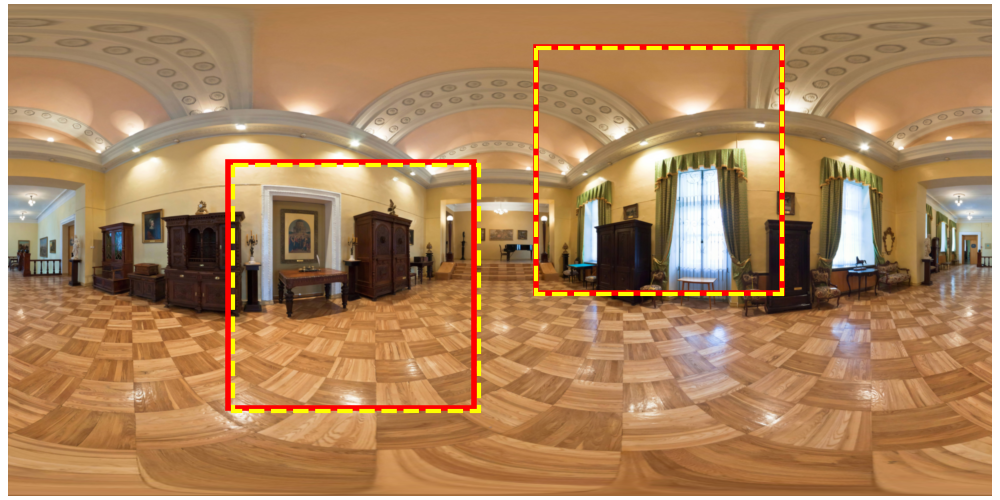}}{}
    \\ 
    \jsubfig{\includegraphics[height=1.7cm]{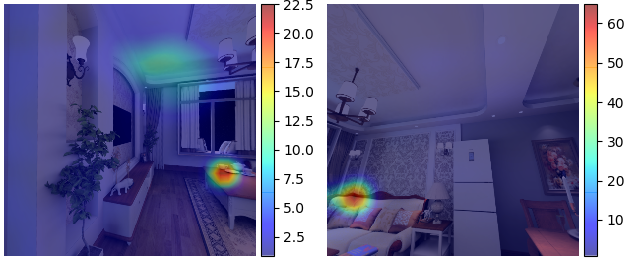}}{} 
    \hfill
    \jsubfig{\includegraphics[height=1.67cm,trim={8.5cm 0 0 0},clip]{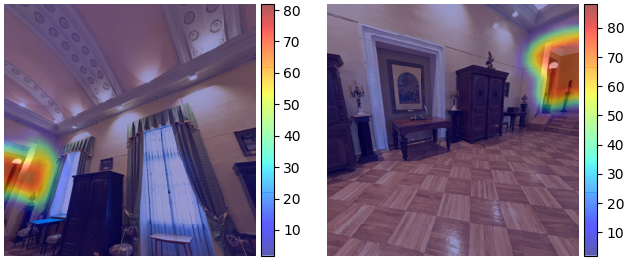}
    \includegraphics[height=1.67cm,trim={0 0 8.5cm 0},clip]{figure/clues_overview/new_cues/sun360_img_53-Copy.png}}{}   
\\ \vspace{7pt}
    \jsubfig{\includegraphics[height=2.06cm]{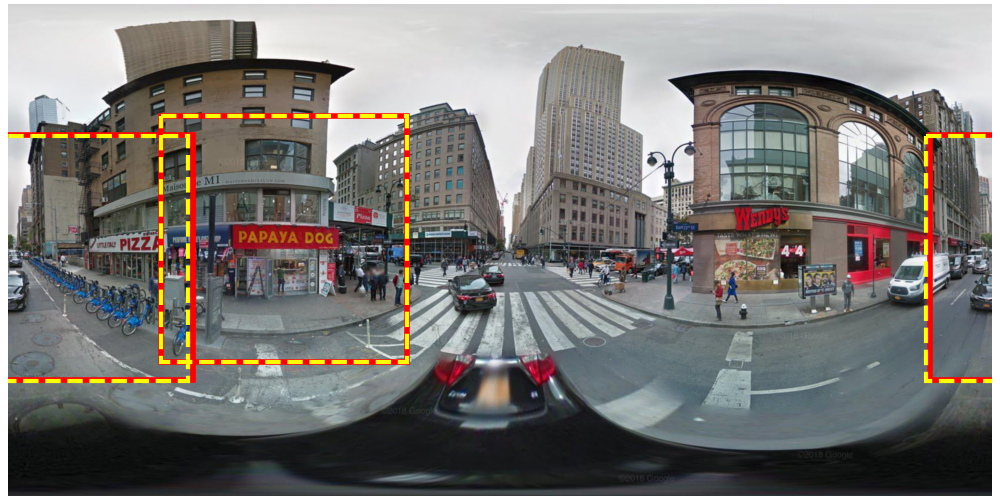}}{}
    \hfill 
\jsubfig{\includegraphics[height=2.06cm]{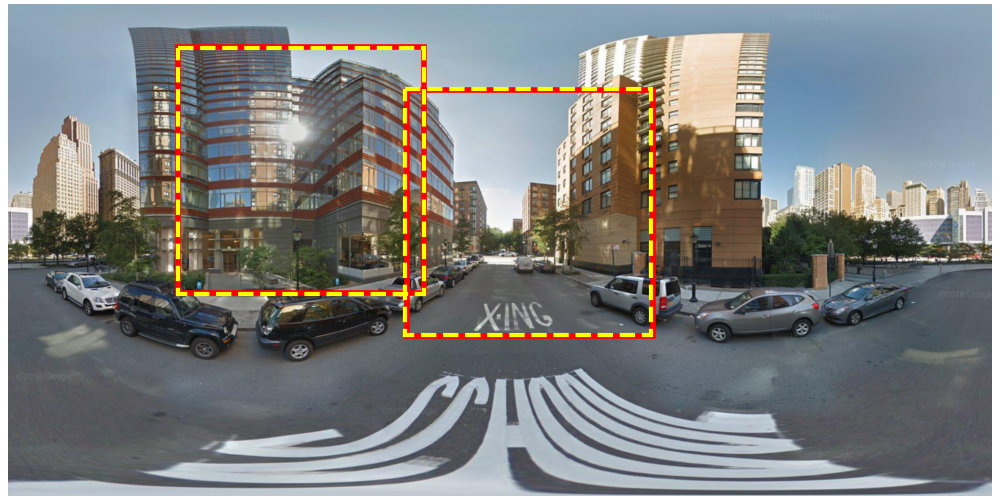}}{}
        \\ 
        \hfill
        \jsubfig{\includegraphics[height=1.67cm]{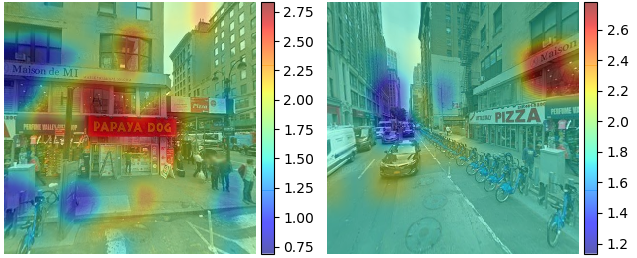}}{} 
        \hfill
\jsubfig{\includegraphics[height=1.67cm]{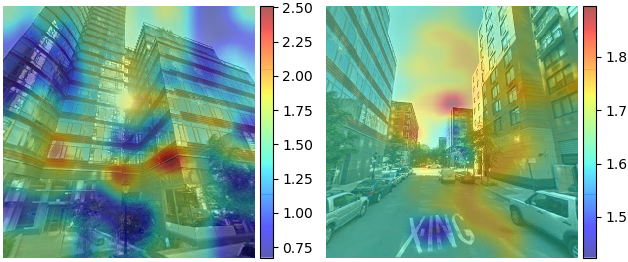}}{} 
\hfill
\\ \vspace{7pt}
    \jsubfig{\includegraphics[height=2.00cm]{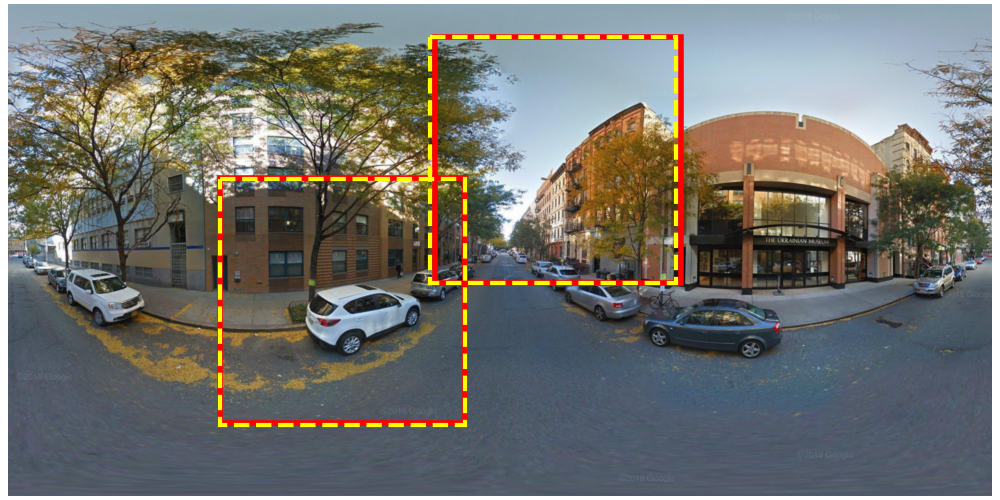}}{}
    \hfill 
\jsubfig{\includegraphics[height=2.00cm]{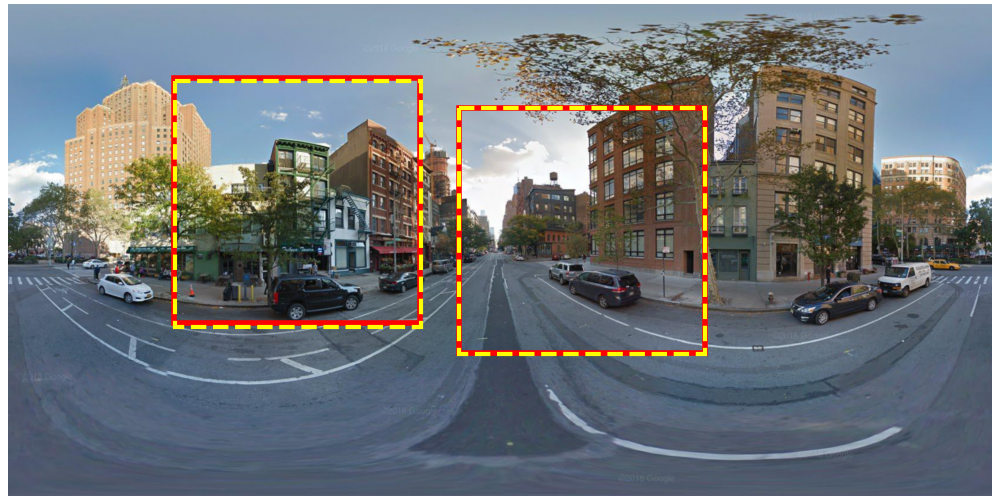}}{}
        \\ 
        \jsubfig{\includegraphics[height=1.67cm]{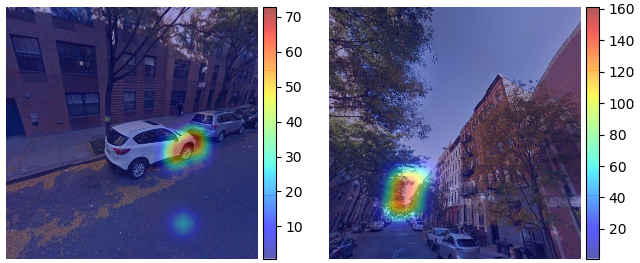}
        }{} 
        \hfill
\jsubfig{\includegraphics[height=1.67cm,trim={8.5cm 0 0 0},clip]{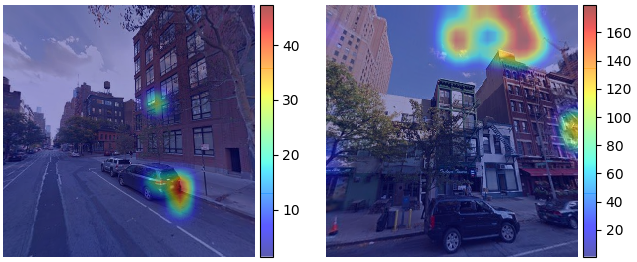}
\includegraphics[height=1.67cm,trim={0 0 8.5cm 0},clip]{figure/clues_overview/clues/st_wo_img_466.png}}{} 
    \vspace{7pt}
    \caption{\textbf{Visualizing cues detected by our model for overlapping and non-overlapping pairs.} We show regions which, when blocked, affect the rotation error, with warmer colors depicting larger errors (according to their associated color bars). The full panoramas are shown above, with the ground-truth and predicted perspective image regions marked in red and yellow, respectively. In indoor scenes, blocking corresponding regions (in overlapping pairs) may result in large errors (first row).  In outdoor scenes that are richer in information, blocking small regions does not seem to affect the model's predictions in overlapping cases (third row). In non-overlapping cases, we can see that the model seems to reason about truncated objects (second row, left) or cues related to vanishing points (bottom row, left), sunlight (bottom row, right) and shadows (second row, right).
   }
    \label{fig:clues_supp}
\end{minipage}
\hfill
\begin{minipage}{0.49\linewidth}
    \begin{center}
    \newcommand{\sizea}{0.475\textwidth}
    \newcommand{\tal}{0cm}
    \newcommand{\tab}{0cm}
    \newcommand{\tar}{0cm}
    \newcommand{\tat}{0cm}
    \newcommand{\smb}{0cm}
    \newcommand{\T}[1]{\raisebox{-0.5\height}{#1}}
    \setlength{\tabcolsep}{0pt}
    \renewcommand{\arraystretch}{0}
    \begin{tabular}{@{}ccc@{}}
        \multirow{3}{*}{\rotatebox[origin=c]{90}{Large \whitetxt{ssssssssssss}}} &
        \T{\includegraphics[width=\sizea, trim={\tal} {\tab} {\tar} {\tat},clip]{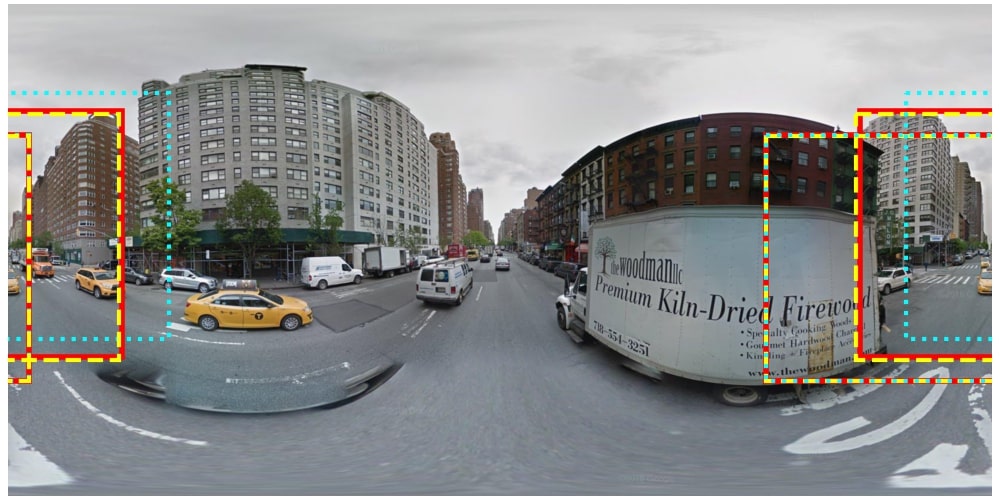}}& 
        \T{\includegraphics[width=\sizea, trim={\tal} {\tab} {\tar} {\tat},clip]{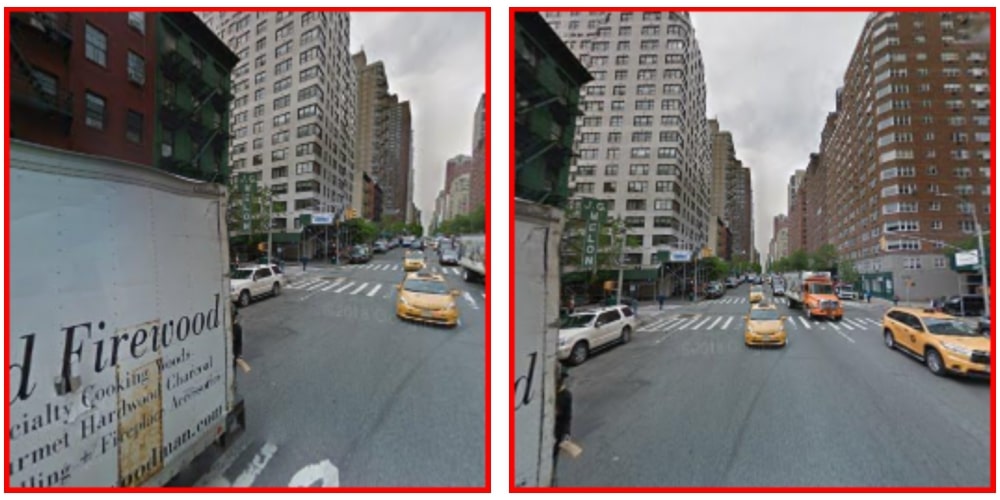}}
        \\ 
        &
        \T{\includegraphics[width=\sizea, trim={\tal} {\tab} {\tar} {\tat},clip]{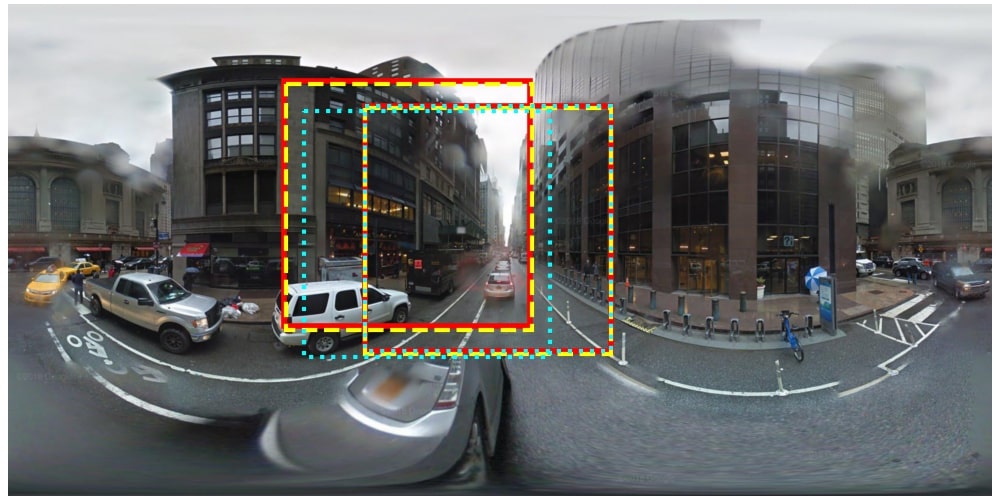}}& 
        \T{\includegraphics[width=\sizea, trim={\tal} {\tab} {\tar} {\tat},clip]{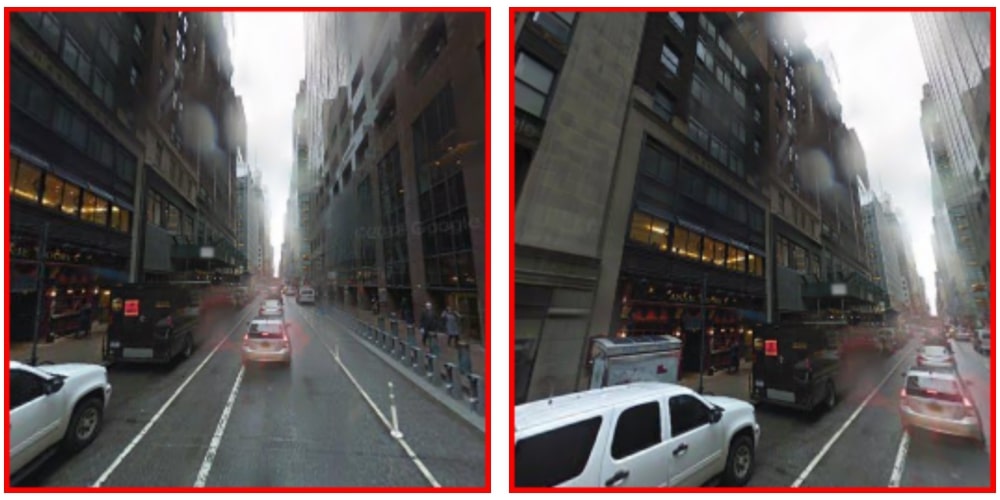}}
       \\ \vspace{+2pt}
       &
       \T{\includegraphics[width=\sizea, trim={\tal} {\tab} {\tar} {\tat},clip]{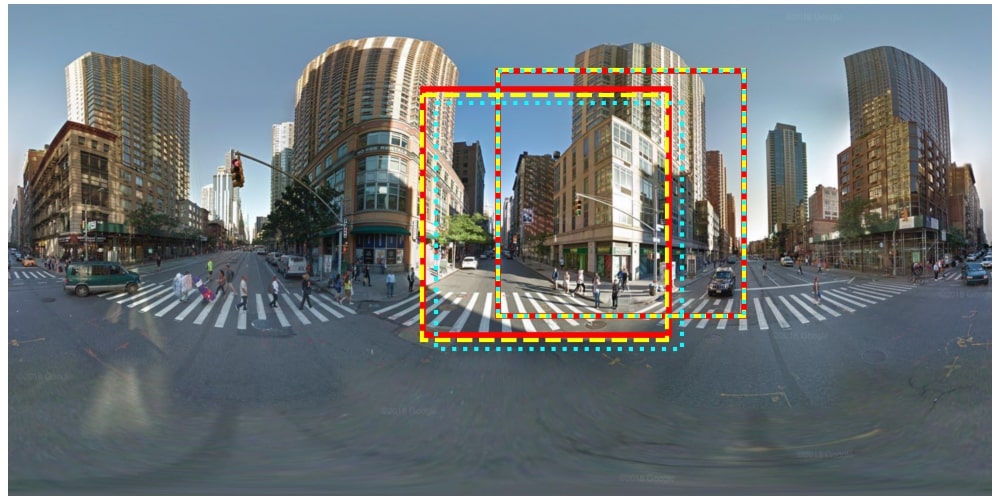}}& 
        \T{\includegraphics[width=\sizea, trim={\tal} {\tab} {\tar} {\tat},clip]{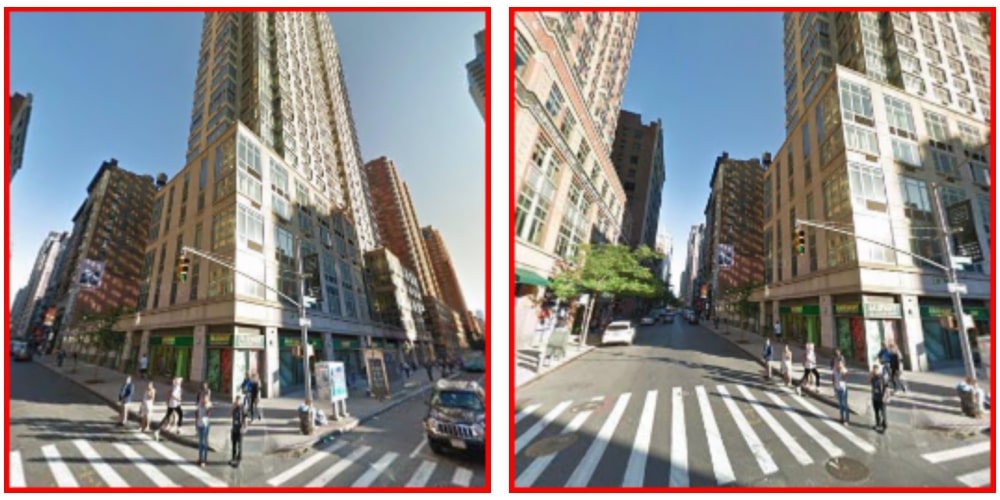}}
        \\ 
        
        \multirow{3}{*}{\rotatebox[origin=c]{90}{Small\whitetxt{ssssssssssss}}} &
        \T{\includegraphics[width=\sizea, trim={\tal} {\tab} {\tar} {\tat},clip]{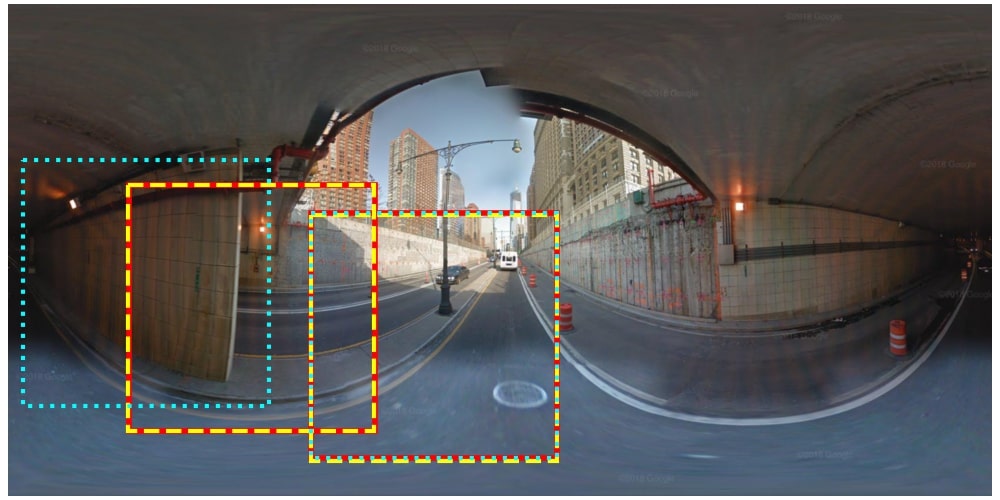}}& 
        \T{\includegraphics[width=\sizea, trim={\tal} {\tab} {\tar} {\tat},clip]{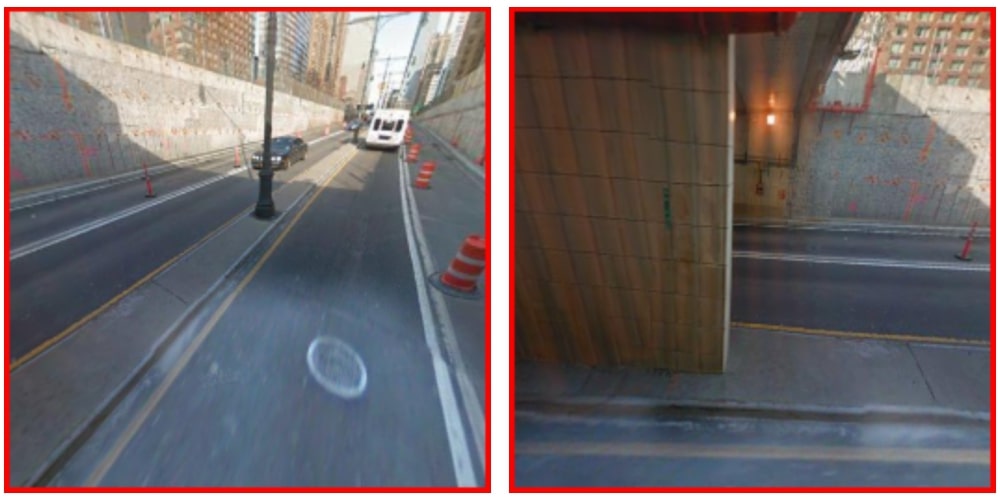}}
        \\ 
        &
        \T{\includegraphics[width=\sizea, trim={\tal} {\tab} {\tar} {\tat},clip]{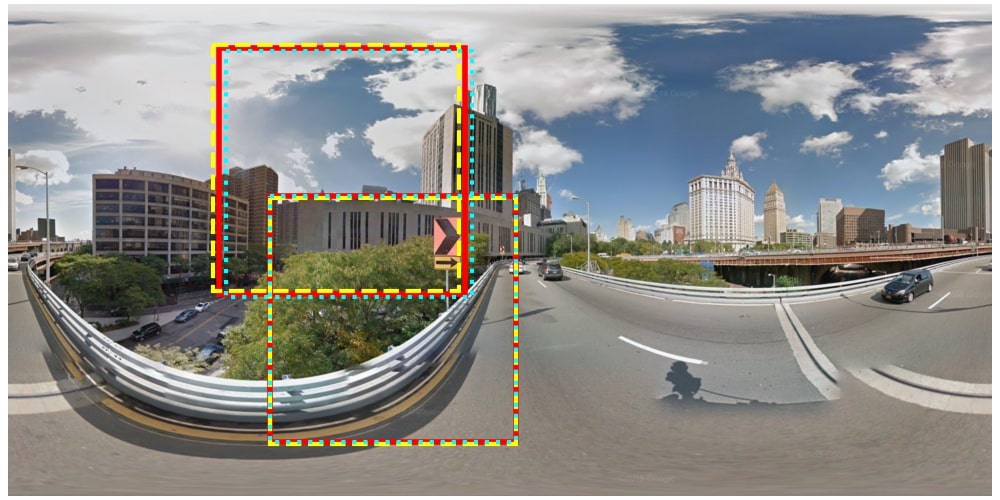}}& 
        \T{\includegraphics[width=\sizea, trim={\tal} {\tab} {\tar} {\tat},clip]{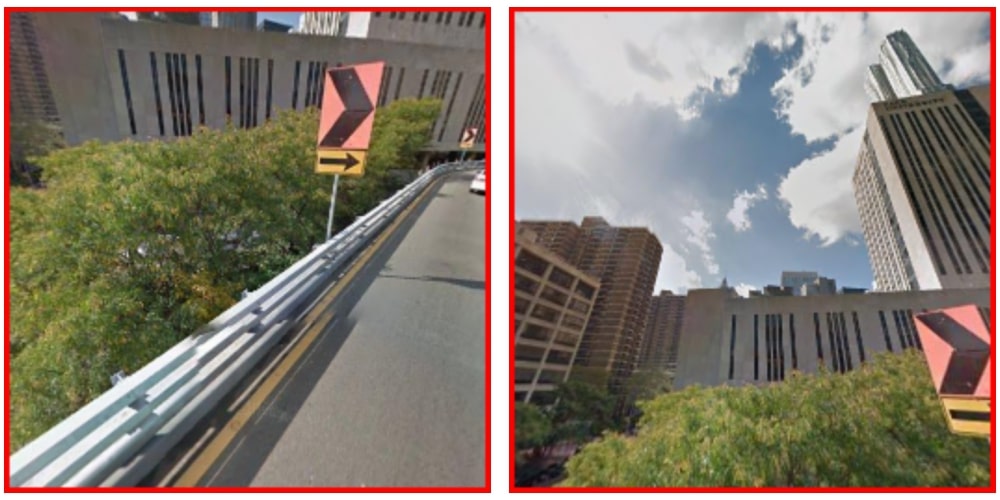}}
       \\ \vspace{+2pt}
       &
       \T{\includegraphics[width=\sizea, trim={\tal} {\tab} {\tar} {\tat},clip]{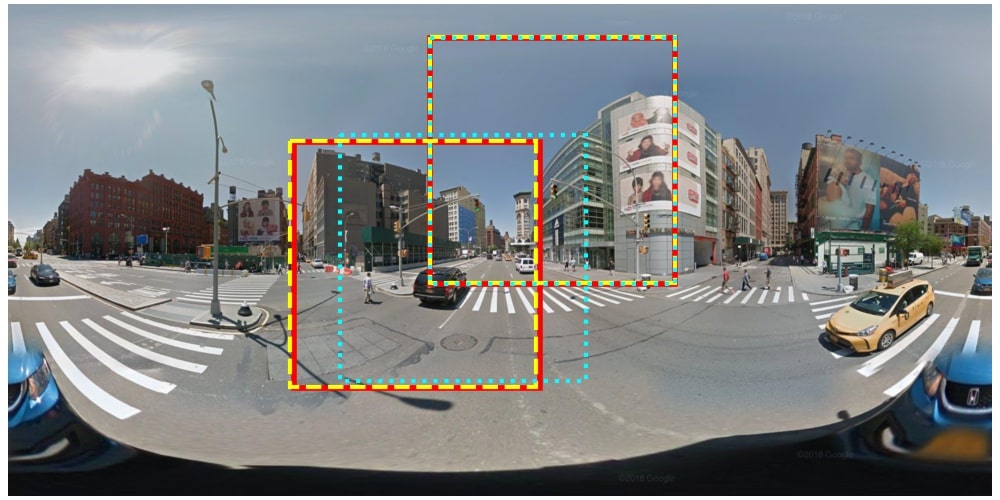}}& 
        \T{\includegraphics[width=\sizea, trim={\tal} {\tab} {\tar} {\tat},clip]{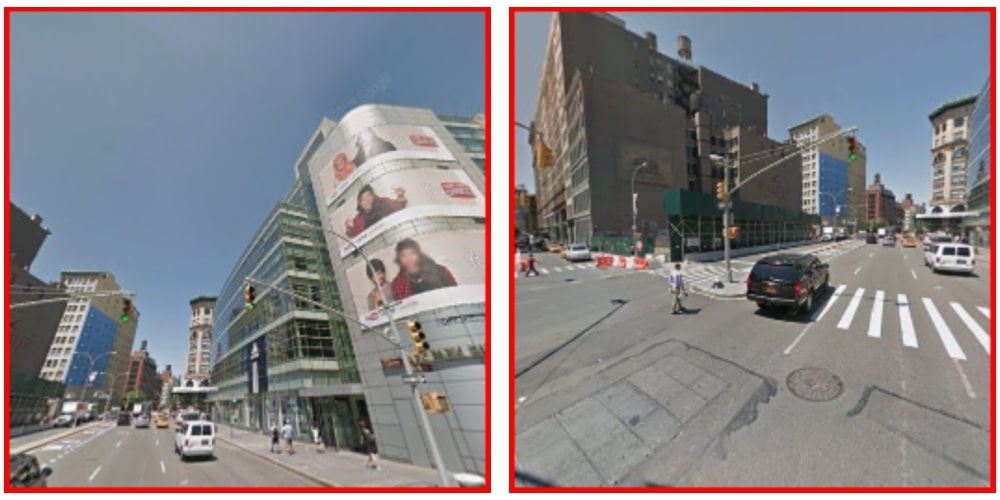}}
        
        \\
        \multirow{3}{*}{\rotatebox[origin=c]{90}{None\whitetxt{sssssssssssss}}} &
        \T{\includegraphics[width=\sizea, trim={\tal} {\tab} {\tar} {\tat},clip]{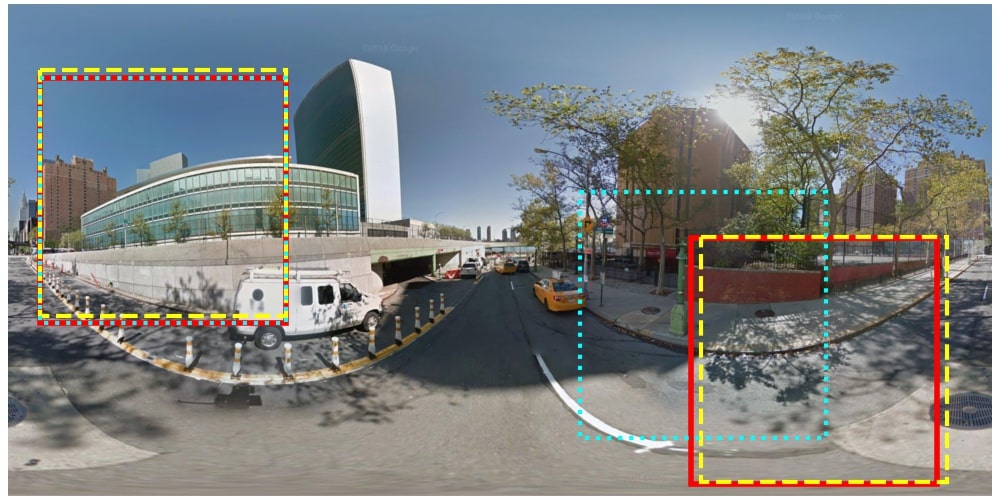}}& 
        \T{\includegraphics[width=\sizea, trim={\tal} {\tab} {\tar} {\tat},clip]{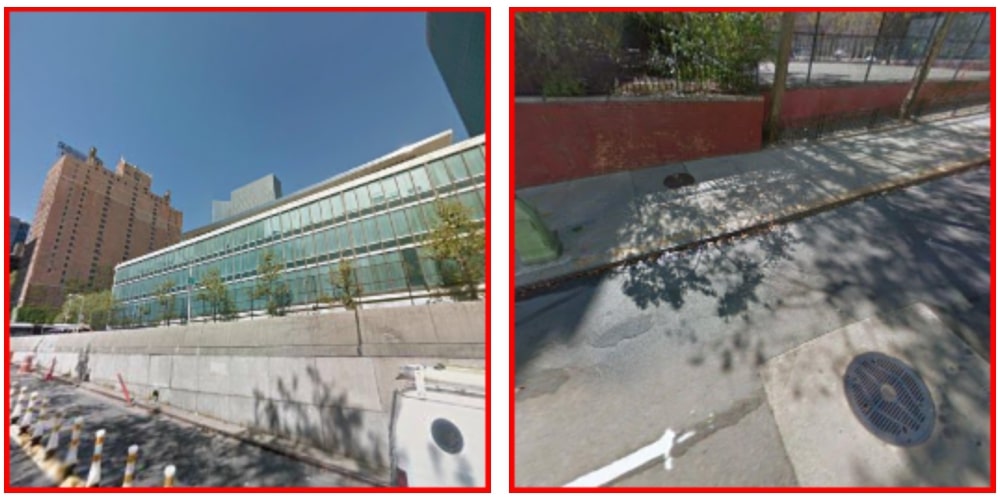}}
        \\ 
        &
        \T{\includegraphics[width=\sizea, trim={\tal} {\tab} {\tar} {\tat},clip]{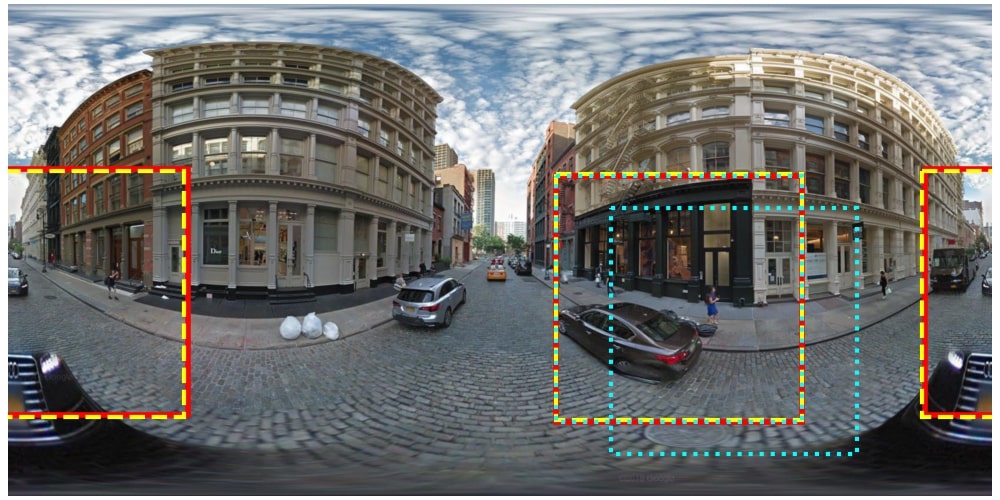}}& 
        \T{\includegraphics[width=\sizea, trim={\tal} {\tab} {\tar} {\tat},clip]{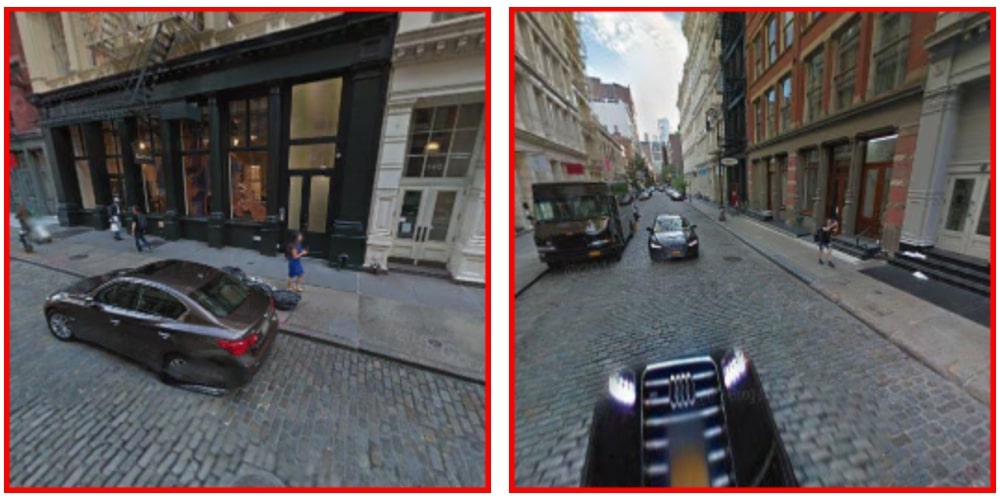}}
       \\ \vspace{+2pt}
       &
       \T{\includegraphics[width=\sizea, trim={\tal} {\tab} {\tar} {\tat},clip]{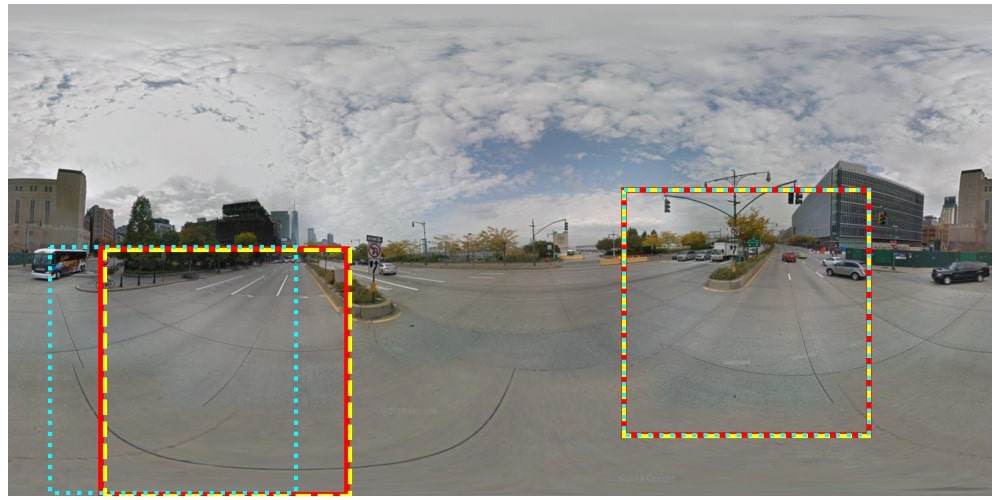}}& 
        \T{\includegraphics[width=\sizea, trim={\tal} {\tab} {\tar} {\tat},clip]{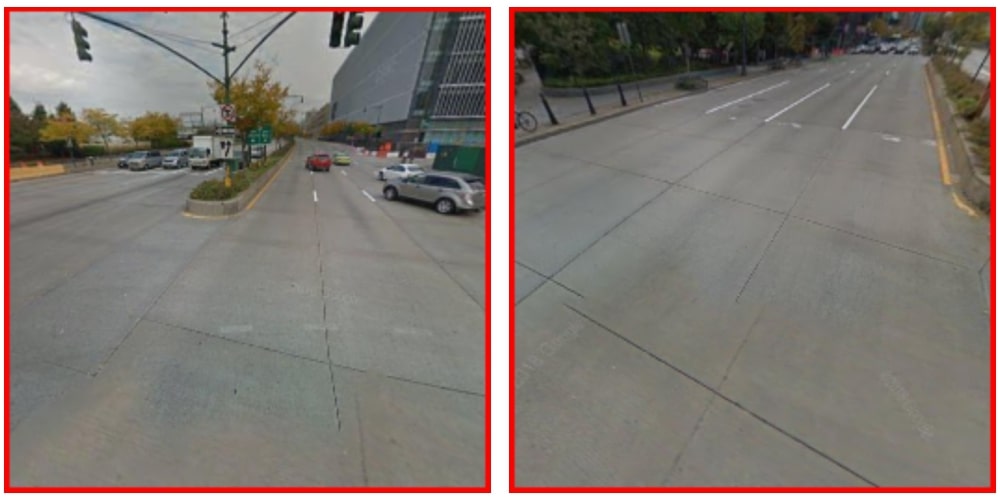}}
        \\ 
        & \multicolumn{2}{c}{StreetLearn} 
    \end{tabular}
    \end{center}
    \caption{\textbf{Predicted rotation results.} Full panoramas are shown on the left, with the ground-truth perspective images marked in red. We show our predicted viewpoints in yellow, and results obtained using the regression model of Zhou \etal~\cite{zhou2019continuity} in blue.
    }
    \label{fig:predicted_street}
\end{minipage}
\end{figure*}

\begin{figure*}
    \begin{center}
    \newcommand{\sizea}{0.2375\textwidth}
    \newcommand{\tal}{0cm}
    \newcommand{\tab}{0cm}
    \newcommand{\tar}{0cm}
    \newcommand{\tat}{0cm}
    \newcommand{\smb}{0.5cm}
    \newcommand{\T}[1]{\raisebox{-0.5\height}{#1}}
    \setlength{\tabcolsep}{0pt}
    \renewcommand{\arraystretch}{0}
    \begin{tabular}{@{}ccccc@{}}
        \multirow{3}{*}{\rotatebox[origin=c]{90}{Large \whitetxt{ssssssssssss}}} &
        \T{\includegraphics[width=\sizea, trim={\tal} {\tab} {\tar} {\tat},clip]{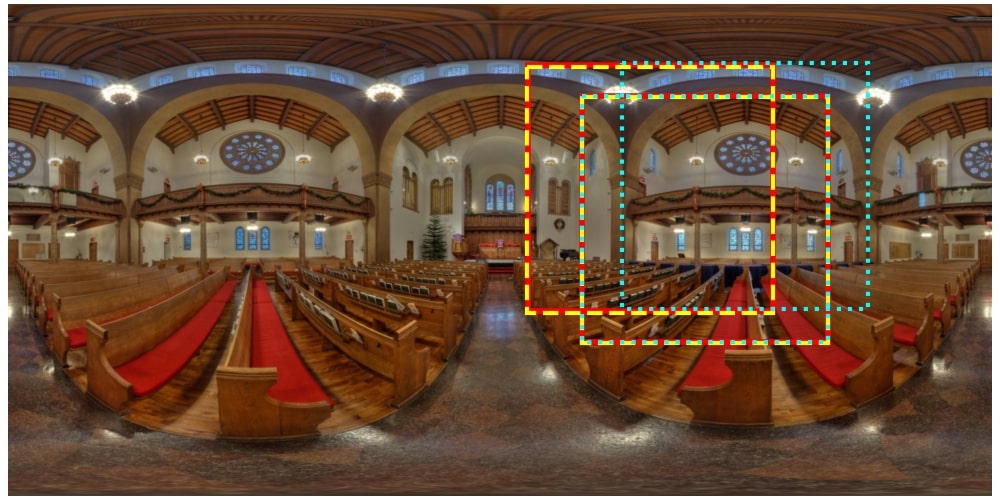}}&
        \T{\includegraphics[width=\sizea, trim={\tal} {\tab} {\tar} {\tat},clip]{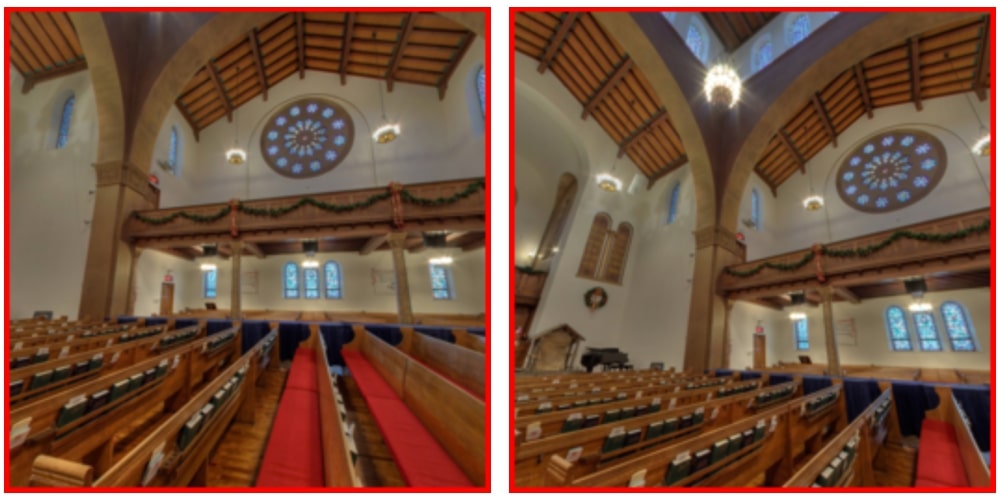}}&
        \T{\includegraphics[width=\sizea, trim={\tal} {\tab} {\tar} {\tat},clip]{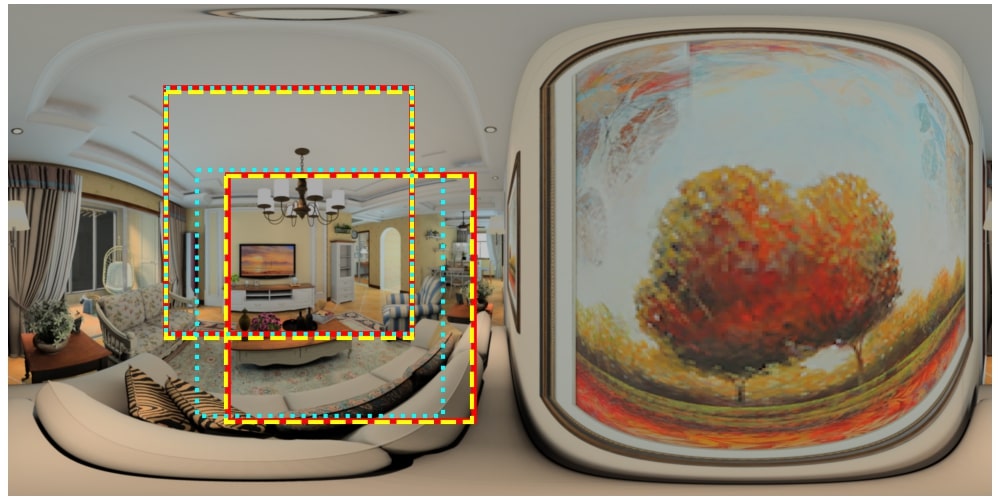}} &
        \T{\includegraphics[width=\sizea, trim={\tal} {\tab} {\tar} {\tat},clip]{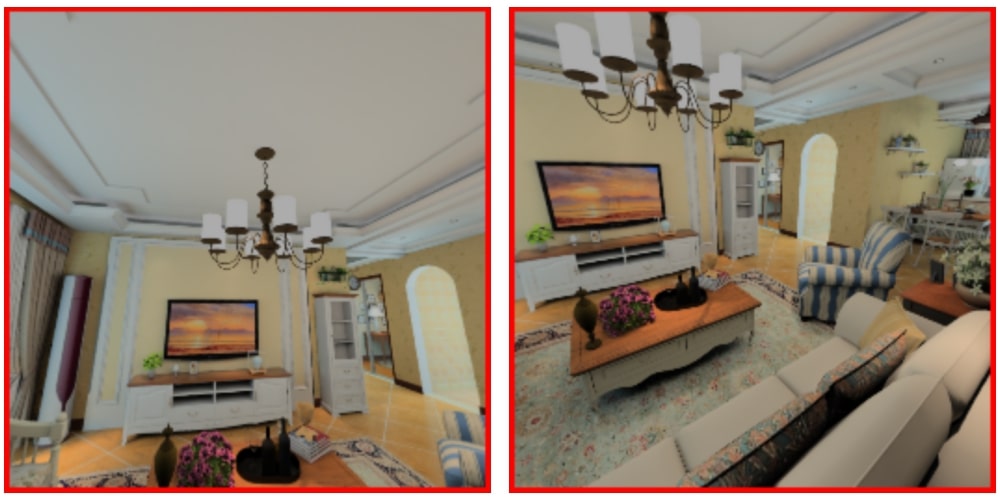}} 
        \\ 
        &
        \T{\includegraphics[width=\sizea, trim={\tal} {\tab} {\tar} {\tat},clip]{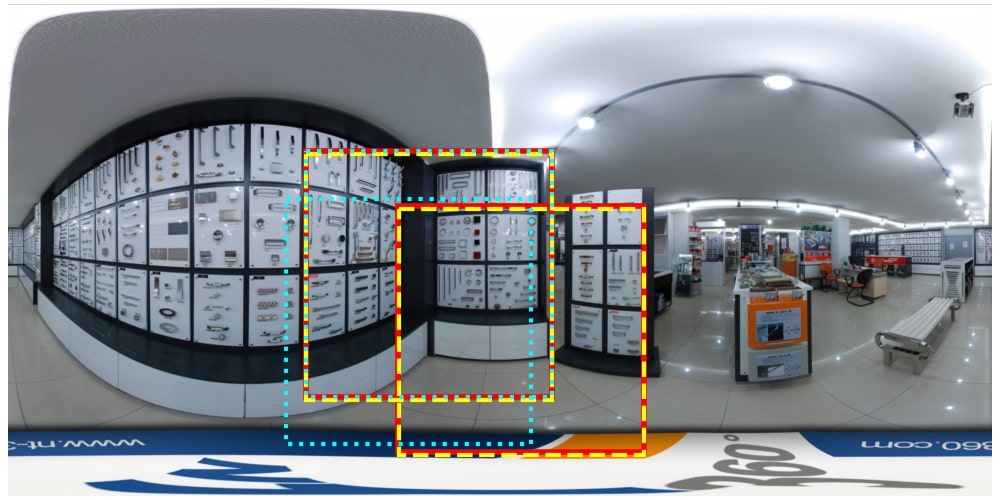}}&
        \T{\includegraphics[width=\sizea, trim={\tal} {\tab} {\tar} {\tat},clip]{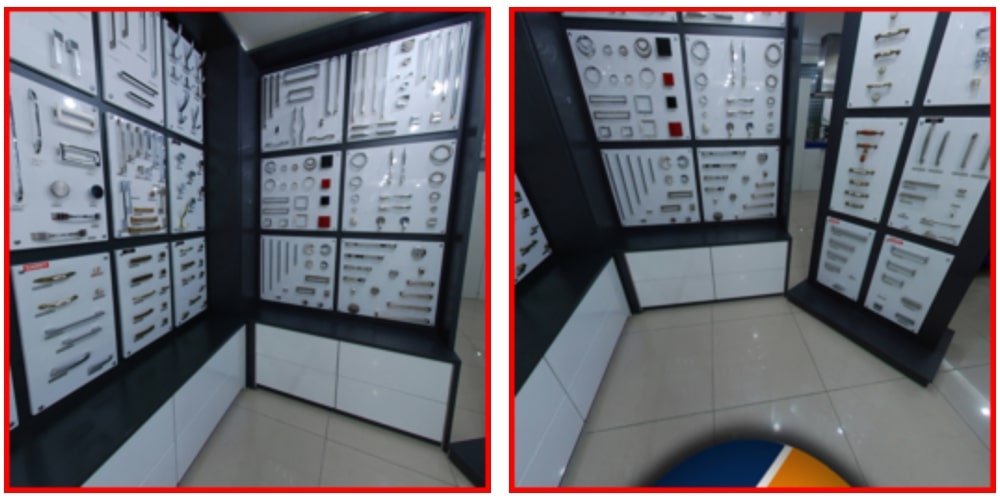}}&
        \T{\includegraphics[width=\sizea, trim={\tal} {\tab} {\tar} {\tat},clip]{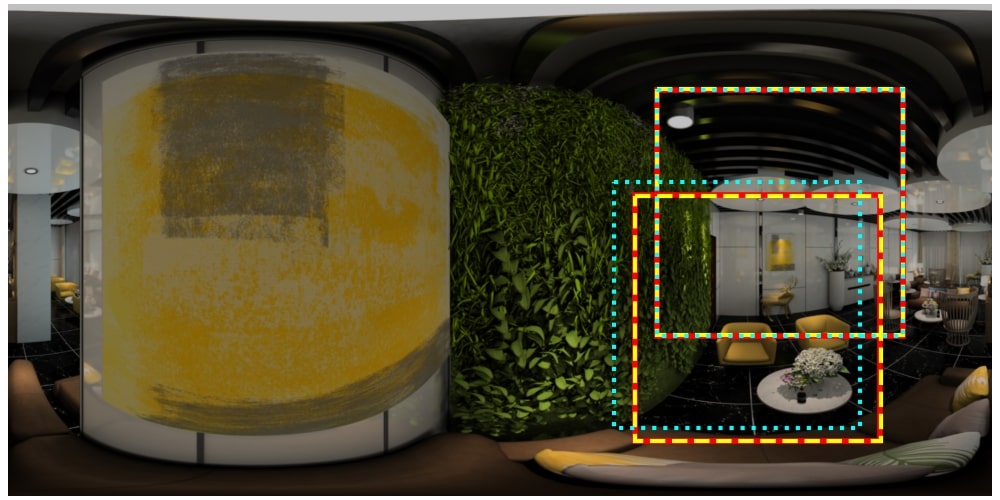}} &
        \T{\includegraphics[width=\sizea, trim={\tal} {\tab} {\tar} {\tat},clip]{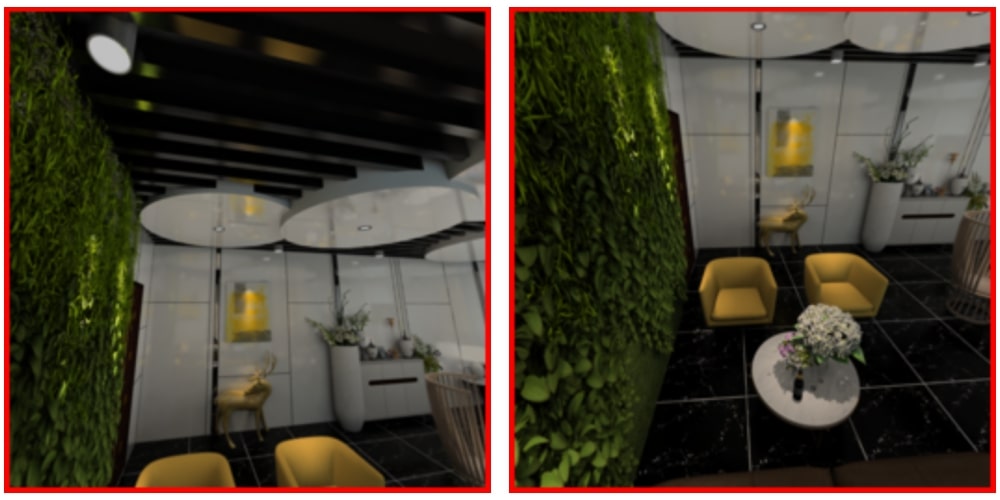}} 
      \\ \vspace{+2pt}
      &
      \T{\includegraphics[width=\sizea, trim={\tal} {\tab} {\tar} {\tat},clip]{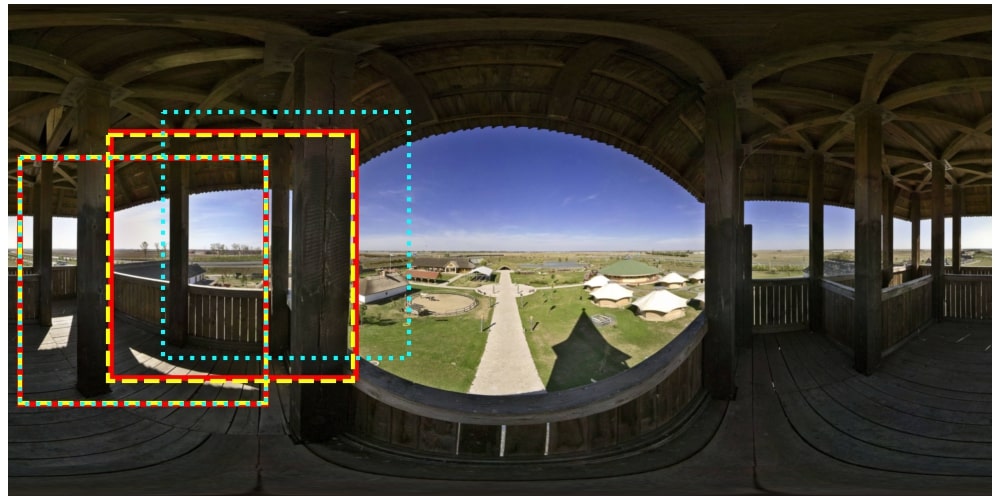}}&
        \T{\includegraphics[width=\sizea, trim={\tal} {\tab} {\tar} {\tat},clip]{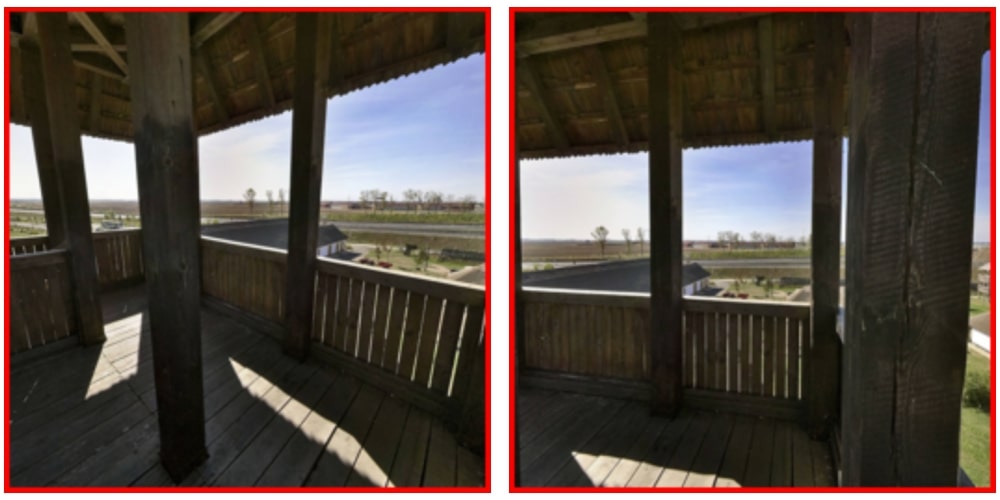}}&
        \T{\includegraphics[width=\sizea, trim={\tal} {\tab} {\tar} {\tat},clip]{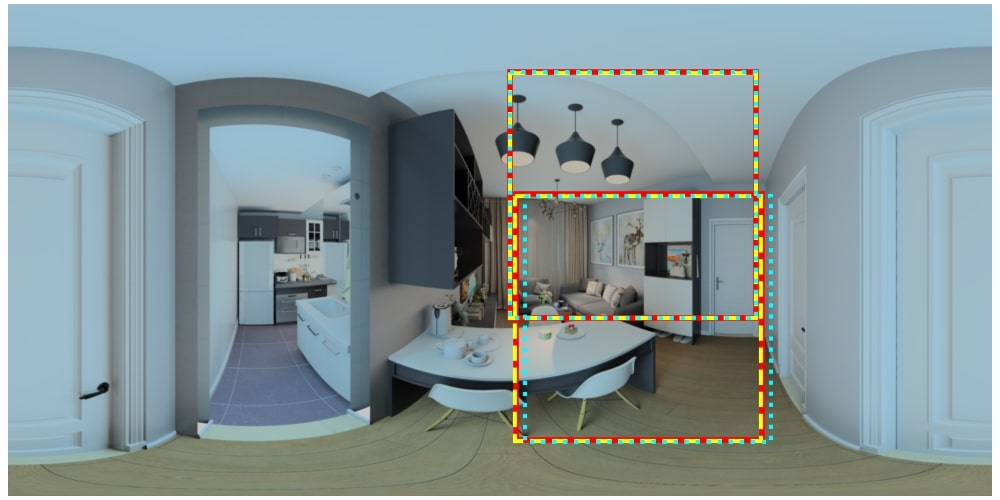}} &
        \T{\includegraphics[width=\sizea, trim={\tal} {\tab} {\tar} {\tat},clip]{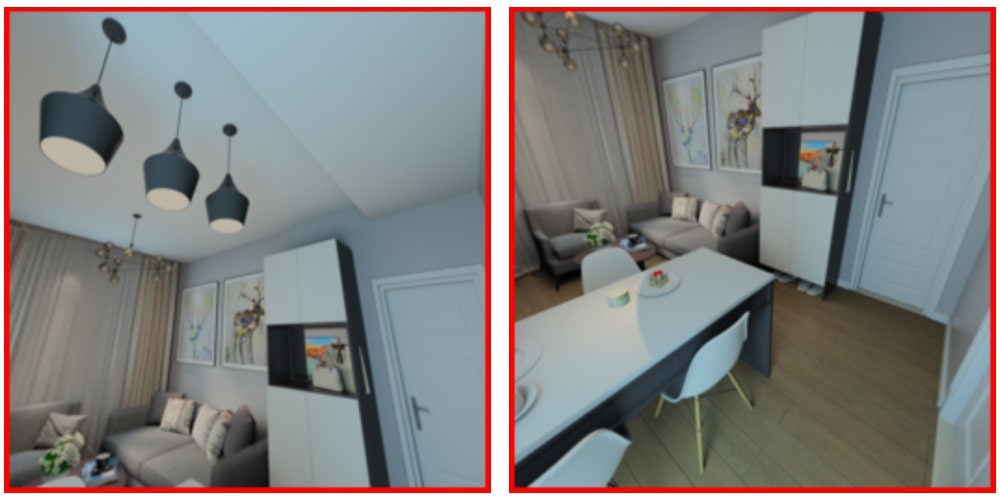}} 
        \\ 
        
        \multirow{3}{*}{\rotatebox[origin=c]{90}{Small\whitetxt{ssssssssssss}}} &
        \T{\includegraphics[width=\sizea, trim={\tal} {\tab} {\tar} {\tat},clip]{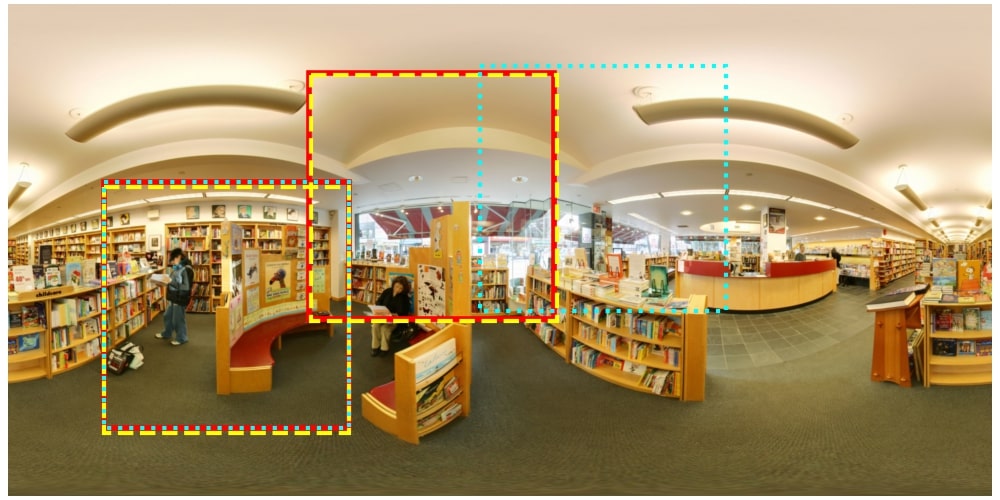}}&
        \T{\includegraphics[width=\sizea, trim={\tal} {\tab} {\tar} {\tat},clip]{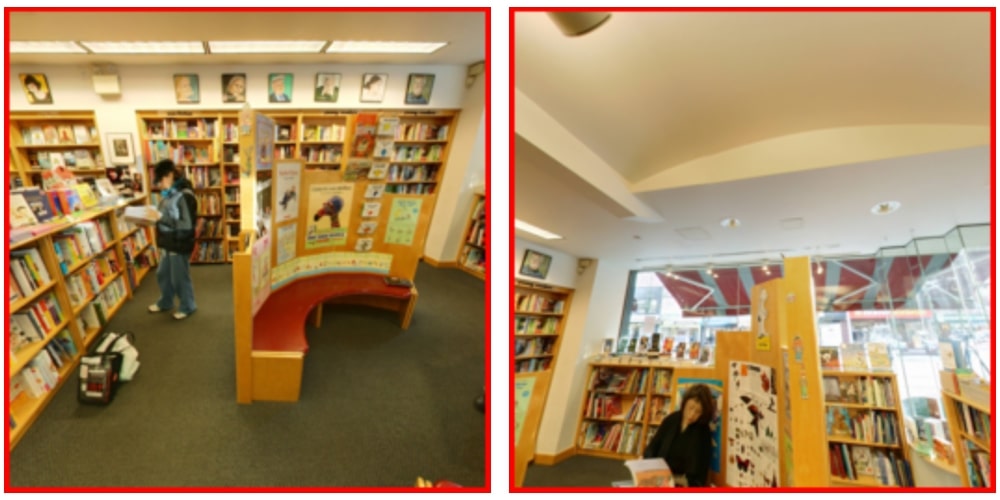}}&
        \T{\includegraphics[width=\sizea, trim={\tal} {\tab} {\tar} {\tat},clip]{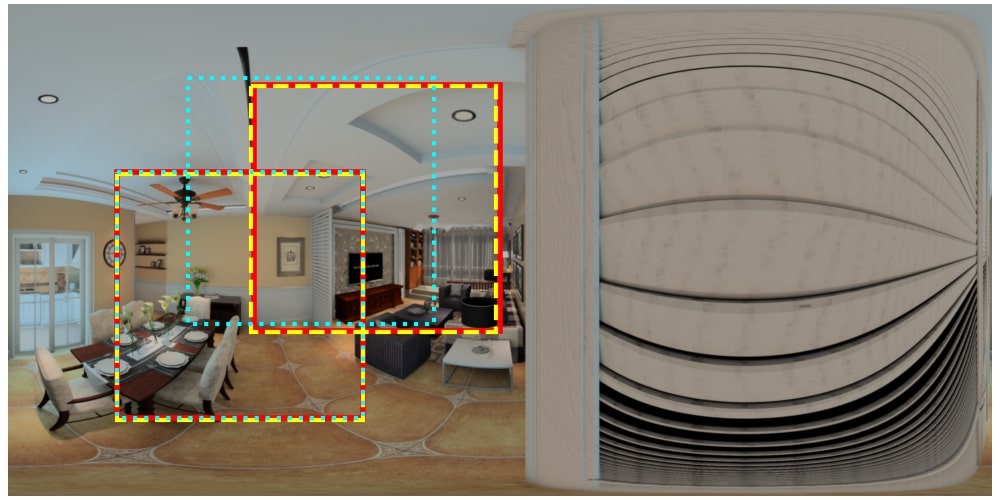}} &
        \T{\includegraphics[width=\sizea, trim={\tal} {\tab} {\tar} {\tat},clip]{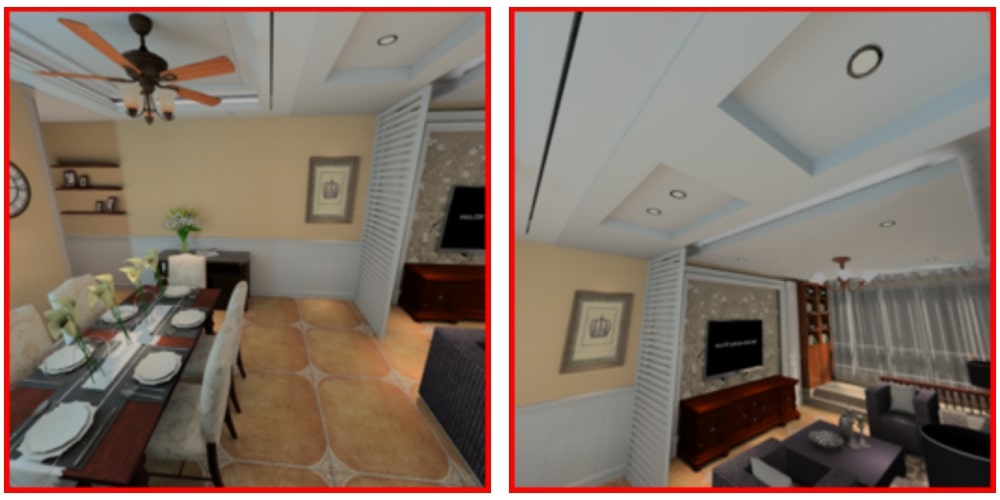}} 
        \\ 
        &
        \T{\includegraphics[width=\sizea, trim={\tal} {\tab} {\tar} {\tat},clip]{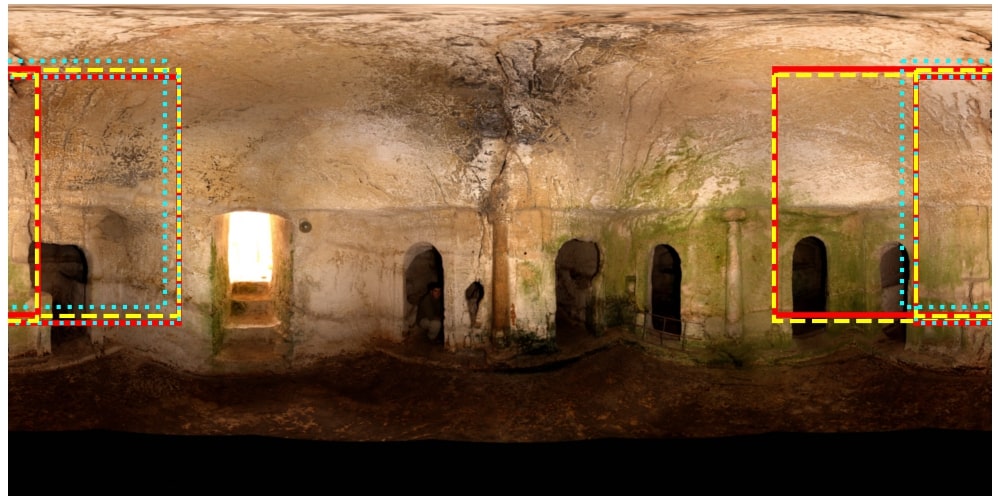}}&
        \T{\includegraphics[width=\sizea, trim={\tal} {\tab} {\tar} {\tat},clip]{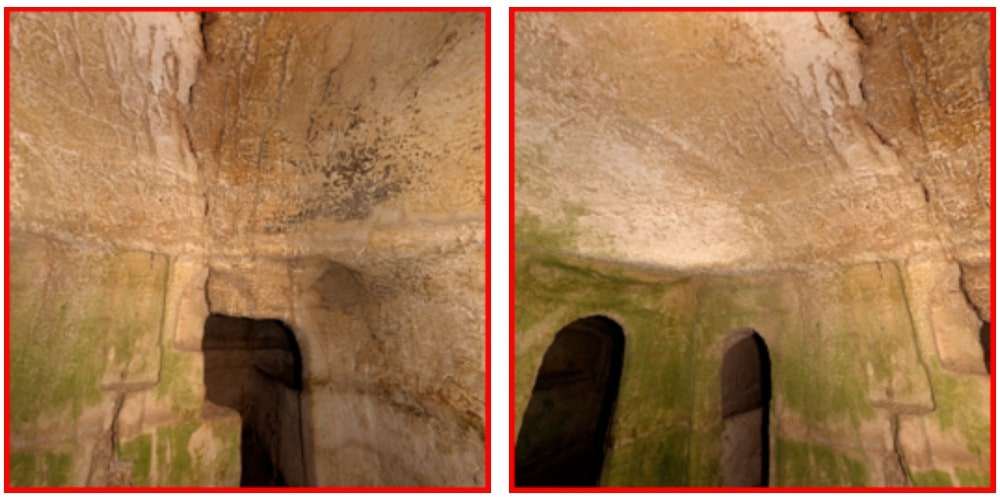}}&
        \T{\includegraphics[width=\sizea, trim={\tal} {\tab} {\tar} {\tat},clip]{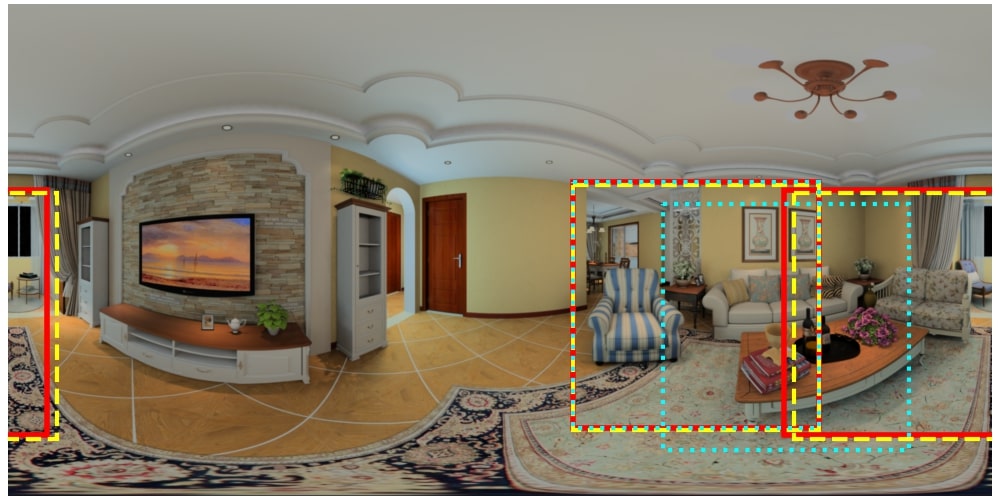}} &
        \T{\includegraphics[width=\sizea, trim={\tal} {\tab} {\tar} {\tat},clip]{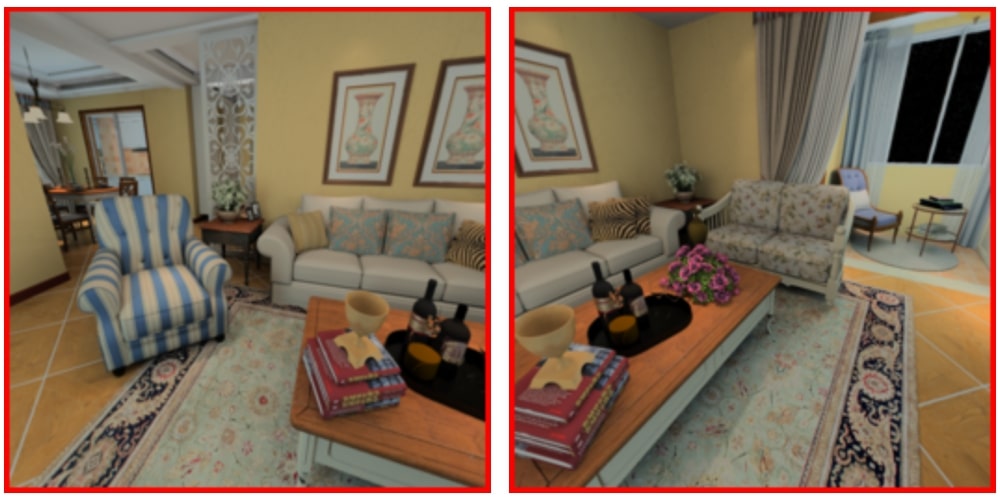}} 
      \\ \vspace{+2pt}
      &
      \T{\includegraphics[width=\sizea, trim={\tal} {\tab} {\tar} {\tat},clip]{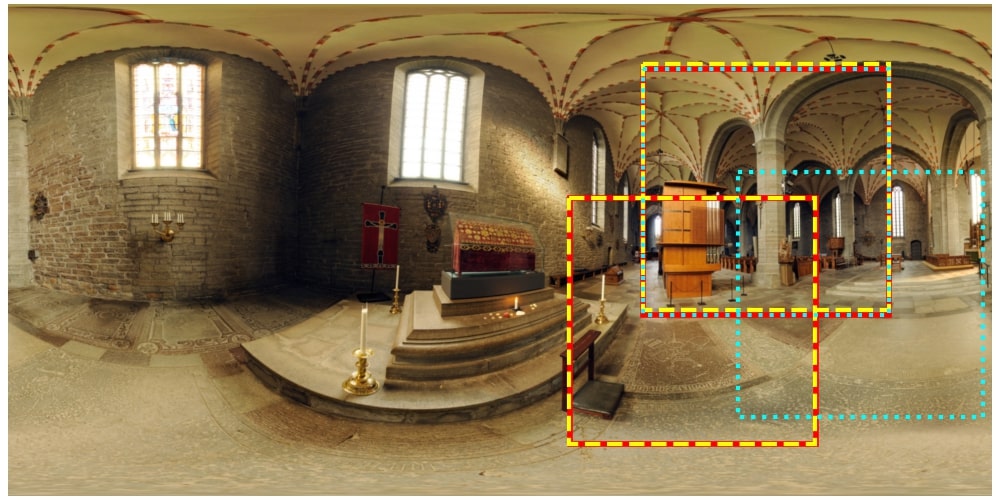}}&
        \T{\includegraphics[width=\sizea, trim={\tal} {\tab} {\tar} {\tat},clip]{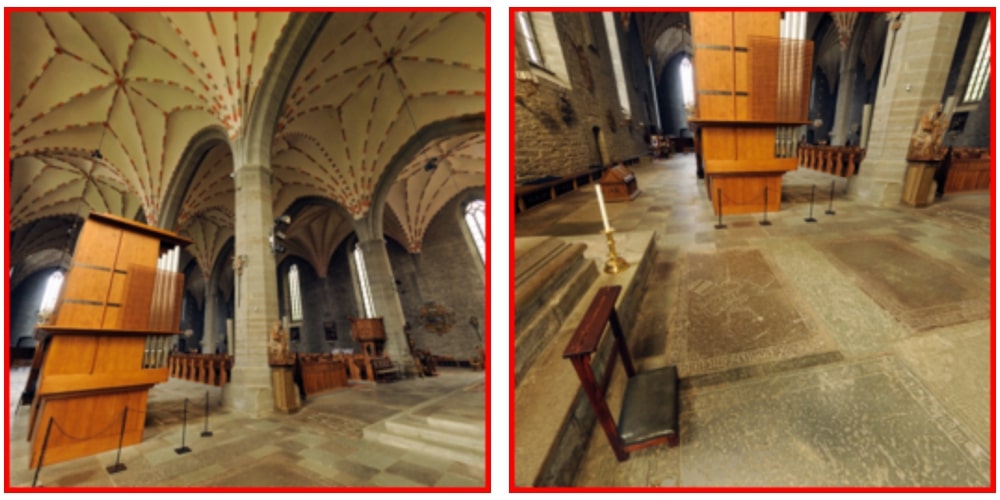}}&
        \T{\includegraphics[width=\sizea, trim={\tal} {\tab} {\tar} {\tat},clip]{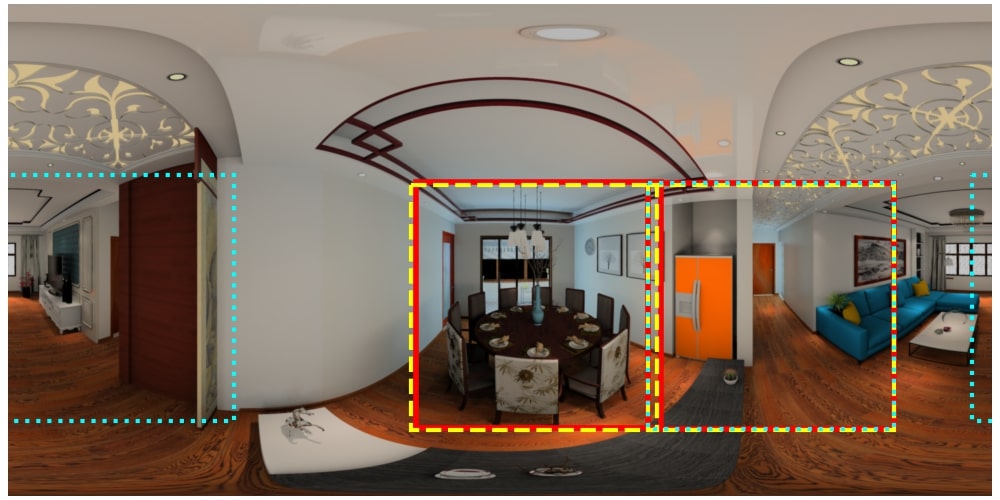}} &
        \T{\includegraphics[width=\sizea, trim={\tal} {\tab} {\tar} {\tat},clip]{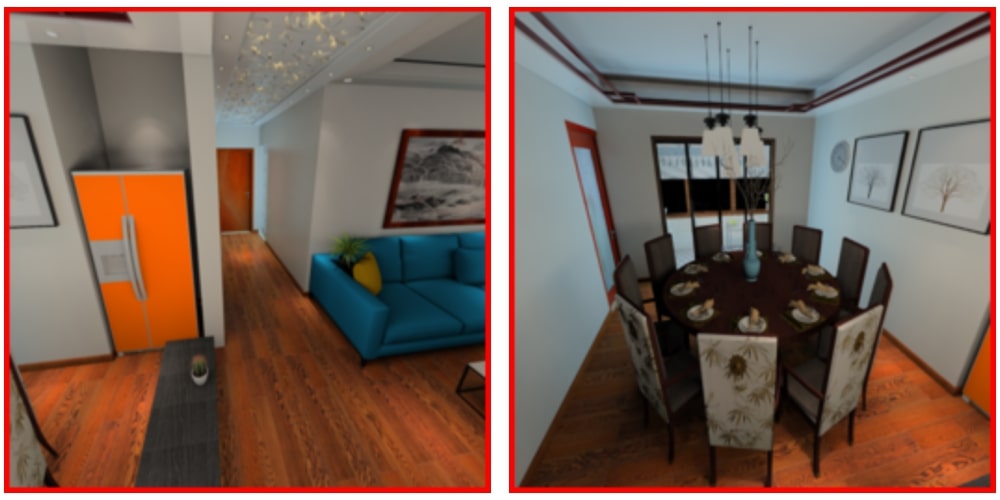}} 
        
        \\
        \multirow{3}{*}{\rotatebox[origin=c]{90}{None\whitetxt{sssssssssssss}}} &
        \T{\includegraphics[width=\sizea, trim={\tal} {\tab} {\tar} {\tat},clip]{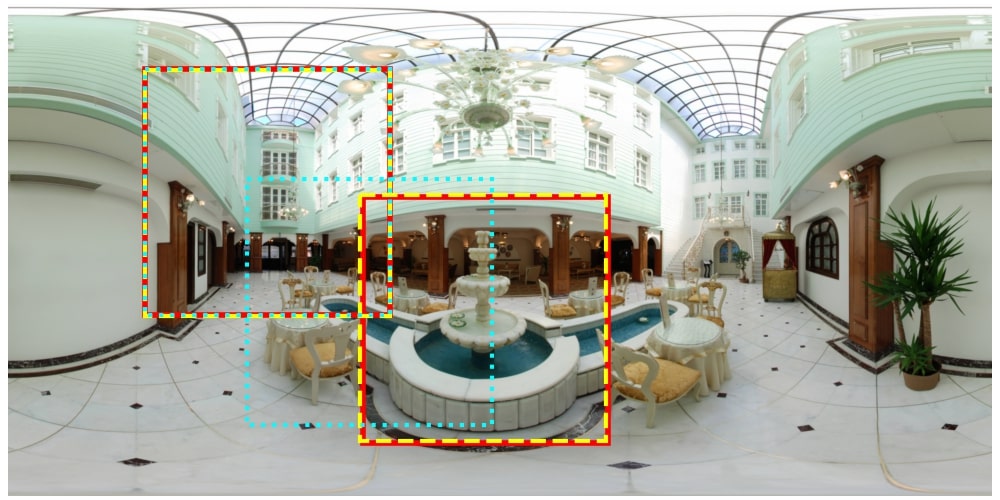}}&
        \T{\includegraphics[width=\sizea, trim={\tal} {\tab} {\tar} {\tat},clip]{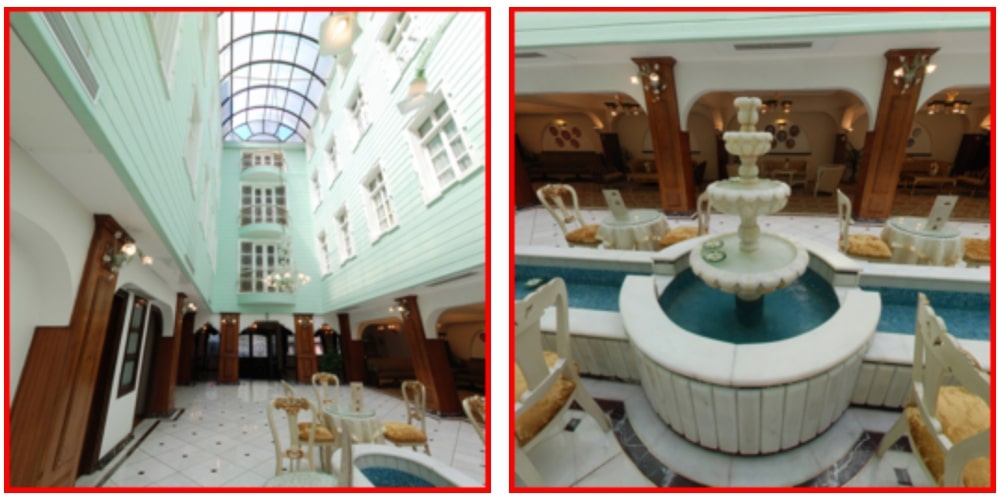}}&
        \T{\includegraphics[width=\sizea, trim={\tal} {\tab} {\tar} {\tat},clip]{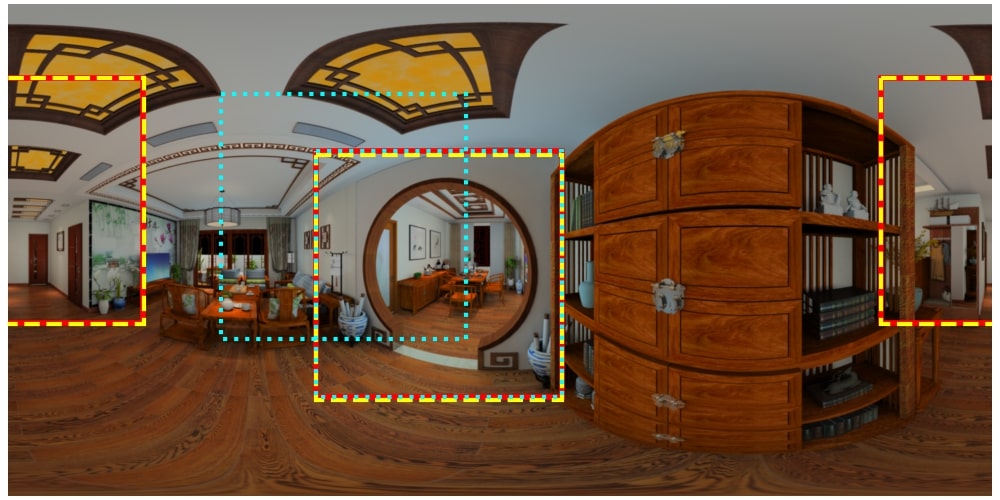}} &
        \T{\includegraphics[width=\sizea, trim={\tal} {\tab} {\tar} {\tat},clip]{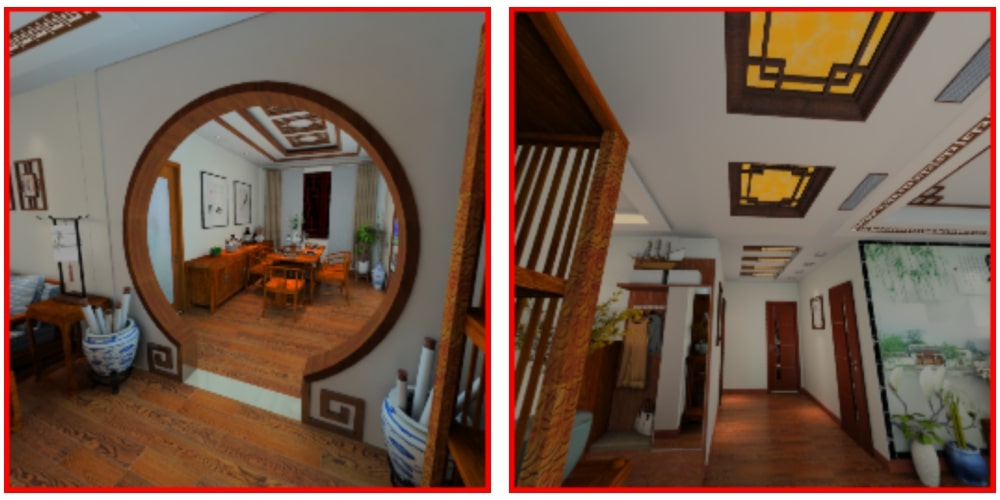}} 
        \\ 
        &
        \T{\includegraphics[width=\sizea, trim={\tal} {\tab} {\tar} {\tat},clip]{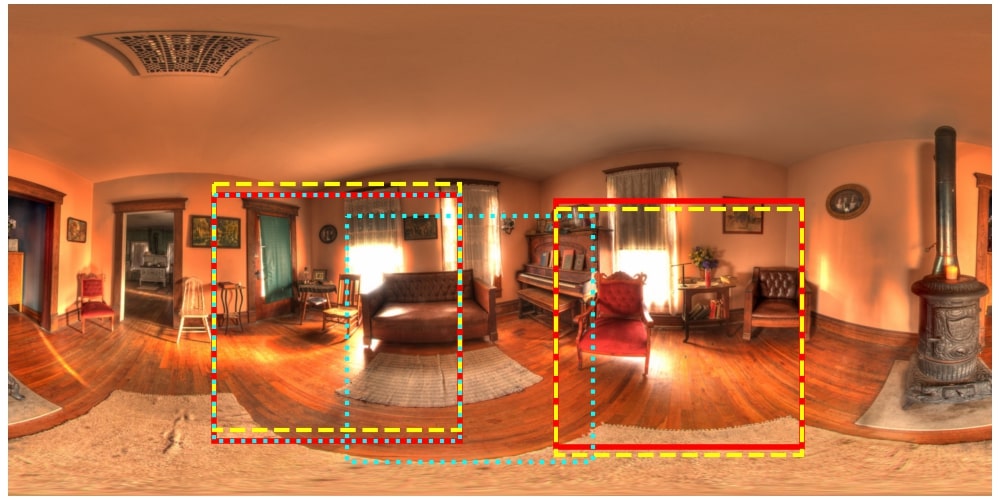}}&
        \T{\includegraphics[width=\sizea, trim={\tal} {\tab} {\tar} {\tat},clip]{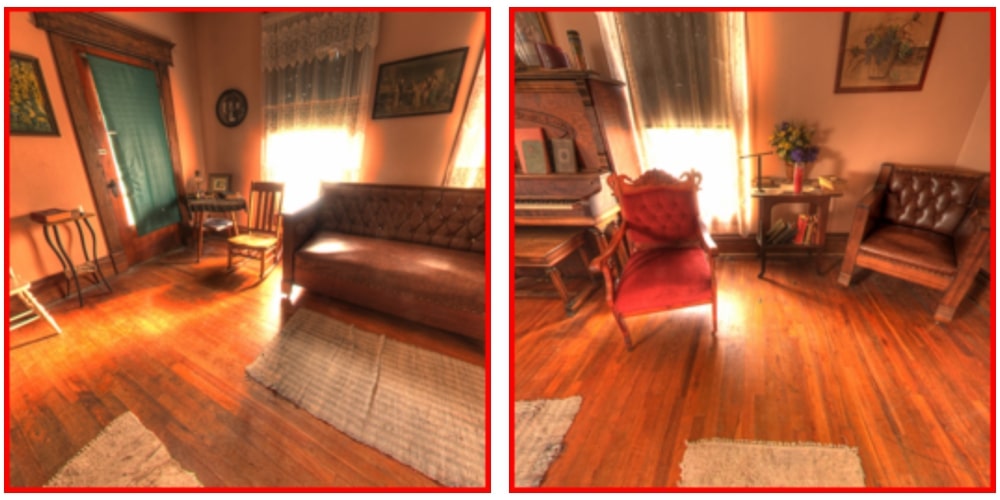}}&
        \T{\includegraphics[width=\sizea, trim={\tal} {\tab} {\tar} {\tat},clip]{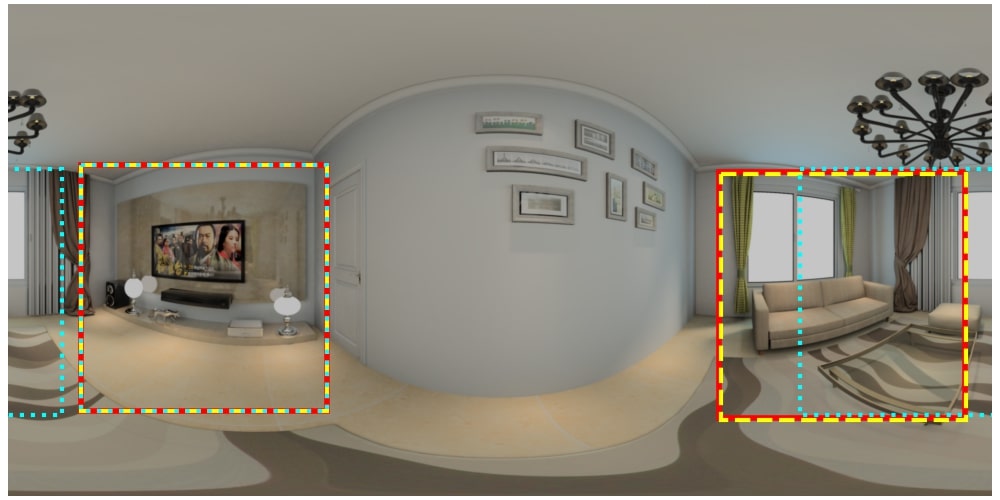}} &
        \T{\includegraphics[width=\sizea, trim={\tal} {\tab} {\tar} {\tat},clip]{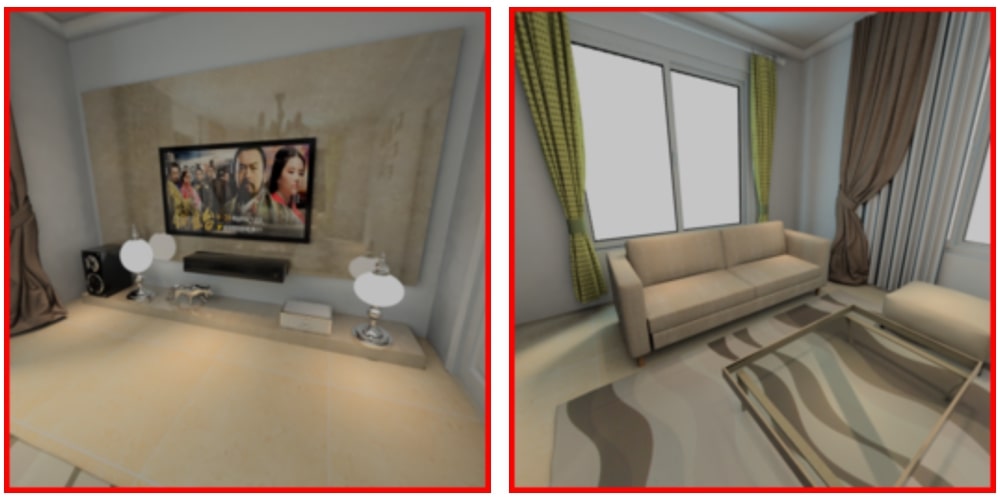}} 
      \\ \vspace{+2pt}
      &
      \T{\includegraphics[width=\sizea, trim={\tal} {\tab} {\tar} {\tat},clip]{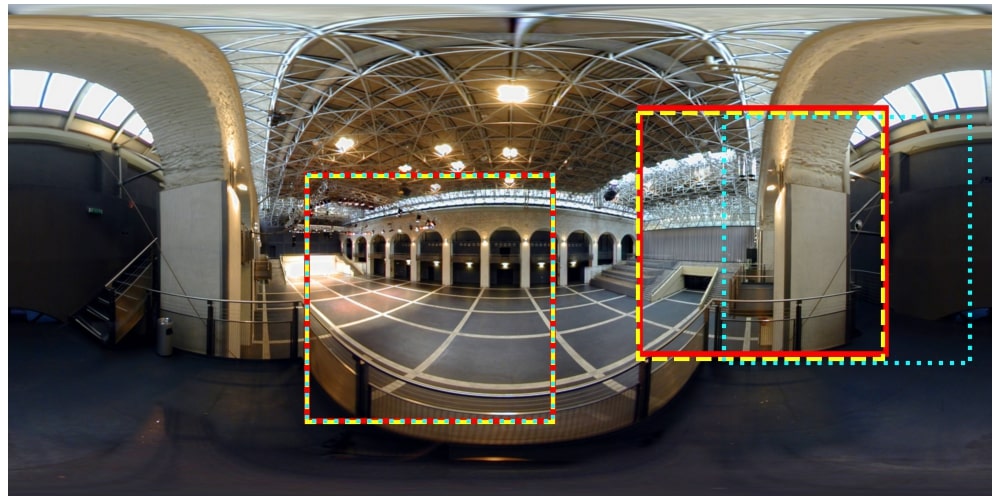}}&
        \T{\includegraphics[width=\sizea, trim={\tal} {\tab} {\tar} {\tat},clip]{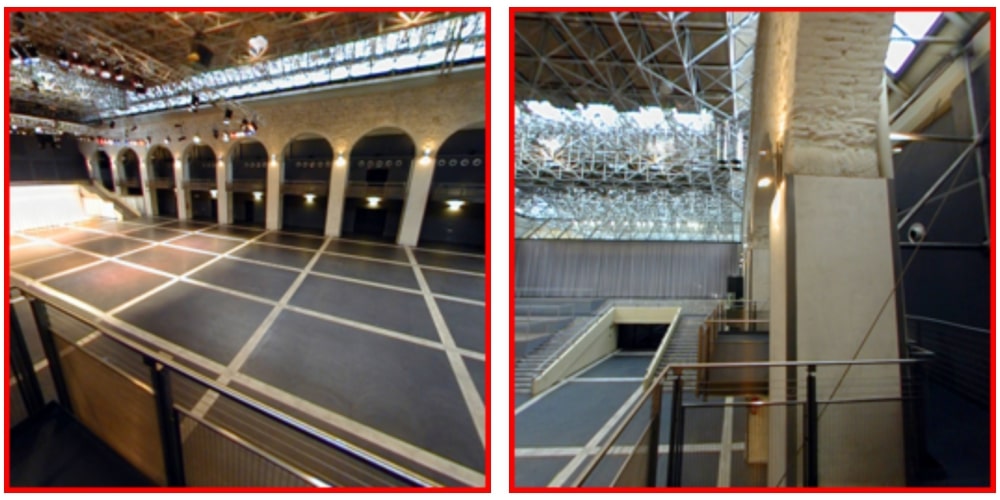}}&
        \T{\includegraphics[width=\sizea, trim={\tal} {\tab} {\tar} {\tat},clip]{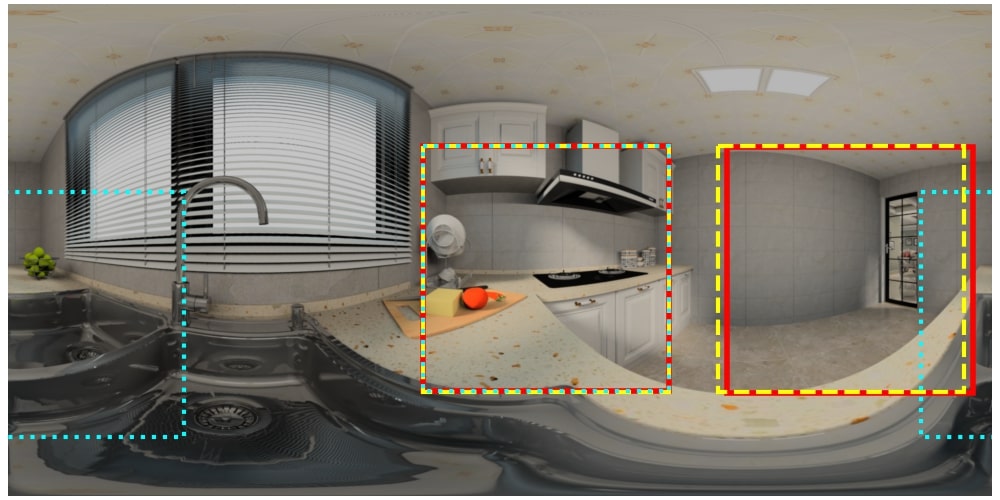}} &
        \T{\includegraphics[width=\sizea, trim={\tal} {\tab} {\tar} {\tat},clip]{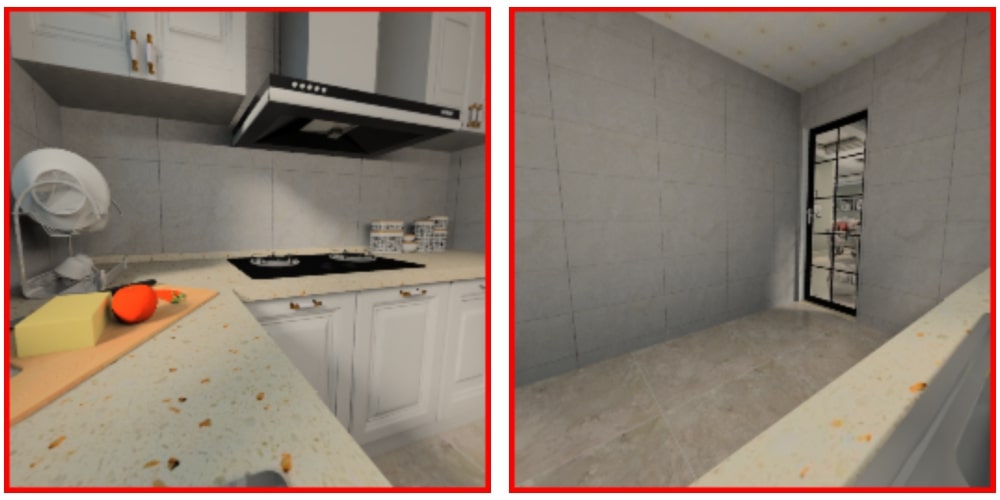}} 
        \\ 
        & \multicolumn{2}{c}{SUN360} & \multicolumn{2}{c}{InteriorNet} 
    \end{tabular}
    \end{center}
    \caption{\textbf{Predicted rotation results of indoor datasets.} Full panoramas are shown on the left, with the ground-truth perspective images marked in red. We show our predicted viewpoints in yellow, and results obtained using the regression model of Zhou \etal~\cite{zhou2019continuity} in blue.
    }
    \label{fig:predicted_indoor}
\end{figure*}

In Figure \ref{fig:clues_supp}, we provide additional visualizations of cues detected in image pairs.
We provide more qualitative results on StreetLearn, SUN360, and InteriorNet in Figures \ref{fig:predicted_street} and \ref{fig:predicted_indoor}.
Additional qualitative results for London and Pittsburgh are shown in Figure \ref{fig:new_cities}.

\begin{figure*}
    \begin{center}
    \newcommand{\sizea}{0.2375\textwidth}
    \newcommand{\tal}{0cm}
    \newcommand{\tab}{0cm}
    \newcommand{\tar}{0cm}
    \newcommand{\tat}{0cm}
    \newcommand{\smb}{0cm}
    \newcommand{\T}[1]{\raisebox{-0.5\height}{#1}}
    \setlength{\tabcolsep}{0pt}
    \renewcommand{\arraystretch}{0}
    \begin{tabular}{@{}ccccc@{}}
        \multirow{3}{*}{\rotatebox[origin=c]{90}{Large \whitetxt{ssssssssssss}}} &
        \T{\includegraphics[width=\sizea, trim={\tal} {\tab} {\tar} {\tat},clip]{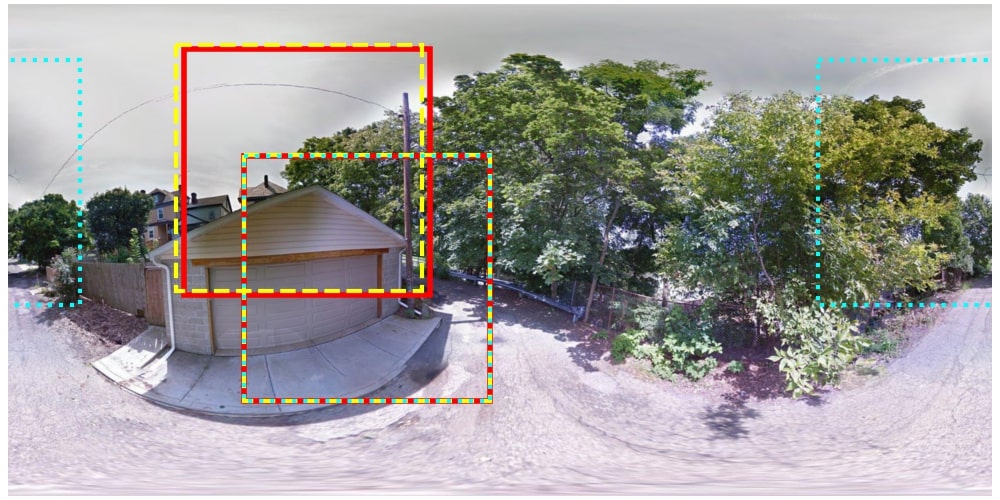}} &
        \T{\includegraphics[width=\sizea, trim={\tal} {\tab} {\tar} {\tat},clip]{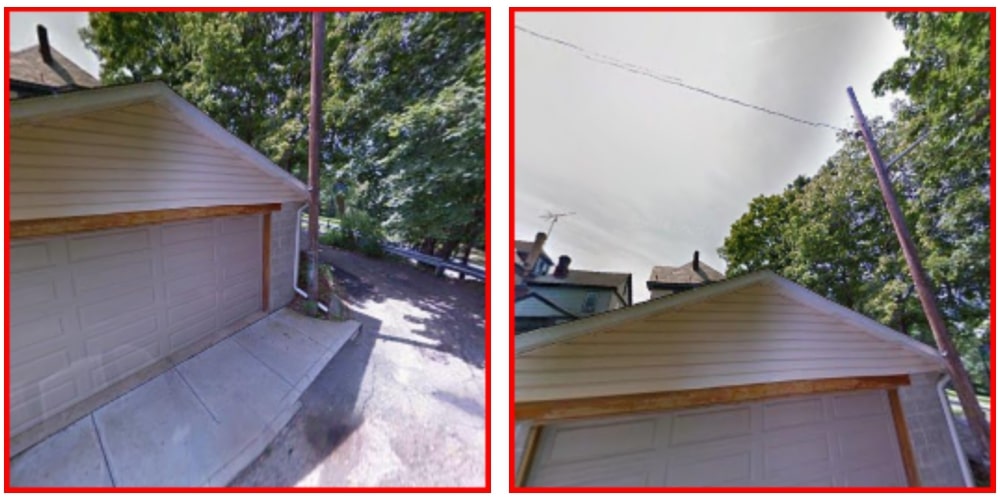}} &
        \T{\includegraphics[width=\sizea, trim={\tal} {\tab} {\tar} {\tat},clip]{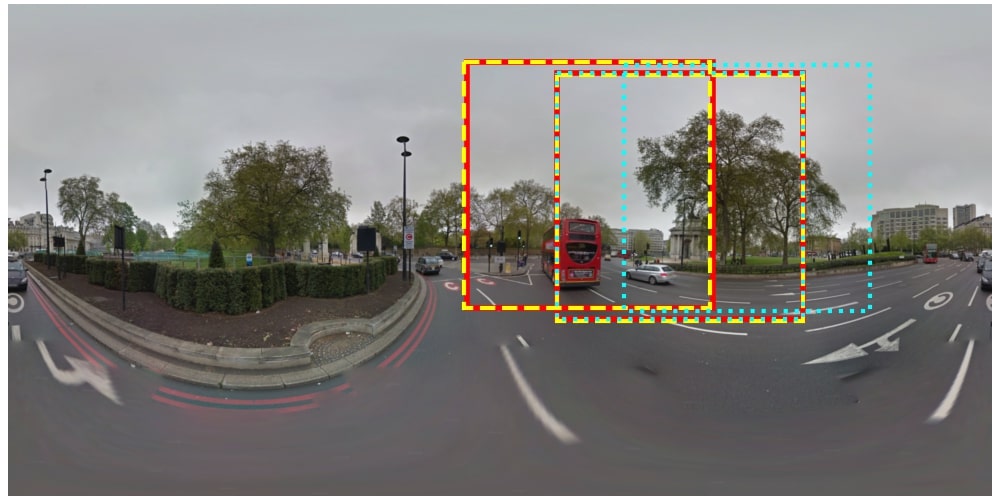}}&
        \T{\includegraphics[width=\sizea, trim={\tal} {\tab} {\tar} {\tat},clip]{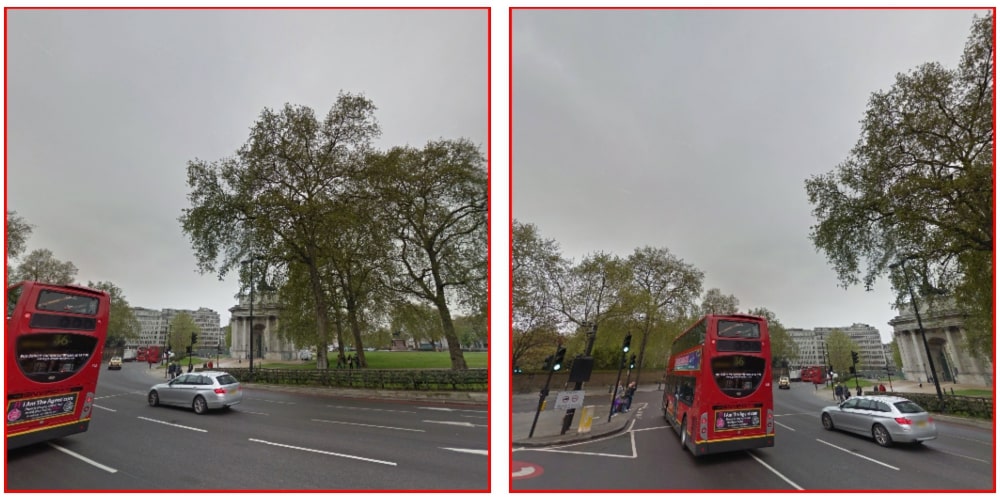}}
        \\ 
        &
        \T{\includegraphics[width=\sizea, trim={\tal} {\tab} {\tar} {\tat},clip]{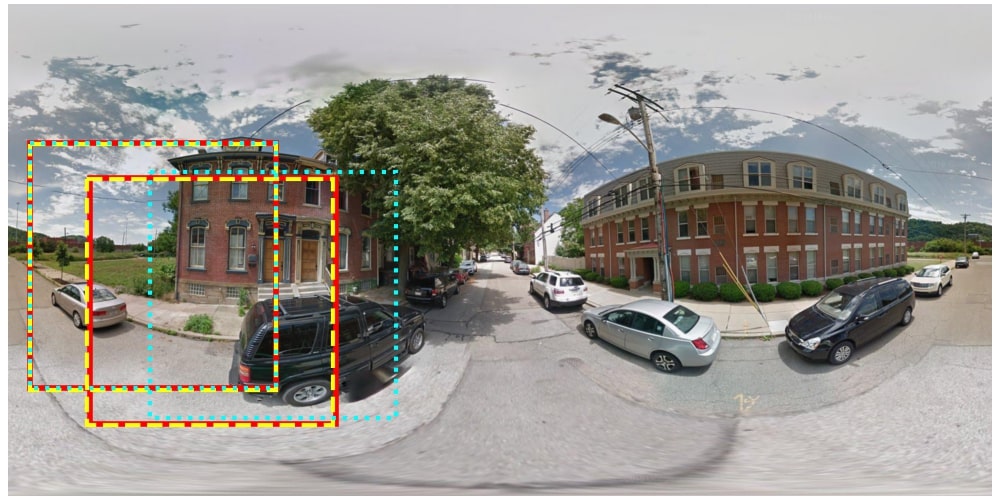}} &
        \T{\includegraphics[width=\sizea, trim={\tal} {\tab} {\tar} {\tat},clip]{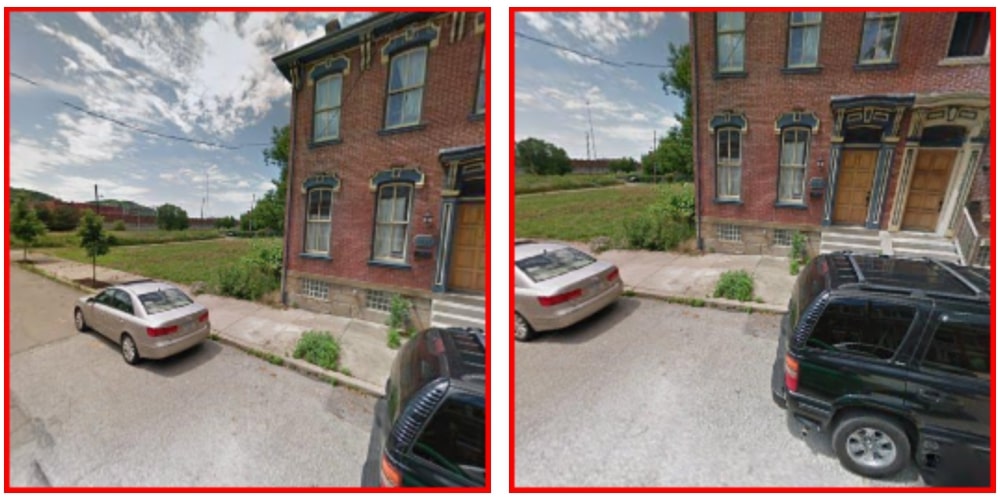}} &
        \T{\includegraphics[width=\sizea, trim={\tal} {\tab} {\tar} {\tat},clip]{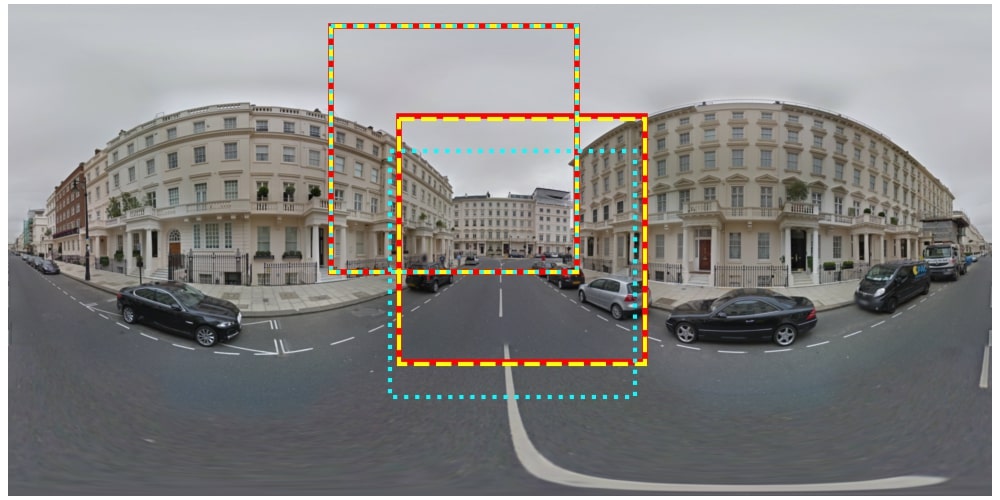}}&
        \T{\includegraphics[width=\sizea, trim={\tal} {\tab} {\tar} {\tat},clip]{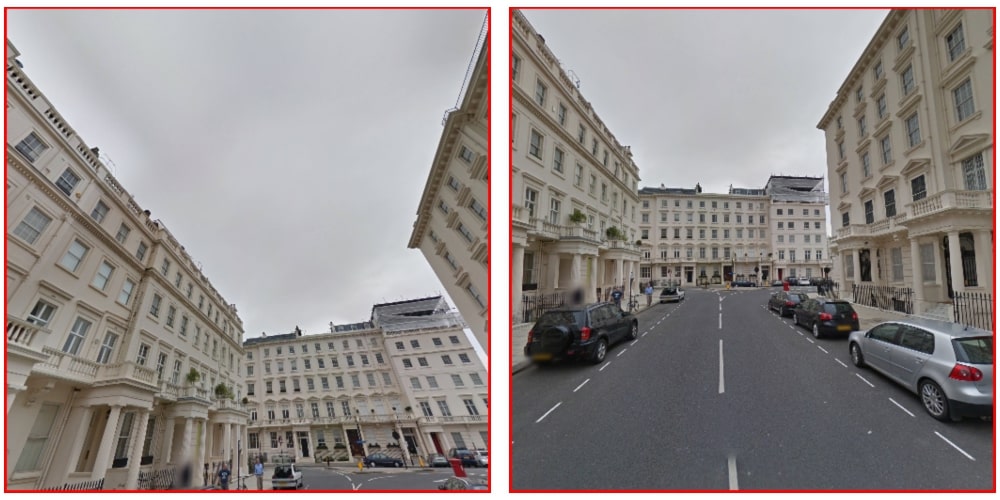}}
       \\ \vspace{+2pt}
       &
       \T{\includegraphics[width=\sizea, trim={\tal} {\tab} {\tar} {\tat},clip]{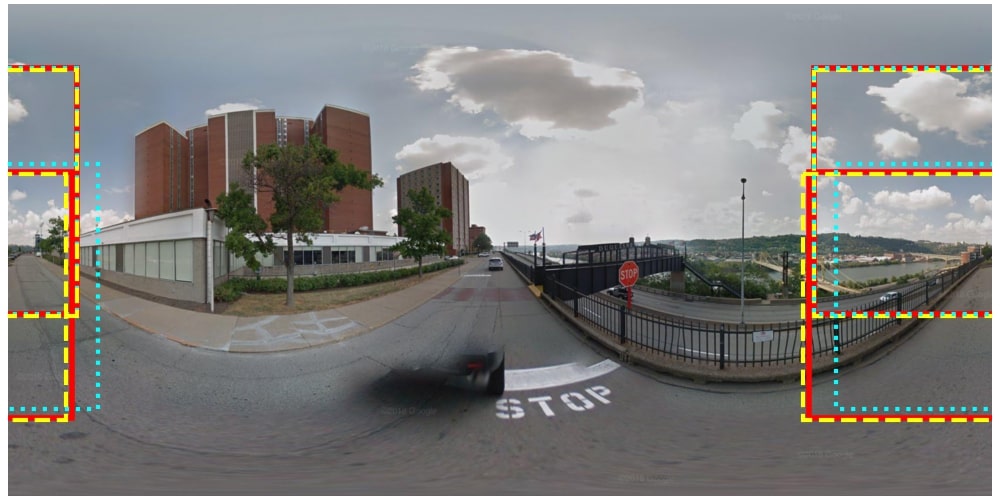}} &
        \T{\includegraphics[width=\sizea, trim={\tal} {\tab} {\tar} {\tat},clip]{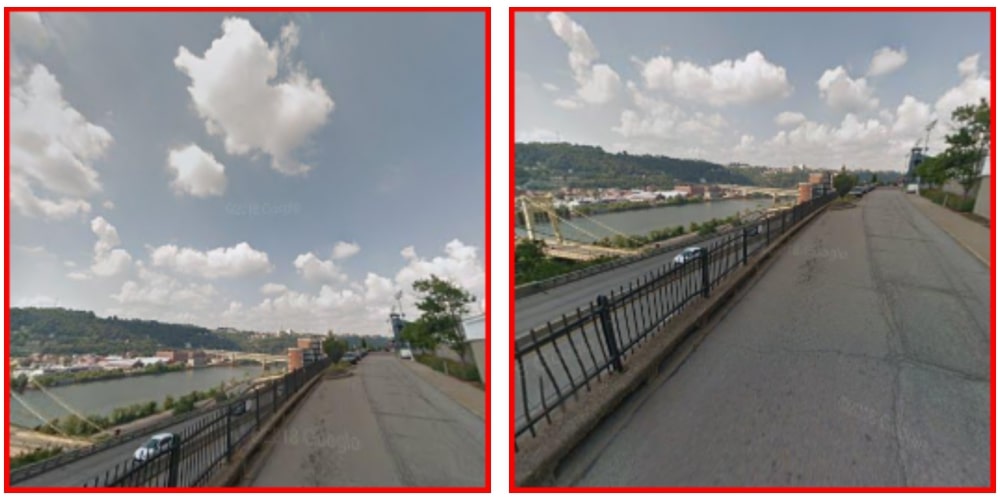}} &
        \T{\includegraphics[width=\sizea, trim={\tal} {\tab} {\tar} {\tat},clip]{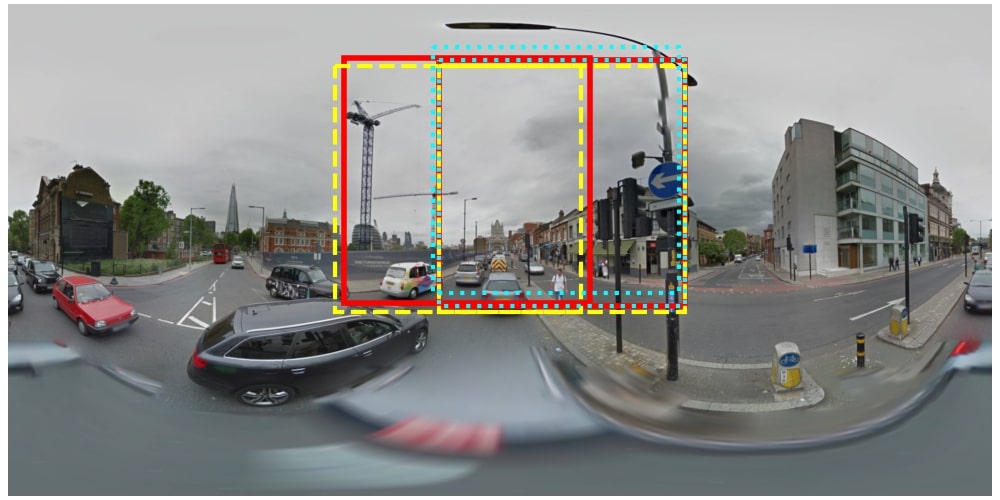}}&
        \T{\includegraphics[width=\sizea, trim={\tal} {\tab} {\tar} {\tat},clip]{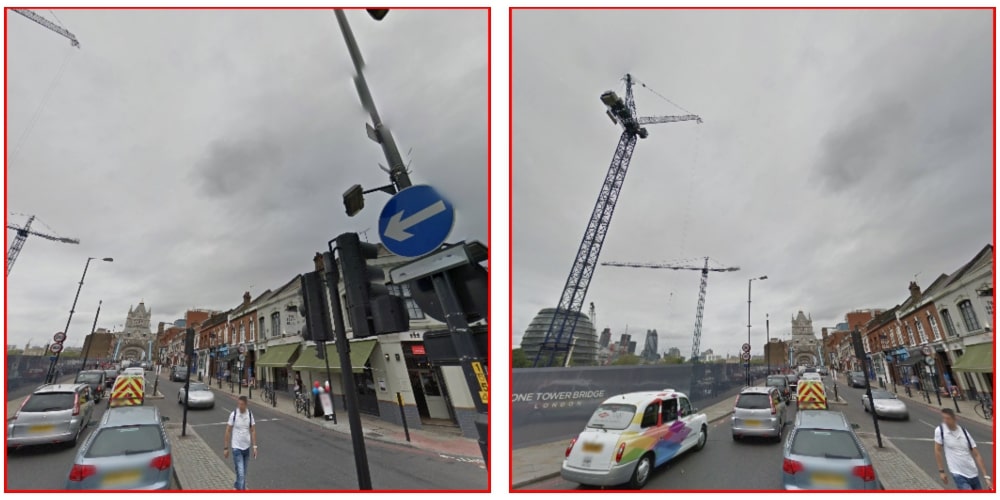}}
        \\ 
        
        \multirow{3}{*}{\rotatebox[origin=c]{90}{Small\whitetxt{ssssssssssss}}} &
        \T{\includegraphics[width=\sizea, trim={\tal} {\tab} {\tar} {\tat},clip]{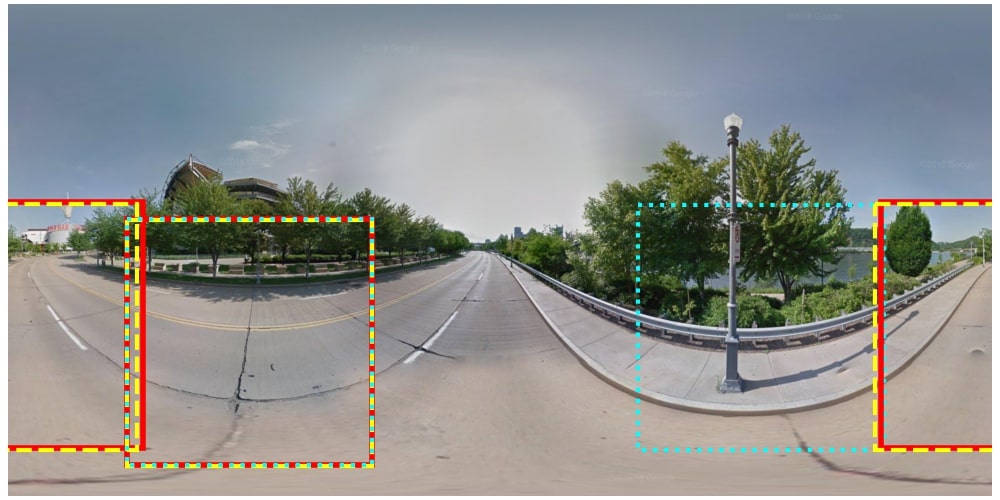}} &
        \T{\includegraphics[width=\sizea, trim={\tal} {\tab} {\tar} {\tat},clip]{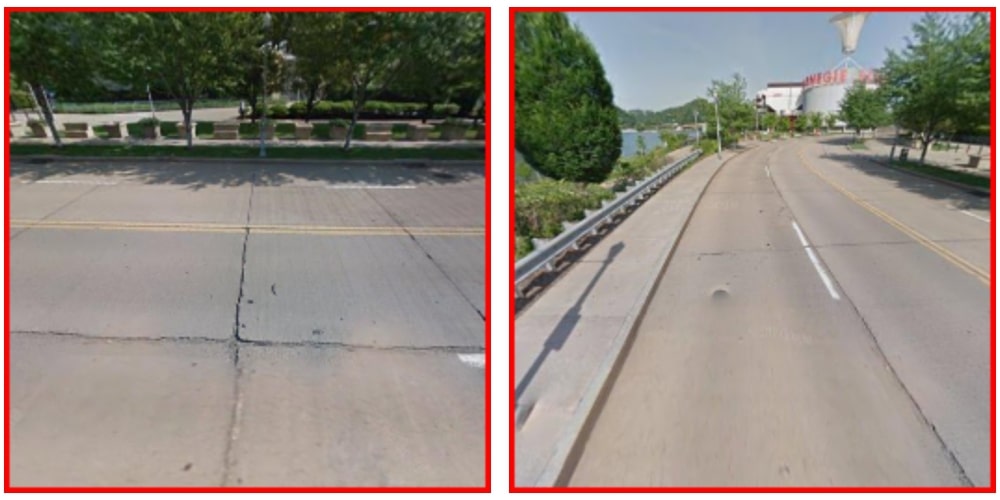}} &
        \T{\includegraphics[width=\sizea, trim={\tal} {\tab} {\tar} {\tat},clip]{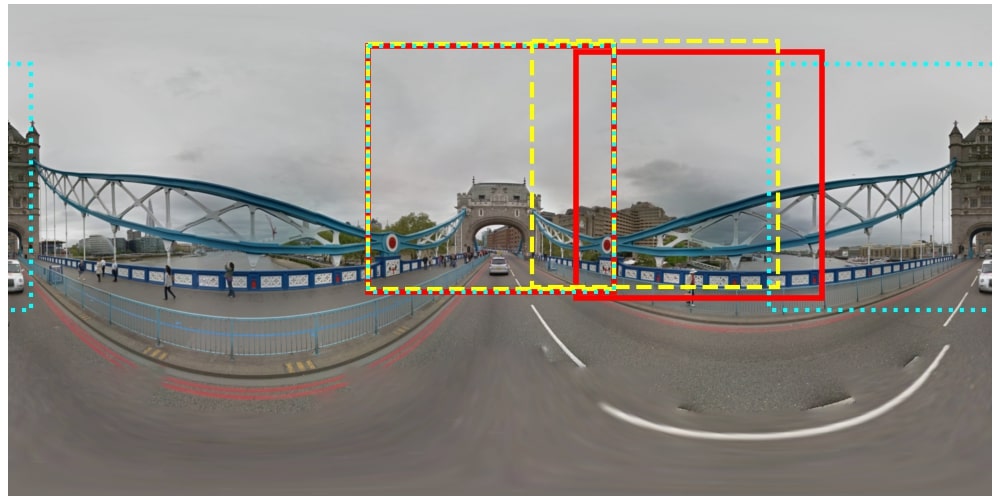}}&
        \T{\includegraphics[width=\sizea, trim={\tal} {\tab} {\tar} {\tat},clip]{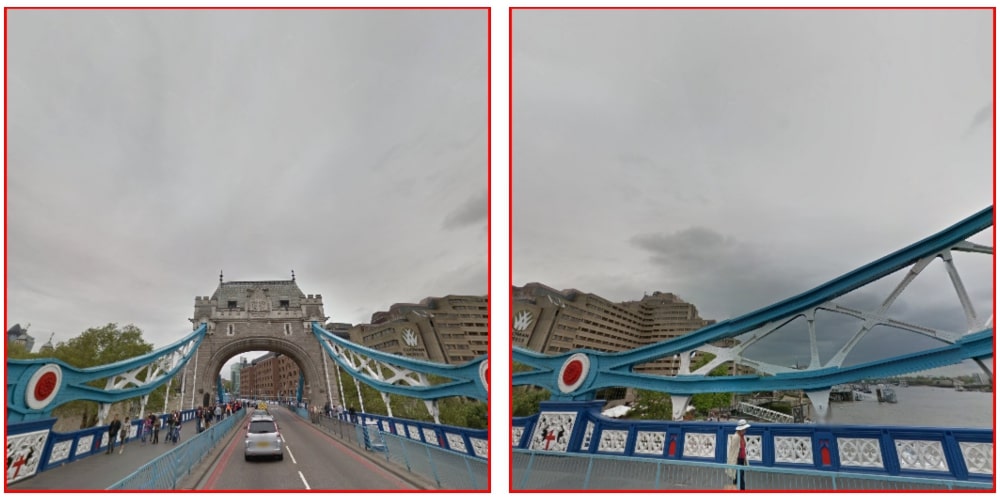}}
        \\ 
        &
        \T{\includegraphics[width=\sizea, trim={\tal} {\tab} {\tar} {\tat},clip]{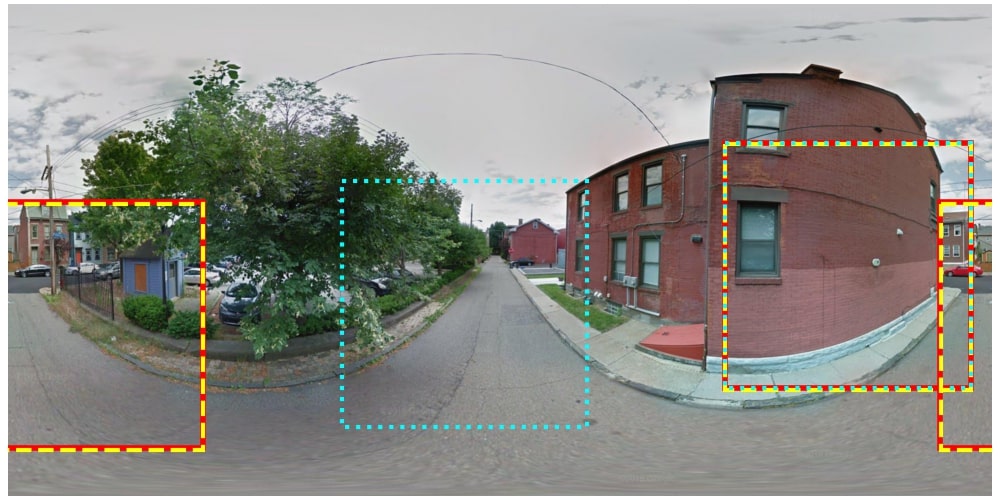}} &
        \T{\includegraphics[width=\sizea, trim={\tal} {\tab} {\tar} {\tat},clip]{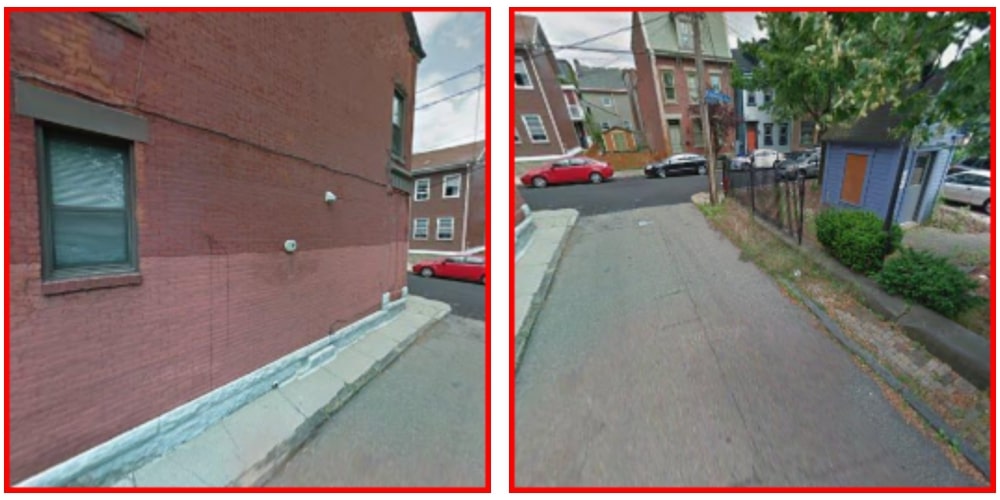}} &
        \T{\includegraphics[width=\sizea, trim={\tal} {\tab} {\tar} {\tat},clip]{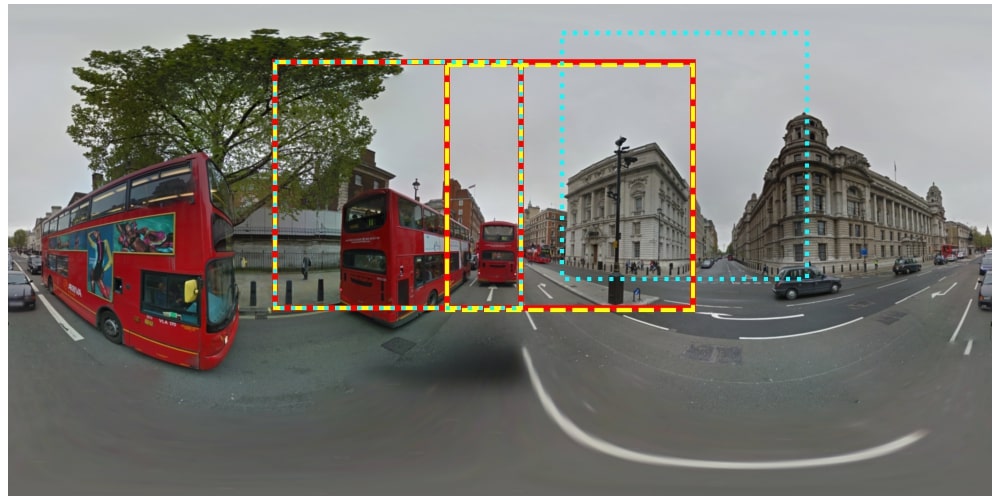}}&
        \T{\includegraphics[width=\sizea, trim={\tal} {\tab} {\tar} {\tat},clip]{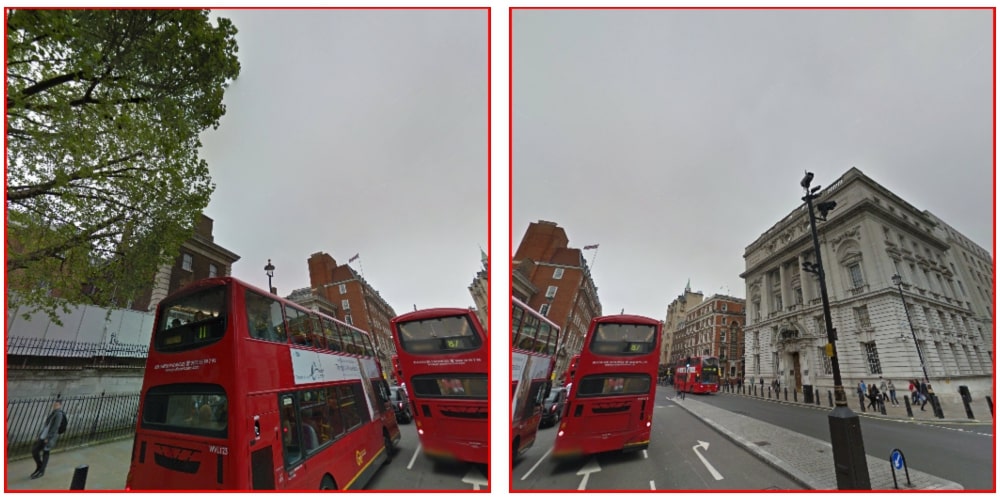}}
       \\ \vspace{+2pt}
       &
       \T{\includegraphics[width=\sizea, trim={\tal} {\tab} {\tar} {\tat},clip]{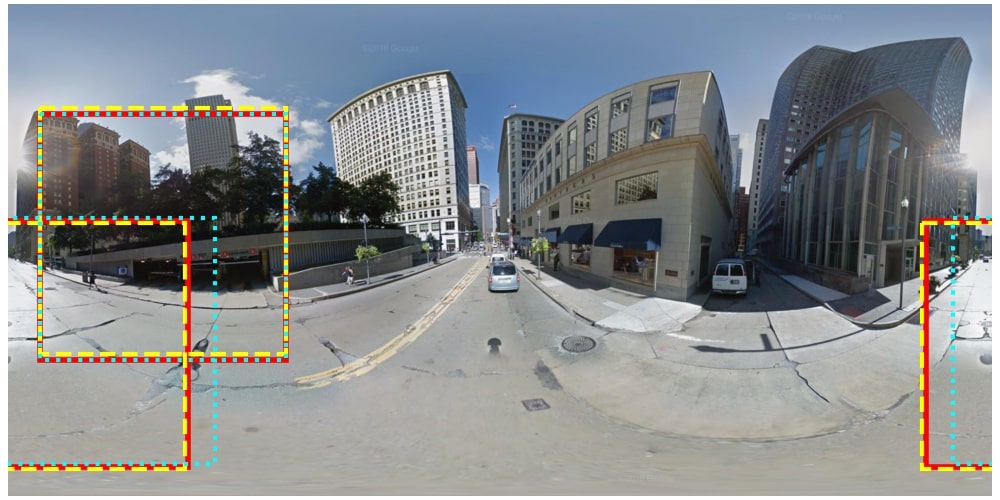}} &
        \T{\includegraphics[width=\sizea, trim={\tal} {\tab} {\tar} {\tat},clip]{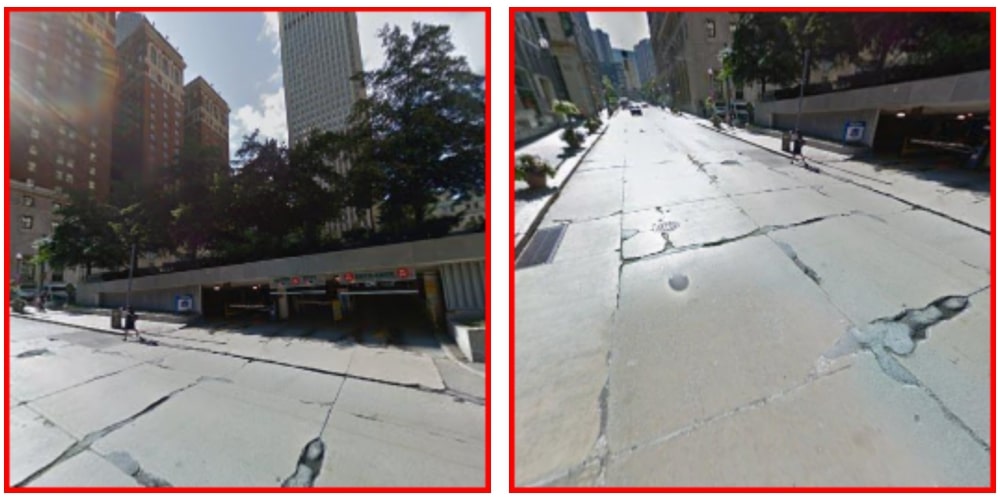}} &
        \T{\includegraphics[width=\sizea, trim={\tal} {\tab} {\tar} {\tat},clip]{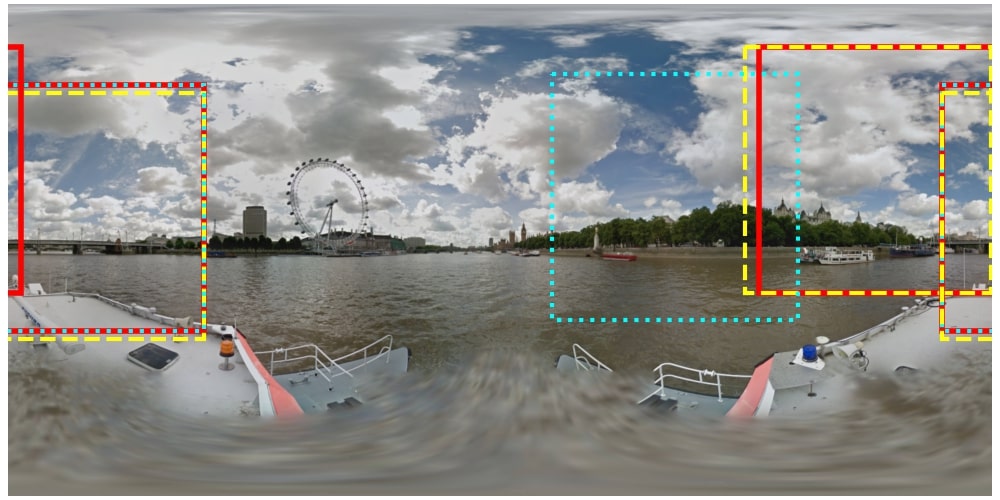}}&
        \T{\includegraphics[width=\sizea, trim={\tal} {\tab} {\tar} {\tat},clip]{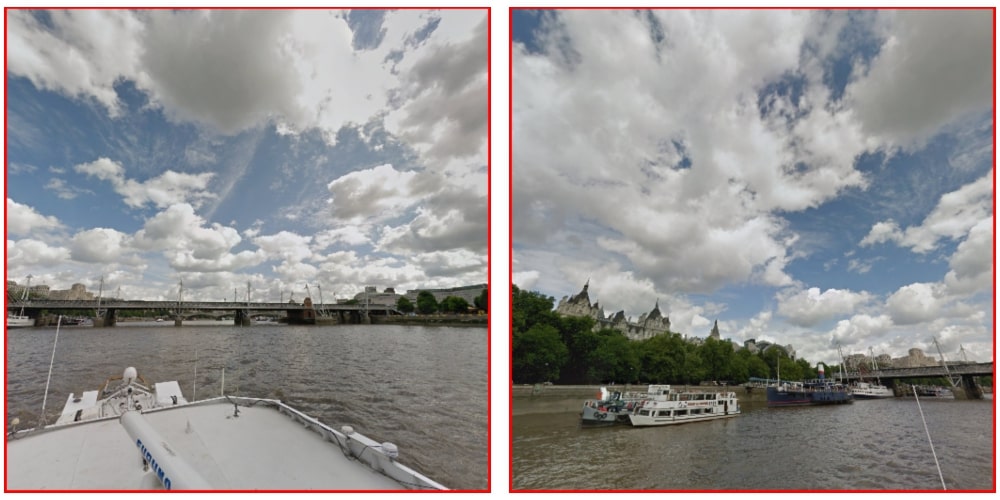}}
        
        \\
        \multirow{3}{*}{\rotatebox[origin=c]{90}{None\whitetxt{sssssssssssss}}} &
        \T{\includegraphics[width=\sizea, trim={\tal} {\tab} {\tar} {\tat},clip]{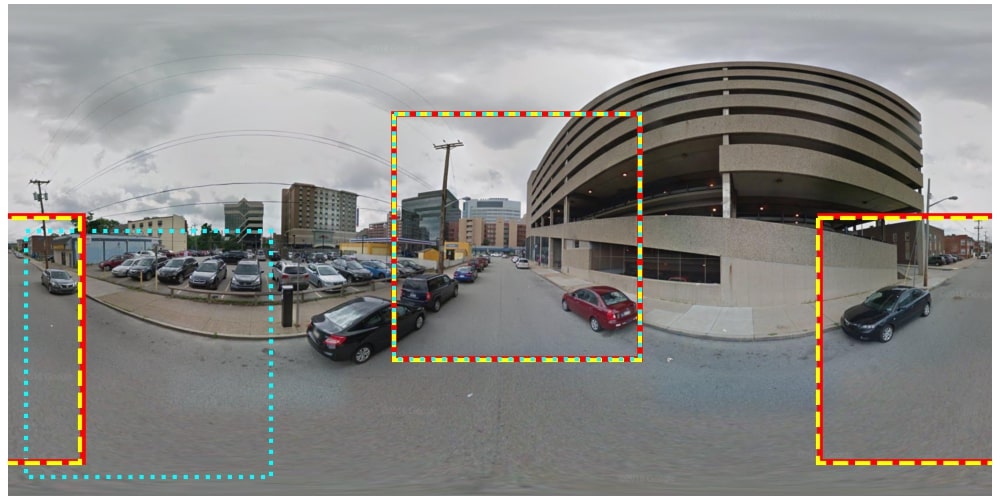}} &
        \T{\includegraphics[width=\sizea, trim={\tal} {\tab} {\tar} {\tat},clip]{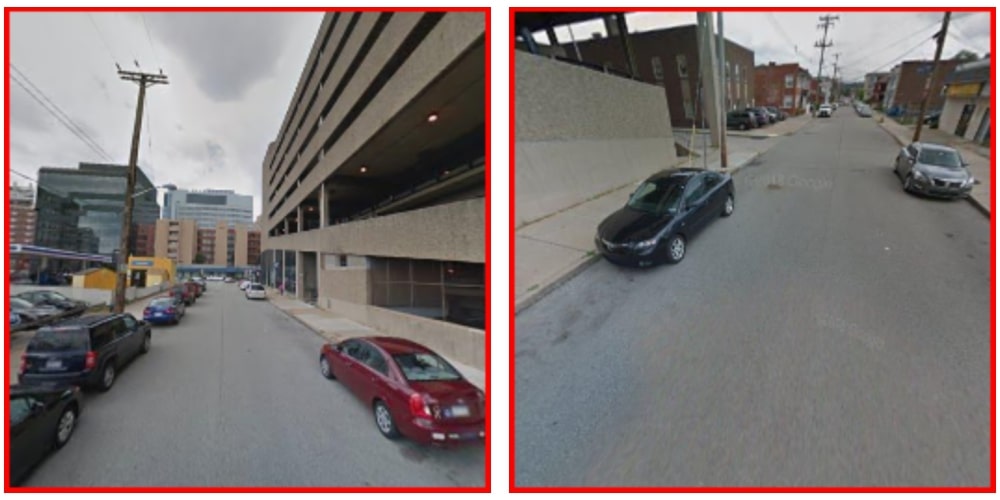}} &
        \T{\includegraphics[width=\sizea, trim={\tal} {\tab} {\tar} {\tat},clip]{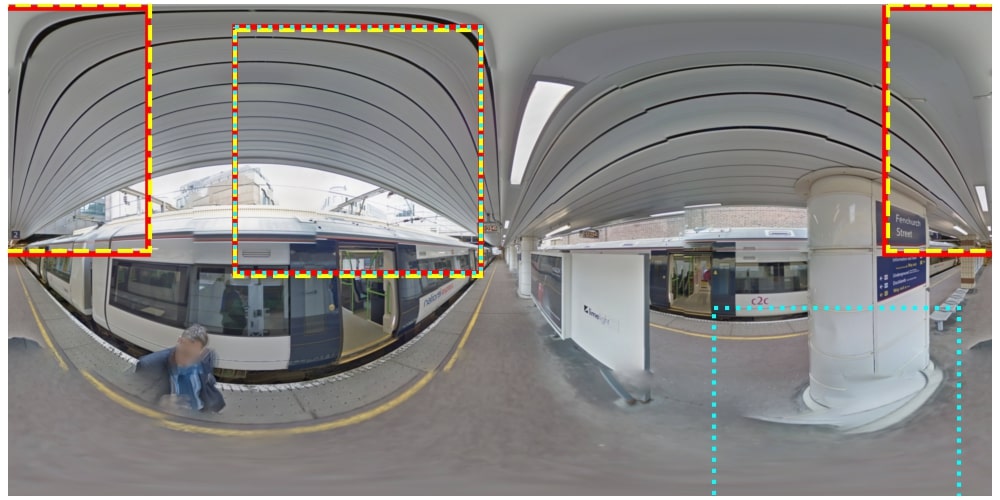}}&
        \T{\includegraphics[width=\sizea, trim={\tal} {\tab} {\tar} {\tat},clip]{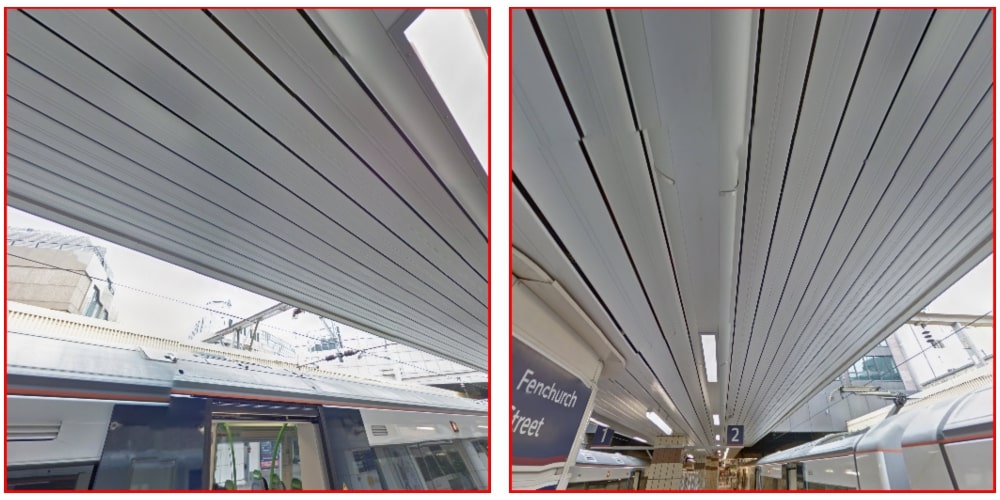}}
        \\ 
        &
        \T{\includegraphics[width=\sizea, trim={\tal} {\tab} {\tar} {\tat},clip]{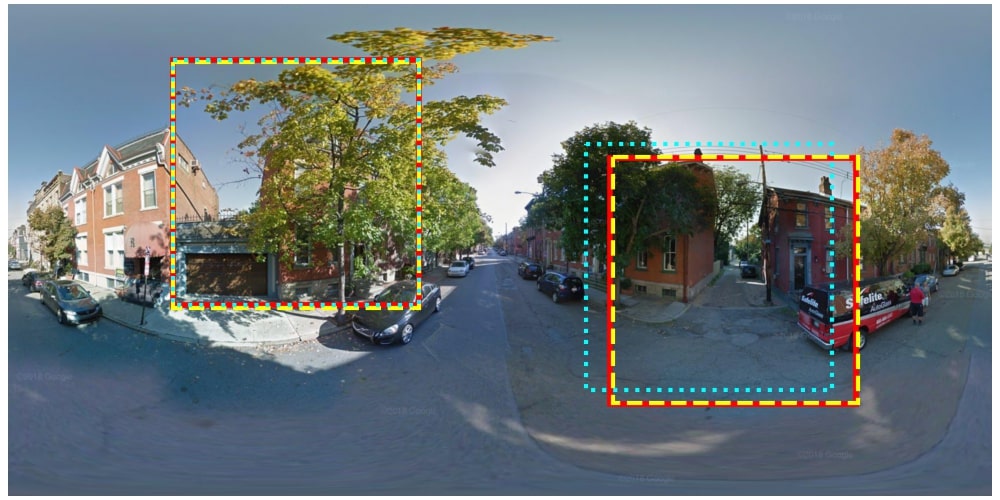}} &
        \T{\includegraphics[width=\sizea, trim={\tal} {\tab} {\tar} {\tat},clip]{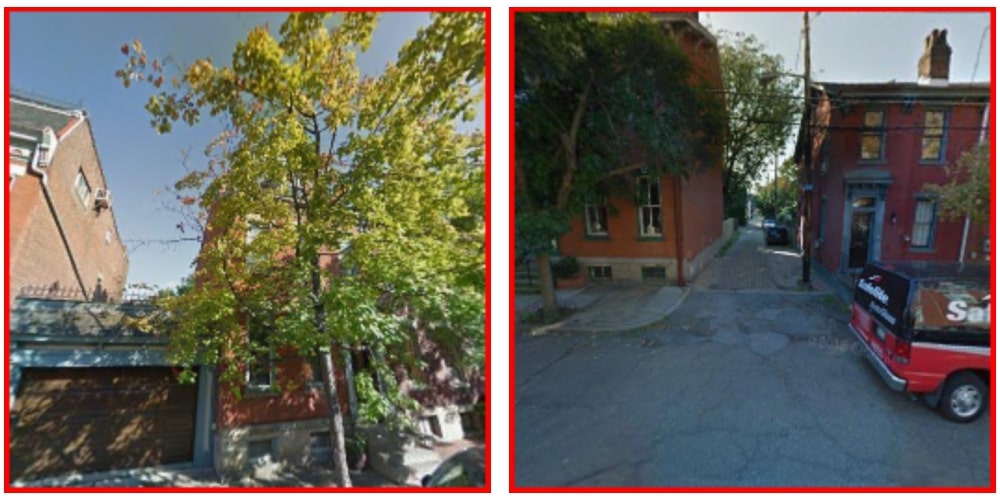}} &
        \T{\includegraphics[width=\sizea, trim={\tal} {\tab} {\tar} {\tat},clip]{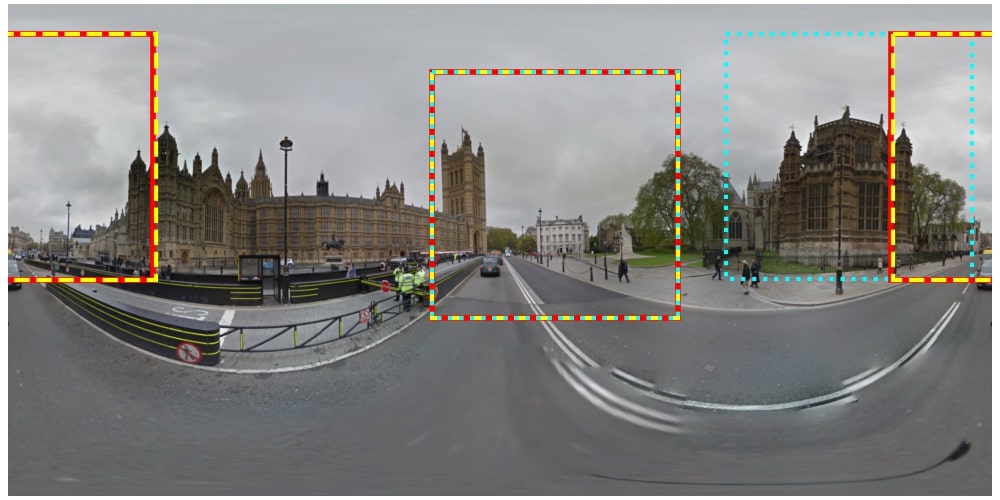}}&
        \T{\includegraphics[width=\sizea, trim={\tal} {\tab} {\tar} {\tat},clip]{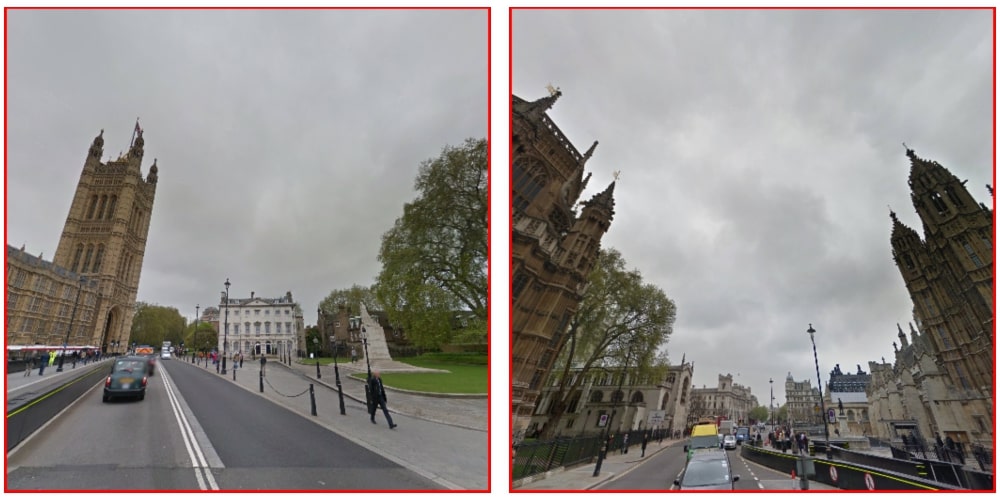}}
       \\ \vspace{+2pt}
       &
       \T{\includegraphics[width=\sizea, trim={\tal} {\tab} {\tar} {\tat},clip]{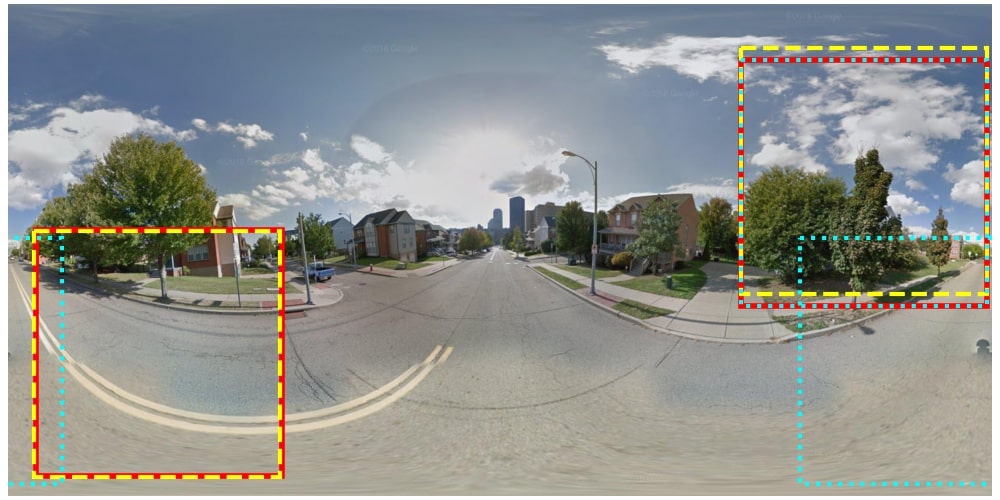}} &
        \T{\includegraphics[width=\sizea, trim={\tal} {\tab} {\tar} {\tat},clip]{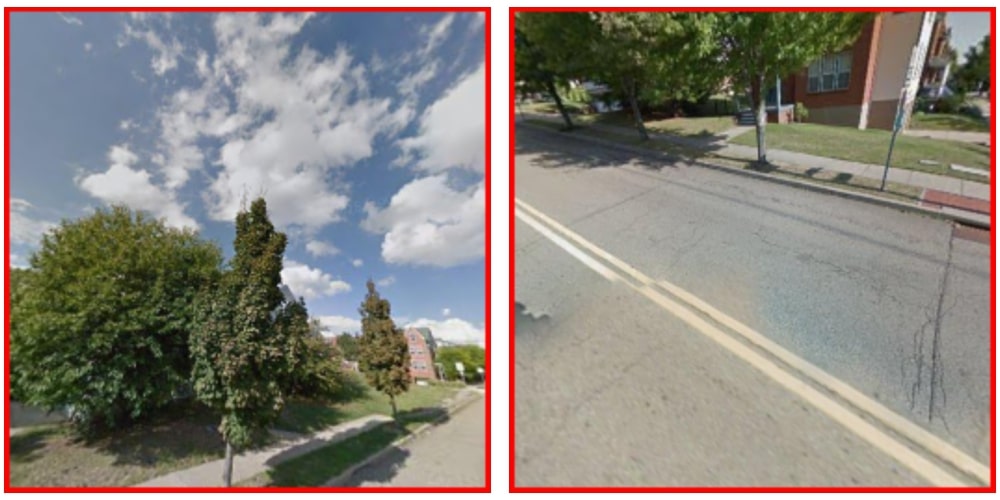}} &
        \T{\includegraphics[width=\sizea, trim={\tal} {\tab} {\tar} {\tat},clip]{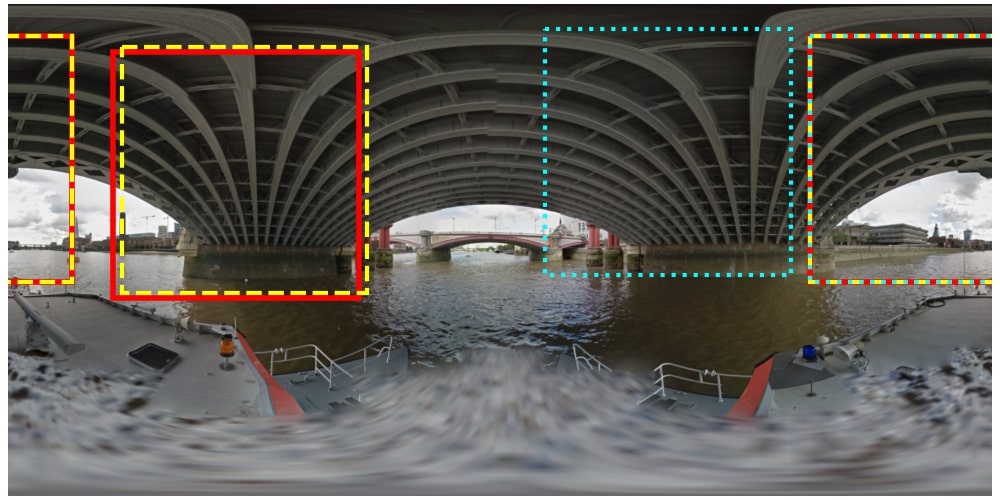}}&
        \T{\includegraphics[width=\sizea, trim={\tal} {\tab} {\tar} {\tat},clip]{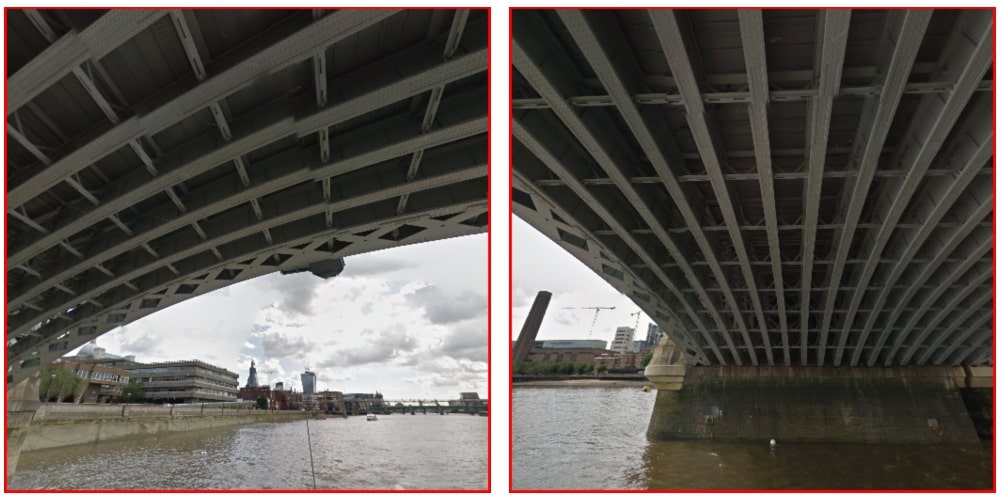}}
        \\ 
        & \multicolumn{2}{c}{Pittsburgh test images}
        & \multicolumn{2}{c}{London test images}
    \end{tabular}
    \end{center}
    \caption{\textbf{Predicted rotation results on new cities.} Full panoramas are shown on the left, with ground-truth perspective images marked in red. We show our predicted viewpoints in yellow and results obtained using the regression model of Zhou \etal~\cite{zhou2019continuity} in blue.
    }
    \label{fig:new_cities}
\end{figure*}

\end{document}